\documentclass[final]{cvpr}

\usepackage{times}
\usepackage{epsfig}
\usepackage{graphicx}
\usepackage{amsmath}
\usepackage{amssymb}
\usepackage{gensymb}
\usepackage{dirtytalk}
\usepackage{mathtools}
\usepackage{multirow}
\usepackage{booktabs}       % professional-quality tables
\newcommand{\bsb}[1]{\boldsymbol{#1}}
\usepackage[pagebackref=true,breaklinks=true,colorlinks,bookmarks=false]{hyperref}

\def\cvprPaperID{2551} % *** Enter the CVPR Paper ID here
\def\confYear{CVPR 2021}
\begin{document}

%%%%%%%%% TITLE
%\title{Modeling Interpretability by Perceptual Simplicity and Spatial Constriction}
\title{Where and What? Examining Interpretable Disentangled Representations}

%\author{Xinqi Zhu \hspace{0.5cm} Chang Xu \hspace{0.5cm} Dacheng Tao\\
\author{Xinqi Zhu, Chang Xu, Dacheng Tao\\
The University of Sydney\\
%School of Computer Science, Faculty of Engineering, The University of Sydney,\\
%6 Cleveland St, Darlington, NSW 2008, Australia\\
{\tt\small \{xzhu7491@uni.,c.xu@,dacheng.dao@\}sydney.edu.au}
}

\maketitle

%%%%%%%%% ABSTRACT
\begin{abstract}
    Capturing interpretable variations has long been one of the goals in
    disentanglement learning. However, unlike the independence assumption,
    interpretability has rarely been exploited to encourage disentanglement
    in the unsupervised setting. In this paper, we examine the
    interpretability of disentangled representations by investigating
    two questions: where to be interpreted and what to be interpreted?
    A latent code is easily to be interpreted if it would consistently
    impact a certain subarea of the resulting generated image. We thus
    propose to learn a spatial mask to localize the effect of each
    individual latent dimension.
    On the other hand, interpretability usually comes from
    latent dimensions that capture simple and basic variations in data.
    %an interpretable code should capture simple
    %variations along dimensions which can be easily identified.
    We thus impose a perturbation on a certain dimension of
    the latent code, and expect to identify the perturbation along this
    dimension from the generated images so that
    the encoding of simple variations can be enforced.
    %On the other hand, considering a
    %perturbation imposed on a certain dimension of the latent code, we
    %expect to identify that perturbed  dimension and its associated
    %perturbation from the generated image.  An auxiliary recognizer is
    %introduced for help to calculate the reconstruction error of latent
    %codes.
    Additionally, we develop an unsupervised model selection method,
    which accumulates perceptual distance scores along axes in the
    latent space. On various datasets, our models can learn
    high-quality disentangled representations without supervision,
    showing the proposed
    modeling of interpretability is an effective proxy for achieving
    unsupervised disentanglement.
\end{abstract}

%%%%%%%%% BODY TEXT
\section{Introduction}
Learning disentangled representations in generative models has gained
increasing interest in recent years \cite{Higgins2017betaVAELB,
Locatello2018ChallengingCA, bau2019gandissect,Karras2020ASG}.
Disentangled representations are supposed to capture independent factors
of variations in data \cite{Bengio2012RepresentationLA},
which should ideally coincide with natural concepts summarized
by humans.
These representations can usually be applied to
various downstream tasks such as controllable image generation and manipulation
\cite{Lample2017FaderNM,Xing2019UnsupervisedDO,Tran2017DisentangledRL,
Kulkarni2015DeepCI,Lee2018DiverseIT},
domain adaptation \cite{Peng2019DomainAL,Cao2018DiDADS},
abstract reasoning \cite{Steenkiste2019AreDR},
and machine learning fairness \cite{Creager2019FlexiblyFR,Locatello2019OnTF}.

Adopting the definition from \cite{Do2020Theory},
we can characterize \emph{disentanglement} from three perspectives:
informativeness, independence, and interpretability.
In the context of unsupervised disentangled representation learning,
the first two properties have been commonly adopted as proxies to
encourage the disentanglement in representations.
Methods built based on the framework of the Generative Adversarial Networks
(GANs) \cite{Goodfellow2014GenerativeAN}
maximize the mutual information between a subset of latent variables
and the generated samples \cite{Chen2016InfoGANIR,Jeon2018IBGANDR,
Lin2019InfoGANCRDG}.
%so that the representations
%can encode the most salient information shown in the data.
On the other hand, the methods based on the Variational Autoencoders (VAEs)
\cite{Kingma2013AutoEncodingVB,Higgins2017betaVAELB,
Kim2018DisentanglingBF,chen2018isolating,
Kumar2017VariationalIO,Jeong2019LearningDA} usually enforce the statistical
independence in latent codes.
%and various regularization terms have
%been imposed on the VAE objective function.
%These models have shown some levels of success in
%learning disentangled representations.

Unlike the informativeness and independence properties,
the interpretability property
in disentanglement has rarely been explored in the unsupervised setting.
Partially due to the meaning of the term
\emph{interpretability} which indicates the correspondence between
the learned representations and human-defined concepts, it sounds
impossible to approach this goal without revealing the ground-truth labels.
Unfortunately, omitting the modeling of interpretability leaves the
existing unsupervised models a huge flaw: the representations satisfying the
informativeness and independence goals are far from unique,
in which case the target representation is indeed included in the
solution pool but not distinguishable from other entangled ones.
A most intuitive example could be the rotation of coordinates
in the latent space, where the existing approaches built
for modeling informativeness and independence are blind to
%the rotations in the latent space
this transformation.
%(see Sec. \ref{sec:tpl} for more detailed explanations).
%For example, the constraint designed in $\beta$-VAE
%\cite{Higgins2017betaVAELB}
%indeed sees the target disentangled representations as a solution,
%but it also sees a lot of entangled representations as solutions
%and it cannot distinguish them based on the designed loss functions
%(\eg rotations in Gaussian distributions).
This nonuniqueness problem is explained by the impossibility conclusion
of unsupervised disentanglement drawn in \cite{Locatello2018ChallengingCA},
and also agrees with the results about the rotation
invariance in \cite{Mathieu2018DisentanglingDI}.
On the contrary, modeling interpretability by providing models with
ground-truth labels solves such nonuniqueness problem, which coincides
with the supervised and semi-supervised settings
\cite{Reed2014LearningTD,Kingma2014SemisupervisedLW,
Dosovitskiy2014LearningTG,Kulkarni2015DeepCI,
Yan2016Attribute2ImageCI,Lample2017FaderNM,Locatello2020WeaklySupervisedDW,
Locatello2020Disentangling}.

A rising problem is, can we enforce interpretability in representations
without supervision?
%Indeed, we cannot achieve guaranteed interpretability without accurate
%guides from the labels, we argue that we can still encourage a model
%to preferably converge to an interpretable disentanglement by
%imposing some general assumptions about the data.
A precise matching between the target concepts and
the learned representations is unrealistic because
it depends on how the target solution is defined.
For example, digital color can be represented in RGB or HSV,
but an unsupervised model does not know which one is more preferable
without being told which one is wanted.
However in more general cases, there is no doubt that
interpretable variations are
identifiable out of noninterpretable ones by humans without effort,
\ie the complex world is decomposed into basic concepts
that is comprehensible to most people.
%\eg we can tell the separate factors controlling smiling and
%hair color are more interpretable than those controlling
%any mixtures of these two variations,
%no matter if they are statistically independent.
The insight is that there exist some general biases
in humans' definition of concepts, and they can be borrowed to
heuristically guide a model to prefer a more interpretable
representation than a noninterpretable one.
%These general patterns can be something like:
%a concept should be easily understandable,
%or a concept should be visually separable from each other, \etc.
%%or a concept should be distinct from each other, \etc.
These biases are not precise knowledge about individual concepts,
but some general information that
is assumed to be shared by the interpretable concepts,
%we are interested in learning,
so that noninterpretable ones are filtered out.

In this paper, we exploit two hypotheses about interpretability
to learn disentangled representations.
The first one is \emph{Spatial Constriction}:
%the concept represented by
%a latent code should be spatially restricted in an image.
a representation is usually interpretable if we can consistently tell
where the controlled variations are in an image.
%The second hypothesis is that the integrated perceptual distance
%of the variations in data space
%along the latent axes should be shorter than it along other directions.
The second hypothesis is \emph{Perceptual Simplicity}: an interpretable
code usually corresponds to a concept
consisting of perceptually simple variations.
%We develop two modules based on these assumptions.
For the first one, we design a module to
restrict the impact of each
latent code in specific areas on feature maps during generation.
For the second one, we design a loss to encourage the model
to embed simple data variations along each latent dimension.
%while training an encoder for a GAN.
These two contributions are orthogonal and can be used jointly.
%In addition, we show that we can quantitatively differentiate the
%entangled and disentangled representations in Fig. \ref{fig:rot_impression}
%(a) and (b) by computing their accumulated traversal perceptual lengths (TPLs),
%which serves as an approach for unsupervised model selection
%in disentanglement learning.
In addition, we show that for a disentangled model, its accumulated perceptual
distance along latent axes are generally smaller than on other latent
directions. This observation corresponds
to the Perceptual Simplicity assumption,
and inspires us to propose an unsupervised model selection method.
%for learning disentangled representations.
We conduct experiments on various datasets including CelebA, Shoes,
Clevr, FFHQ, DSprites and 3DShapes to evaluate our proposed modules.
%unsupervised disentanglement learning.
%Particularly on 512 $\times$ 512 FFHQ, our models can learn
%many interpretable concepts such as azimuth,
%smile, gender, hair color, \etc (see Fig. \ref{fig:traversal_1}
%and Fig. \ref{fig:traversal_full}
%for an impression, and refer \texttt{traversals.gif} in the
%supplementary material to view more learned semantics with animation).
We also conduct experiments to show that the
proposed TPL score is an effective method for
unsupervised model selection. These experiments justify modeling
of interpretability in learning disentangled representations.

% !!! Add in the model description section.
%The most economic learning strategy of the model is to quickest converge
%to a state that is able to separate the image variations
%controlled by each dimension, so that it can
%most efficiently update the weights, leading to the axis-specific changes.
%For the model to quickest converge to a state that
%can distinguish the types of variations
%on each latent dimension, the generator should thus learn to synthesize
%images whose variations along individual dimensions are
%\emph{easiest to be distinguished}.
%In summary, this loss not only
%trains an encoder of a GAN as in \cite{Chen2016InfoGANIR,Jeon2018IBGANDR},
%but implicitly encourages the simplest separation of variations
%corresponding to each latent dimension (similar to the
%goal in \cite{Lin2019InfoGANCRDG,VPdis_eccv20}).

\section{Related Work}
\label{sec:related_work}
Learning interpretable representations has been
commonly tackled as a supervised
or semi-supervised problem for a long time \cite{Reed2014LearningTD,
Kingma2014SemisupervisedLW,Dosovitskiy2014LearningTG,Kulkarni2015DeepCI,
Yan2016Attribute2ImageCI,Lample2017FaderNM} under the subject of
attributes-based generation and conditional generation
(matching the interpretability property defined in \cite{Do2020Theory}),
until the emergence of unsupervised models like the InfoGAN variants
\cite{Chen2016InfoGANIR,Jeon2018IBGANDR,Lin2019InfoGANCRDG}
and the VAE variants
\cite{Zhao2017LearningHF,Higgins2017betaVAELB,Burgess2018UnderstandingDI,
Kumar2017VariationalIO,Kim2018DisentanglingBF,chen2018isolating,
Dupont2018LearningDJ,Jeong2019LearningDA,Li2020ProgressiveLA}.
%Various definitions of disentanglement have been proposed by
%describing different aspects of it
%\cite{higgins2018definition,Eastwood2018AFF,Do2020Theory},
%and we find the one from \cite{Do2020Theory} characterizing
%informativeness, independence, and interpretability to be
%most commonly accepted.
These unsupervised methods achieve disentanglement from different
directions. The first type is to model the informativeness in latent
codes, such as InfoGAN \cite{Chen2016InfoGANIR}
which maximizes the mutual information between a subset of latent variables and
the generated images, and IB-GAN \cite{Jeon2018IBGANDR} which imposes
another upper bound of the informativeness.
Lin \etal \cite{Lin2019InfoGANCRDG} equip InfoGAN with a contrastive
regularizer, which detects the shared dimension in latent codes of the generated
image pairs. The second type is to model the statistical independence
in the encoded latent variables based on the VAE framework,
starting with the $\beta$-VAE model
\cite{Higgins2017betaVAELB,Burgess2018UnderstandingDI}
which modulates the prior matching term with
a coefficient $\beta$ in the evidence lower bound objective.
Other VAE variants consist of methods minimizing the
total correlation in latent variables via factorizing the aggregated posterior
\cite{Kim2018DisentanglingBF}, moment-matching betweeen the prior
and aggregated posterior distributions \cite{Burgess2018UnderstandingDI},
weighted sampling \cite{chen2018isolating}, and sequentially
relieving the $\beta$ coefficient for different dimensions
during training \cite{Jeong2019LearningDA}.

Other disentanglement methods include
exploiting the hierarchical nature of deep networks in the VAE framework
\cite{Zhao2017LearningHF,Li2020ProgressiveLA} and the GAN framework
\cite{Karras2019AnalyzingAI,Karras2020ASG}. \cite{singh-cvpr2019}
disentangles background, shape and appearance in images
in a hierarchical manner by designing a three-stage architecture.
There are works achieving disentanglement by manipulating subparts of
a latent code. Recent content-style disentanglement
techniques \cite{Gatys2016ImageST,Ulyanov2016TextureNF,Huang2017ArbitraryST,
Kazemi2019StyleAC} can be seen as designing losses by manipulating two
groups of features independently to achieve disentanglement of
content and style, based on the hypothesis of how
content and style information should be encoded in deep generative architectures.
In the domain adaptation area, the learned content features are supposed to
be disentangled from the domain information,
where the varied domain labels serve as a supervision for
the disentanglement \cite{Cao2018DiDADS,Peng2019DomainAL}.
For more general disentanglement learning tasks, grouping information
defined by sharing a subset of latent codes inside a group of data can
be exploited to guide their disentanglement with unshared codes
(identity vs pose) \cite{Bouchacourt2018MultiLevelVA,Hosoya2019GroupbasedLO}.
Different from the existing approaches, we propose to exploit the
interpretability in disentangled representations as a proxy to achieve
the unsupervised disentanglement goal.

%-------------------------------------------------------------------------
\section{Methods}
\label{sec:methods}
\begin{figure}[t]
\begin{center}
   \includegraphics[width=\linewidth]{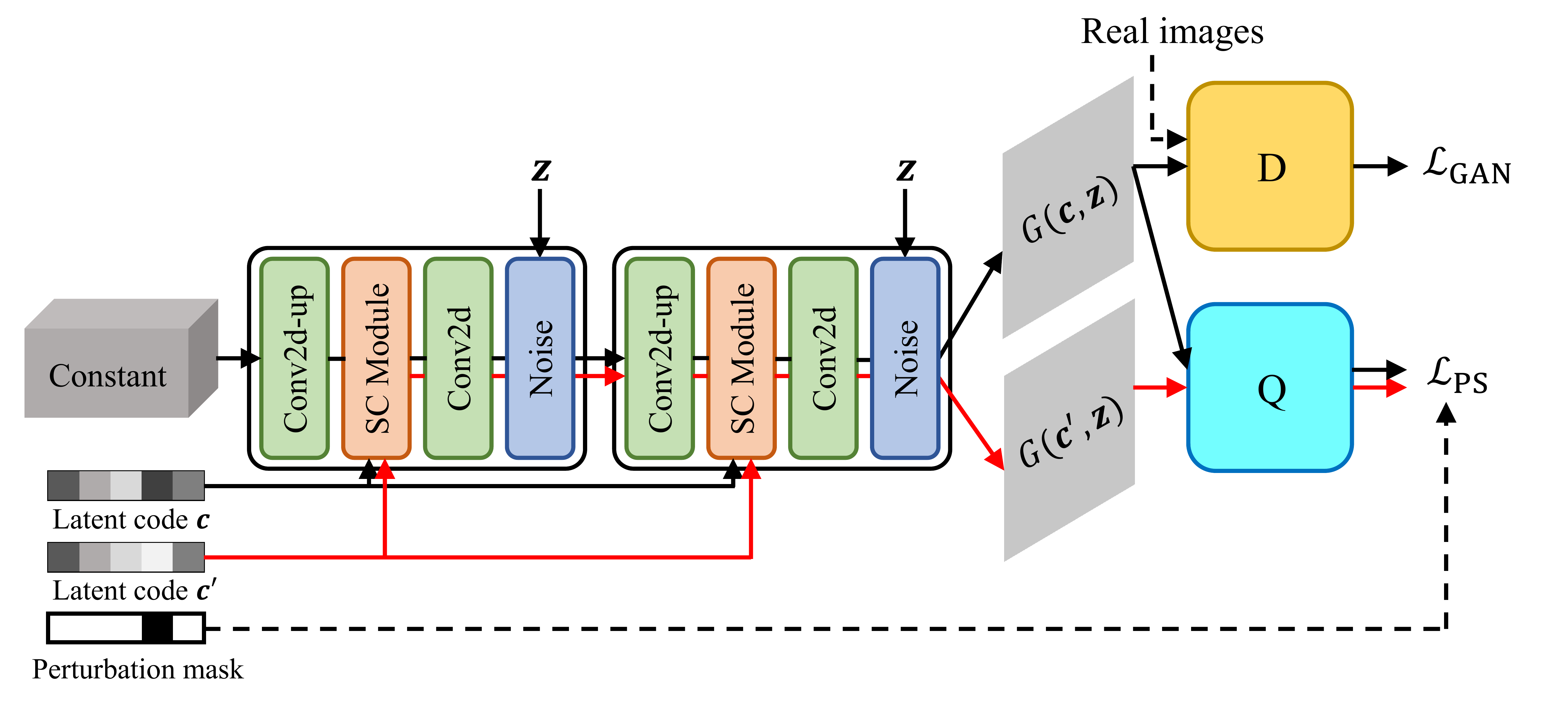}
\end{center}
    \caption{Overview of our proposed PS-SC model.}
\label{fig:architecture}
\end{figure}

We first introduce a module to realize the
Spatial Constriction (Sec. \ref{sec:sc_module}),
then we introduce a loss to enforce the Perceptual Simplicity
assumption (Sec. \ref{sec:pc}).
%Both contributions encourage the axis-alignment objective in
%disentanglement learning without glancing at ground-truth labels.
%The two contributions are orthogonal to each other, and an illustration
%of the combined model is shown in Fig. \ref{fig:architecture}.
A combined model is shown in Fig. \ref{fig:architecture}.
In Sec. \ref{sec:tpl}, we introduce a simple approach
to achieve unsupervised model selection.

\subsection{Enforcing Spatial Constriction}
\label{sec:sc_module}
For a latent code to represent interpretable variations in the image space,
it is usually natural to assume that these variations happen in a
consistently constricted area. For example, if a code is to control
the variation of \emph{fringe}
on a human face, then it should mainly focus on the upper part of the face to
generate how the fringe should be shaped, without paying much attention to
other parts like the background.
%This procedure mimics humans' behavior
%of drawing a picture, during which we focus
%on a subarea of our canvas to detail the sketch
%of a specific component.
We embed this \emph{constricted modification} idea
into generative models.
%by designing a module to restrict
%the manipulatable areas of a latent code.
The key point of our design is that the constricted areas should be shaped by
low degrees of freedom so that a simple and compact area could be constructed,
which is more preferable in terms of interpretability.

How can we simulate \emph{constricted modification} with neural networks?
For the \emph{modification} procedure alone, we can adopt the idea of
adaptive normalization (AdaIN),
a module developed based on instance normalization for
style transferring tasks \cite{Ulyanov2017ImprovedTN,Gatys2016ImageST,
Huang2017ArbitraryST,Kazemi2019StyleAC}, and has been
used for general image generation
\cite{Karras2020ASG,Karras2020AnalyzingAI}. The AdaIN is defined as:
$\text{AdaIN}(\bsb{x}, \bsb{y}) = \sigma(\bsb{y})\Big
(\frac{\bsb{x} - \mu(\bsb{x})}{\sigma(\bsb{x})} \Big ) + \mu(\bsb{y})$,
where $\bsb{x}$ denotes the content input and $\bsb{y}$ denotes
the style input, and $\mu$, $\sigma$ compute
the mean and standard deviation across spatial dimensions.
%If the style input is conditioned on some latent code $\bsb{z}$:
%$\bsb{y} = f(\bsb{z})$, then we can interpret this module as
%modifying the input $\bsb{x}$ with some information from the latent
%code $\bsb{z}$.
To simulate \emph{constricted} modification, it is natural to consider using the
heatmaps computed by the $\mathtt{softmax}$ layer in attention modules
\cite{Vaswani2017AttentionIA,Cao2019GCNetNN,Li_2020_CVPR} to
highlight the focused areas of a latent dimension.
However the $\mathtt{softmax}$
transformation forces the activations to have a summation of 1,
which is more effective for weighted aggregation of features
instead of localized modification
%(the comparisons between using $\mathtt{softmax}$ and our proposed module
%are shown in the experiments).
(see Sec. \ref{sec:faces_exp} for an empirical comparison).
%and the $\mathtt{sigmoid}$ function is prone to output distributed
%activations rather than simply-shaped masks.
Instead, we leverage a gating layer called
$\mathtt{cumax}$, an activation function proposed by \cite{Shen2019OrderedNI},
originally used for structured language modeling,
to realize our goal. The $\mathtt{cumax}$ is defined as:
$\bsb{g} = \mathtt{cumax}(...) = \mathtt{cumsum}(\mathtt{softmax}(...))$,
where the $\mathtt{cumsum}$ function denotes the cumulative summation.
The $\mathtt{cumax}$ layer in practice transforms
a vector of neurons into a soft version of
binary gates $\hat{\bsb{g}} = (0, ..., 0, 1, ..., 1)$, since the
$\mathtt{softmax}$ usually results in a hump in a vector.
By combining two of this function with an element-wise product
$\bsb{g} = \mathtt{cumax}(...) \odot (\bsb{1} - \mathtt{cumax}(...))$,
we create a learnable binary gate (band-pass-filter shaped)
with degrees of freedom on both ends.
%which will be used to generate masks on feature maps.
%When we have this gate on both the height and width of a feature map,
%a mask can be computed by outer product: $\bsb{g}^{h} \otimes \bsb{g}^{w}$.

\begin{figure}[t]
\begin{center}
   \includegraphics[width=\linewidth]{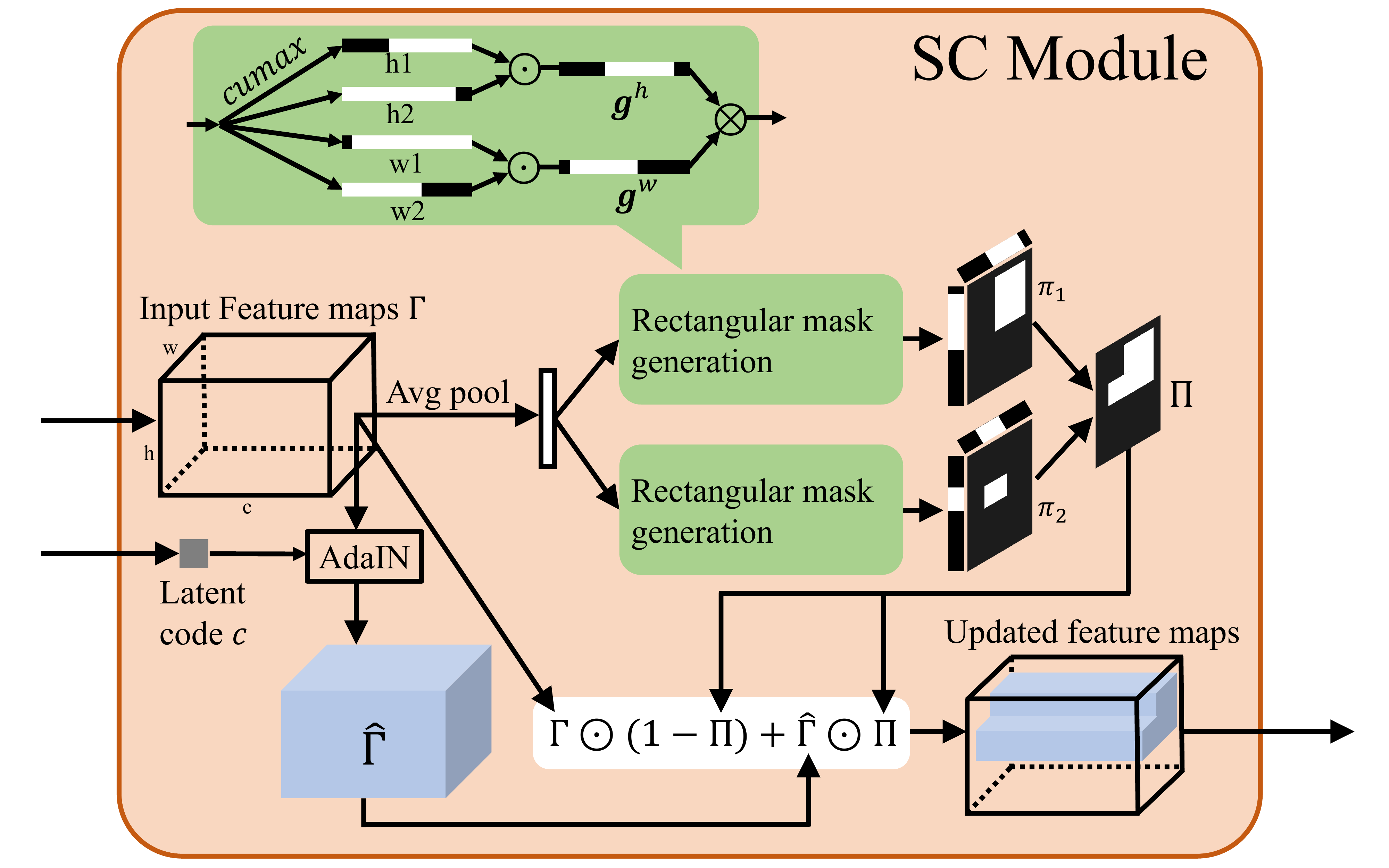}
\end{center}
    \caption{Detailed illustration of the SC module.}
\label{fig:sc_module}
\end{figure}

The proposed Spatial Constriction (SC) module is
illustrated in Fig. \ref{fig:sc_module}.
We use $\bsb{\Gamma}$ to denote the input feature maps,
and $c$ to be the input latent code.
The learnable gate for the height dimension is computed as:
\begin{align}
    \bsb{v}^{h1} &= f^{h1}(\mathtt{avgpool}(\bsb{\Gamma})), \\
    \bsb{v}^{h2} &= f^{h2}(\mathtt{avgpool}(\bsb{\Gamma})), \\
    \bsb{g}^{h} &= \mathtt{cumax}(\bsb{v}^{h1}) \odot
    (\bsb{1} - \mathtt{cumax}(\bsb{v}^{h2}),
\end{align}
where $f^{h1}$ and $f^{h2}$ are functions to map vectors to the length of
the height of the input feature maps. Similarly, we can get
the gate for width dimension $\bsb{g}^{w}$.
A mask on feature maps is computed by performing an outer
product:
\begin{align}
    \bsb{\pi} = \bsb{g}^{h} \otimes \bsb{g}^{w}.
\end{align}
This is a differentiable rectangular mask with learnable sides and position.
We allow a SC mask to be more flexible than a single rectangular and we use
the sum of $J$ rectangles as a direct solution:
\begin{align}
    \bsb{\Pi} = \frac{1}{J}\sum_{j=1}^{J} \bsb{\pi}_{j}. \label{eq:focus_map}
\end{align}
Then a latent code modifies the content on the
input feature maps conditioned on the SC mask:
\begin{align}
    \hat{\bsb{\Gamma}} &= \text{AdaIN}(\bsb{\Gamma}, f(c)), \\
    %\text{SC}(\bsb{\Gamma}, c) &= \bsb{\Pi} \odot
    %\hat{\bsb{\Gamma}} + (\bsb{1} - \bsb{\Pi}) \odot \bsb{\Gamma}
    \text{SC}(\bsb{\Gamma}, c) &= \bsb{\Gamma} \odot (\bsb{1} - \bsb{\Pi}) +
    \hat{\bsb{\Gamma}} \odot \bsb{\Pi}
    \label{eq:draw},
\end{align}
where $\hat{\bsb{\Gamma}}$ is the content modified by code $c$.
%The new content is drawn in the
%constricted subarea $\bsb{\Pi}$ of the input feature maps $\bsb{\Gamma}$.
Function $f$ maps code $c$
to the length of the input channel number of $\bsb{\Gamma}$,
which is required by $\text{AdaIN}$.

\subsection{Encouraging Perceptual Simplicity}
\label{sec:pc}
The second hypothesis about interpretability in
representations is that the data variations captured by individual latent
dimensions should be perceptually simple
(see Sec. \ref{sec:tpl} for a quantitative example).
%This assumption has been used in previous works
%\cite{Lin2019InfoGANCRDG,VPdis_eccv20}, but either of them
%requires to train a separate network to predict which
%dimension in the latent code has or has not been changed.
In this section, we introduce a loss to integrate this
goal into training an encoder for a GAN.
We first introduce how this loss is defined, and then discuss
why such a simple implementation works.

We assume a generator $G$ takes a vector of latent code
$\bsb{c} \in \mathbb{R}^{d}$
and a vector of noise $\bsb{z}$ to generate an image:
$\bsb{x} = G(\bsb{c}, \bsb{z})$, following the notations in
InfoGAN \cite{Chen2016InfoGANIR} (see Sec \ref{ap:gan_infogan}
in the Appendix for a brief introduction about GAN and InfoGAN).
The noise code $\bsb{z}$ is to provide $G$ with sharp details about
an image, which will be omitted in the following text,
and the latent code $\bsb{c}$ is to learn interpretable information.
We impose a perturbation on a randomly
selected dimension in the latent code $c'_k = c_k + p$,
where $k \sim \mathcal{U}_{\text{int}}(0, d-1)$,
$p \sim \mathcal{N}(c_k, p_{\text{var}})$ and $p_{\text{var}}$
is a hyper-parameter. Then we get another image $\bsb{x}'$
generated by the altered latent code $\bsb{x}' = G(\bsb{c}')$ where
$\bsb{c}' = \{\bsb{c}_{\setminus k}, c'_k\}$ .
%For a latent code $\bsb{z}$ of a generator $$
Then we introduce a recognizer $Q$, whose primary goal is to
reconstruct the latent code $\bsb{c}$ and $\bsb{c}'$ based on
the generated images $\bsb{x}$ and $\bsb{x}'$
($\hat{\bsb{c}} = Q(\bsb{x})$, $\hat{\bsb{c}}' = Q(\bsb{x}')$)
respectively with MSE loss.
However, in order to enforce the simple encoding along
dimensions of the latent code, we substitute the
errors computed on the shared dimensions by the
errors between the truth code and the average of both reconstructed values,
leading to the dimension-wise loss defined as:
\begin{align}
    loss_{i} =
    \begin{cases}
        (\hat{c}_i - c_i)^{2} +
        (\hat{c}'_i - c'_i)^{2},
        & \text{if } i = k \\
        2 \times (\frac{\hat{c}_i + \hat{c}'_i}{2}
        - c_i)^{2},
        & \text{if } i \neq k
    \end{cases},
\end{align}
where the $\hat{\bsb{c}}$ and $\hat{\bsb{c}}'$ are outputs from $Q$,
and $k$ is the perturbed dimension index.
Note that $c_i = c'_i$ if $i \neq k$.
%We sum the losses computed on both the original code $\bsb{c}$ and the
%perturbed one $\bsb{c}'$.
We sum the losses on all dimensions to form the complete loss,
which is named as Perceptual Simplicity (PS) loss:
\begin{align}
    %\mathcal{L}_{\text{PS}} &=
    %\frac{1}{Nd} \Big[\bsb{m}^{T} \cdot (\hat{\bsb{c}} - \bsb{c})^{2}
    %+ (\bsb{1} - \bsb{m})^{T} \cdot \big(\bar{\hat{\bsb{c}}} - \bsb{c}\big)^{2} \nonumber \\
    %&+ \bsb{m}^{T} \cdot (\hat{\bsb{c}}' - \bsb{c}')^{2}
    %+ (\bsb{1} - \bsb{m})^{T} \cdot \big(\bar{\hat{\bsb{c}}} - \bsb{c}'\big)^{2}\Big], \\
    %%\widehat{Q(\bsb{x})} &= \widehat{Q(\bsb{x'})} =
    %%\frac{1}{2}(Q(\bsb{x}) + Q(\bsb{x}')),
    %\bar{\hat{\bsb{c}}} &= \frac{\hat{\bsb{c}} + \hat{\bsb{c}}'}{2},
    % ---
    %\mathcal{L}_{\text{PS}} &=
    %\frac{1}{Nd} \Big[(\hat{\bsb{c}}_{i} - \bsb{c}_{i})^{2}
    %+ (\hat{\bsb{c}}'_{i} - \bsb{c}'_{i})^{2}
    %+ 2 \big(\bar{\hat{\bsb{c}}}_{\setminus i}
    %- \bsb{c}_{\setminus i}\big)^{2} \Big]\\
    %\bar{\hat{\bsb{c}}} &= \frac{\hat{\bsb{c}} + \hat{\bsb{c}}'}{2},
    % ---
    \mathcal{L}_{\text{PS}} &= \frac{1}{d}\sum_{i=0}^{d-1}loss_{i}.
\end{align}
This is similar to an ordinary reconstruction loss on latent codes,
but with the losses on shared dimensions calculated in a fuzzy way.
%During implementation, we use a onehot mask $\bsb{m}$ to indicate
%the non-shared $k^{\text{th}}$ dimension, which is
%randomly sampled $k \sim \mathcal{U}_{\text{int}}(0, d-1)$.

\textbf{Discussion}:
The PS loss is more tolerant of the misalignment on the shared dimensions
than on the perturbed dimension, since it only requires the mean of the
shared two latent-code reconstructions to match the truth code.
In other words, the loss will punish the model more on the mistakes made
along the non-shared dimension than on the other directions,
forcing the generator to
embed more easily recognizable variations along this specific latent axis
so that the recognizer $Q$ can more easily regress to the truth
value on this dimension.
Since the non-shared dimension is randomly selected in each iteration,
after convergence the $Q$ will still be an encoder.
The generator $G$ should find a solution that the
data variations controlled
by each latent dimension are necessarily simple to be interpreted,
but the coupled data variations controlled by all dimensions are rich enough
to form data matching the training distribution.
Our PS loss is similar to a series
of losses from VAE-based models \cite{Bouchacourt2018MultiLevelVA,
Hosoya2019GroupbasedLO,Locatello2020WeaklySupervisedDW},
but the paired images used in these works are picked
by varying a known attribute with supervision,
while ours does not rely on any labels.

\begin{figure}[t]
\begin{center}
   \includegraphics[width=\linewidth]{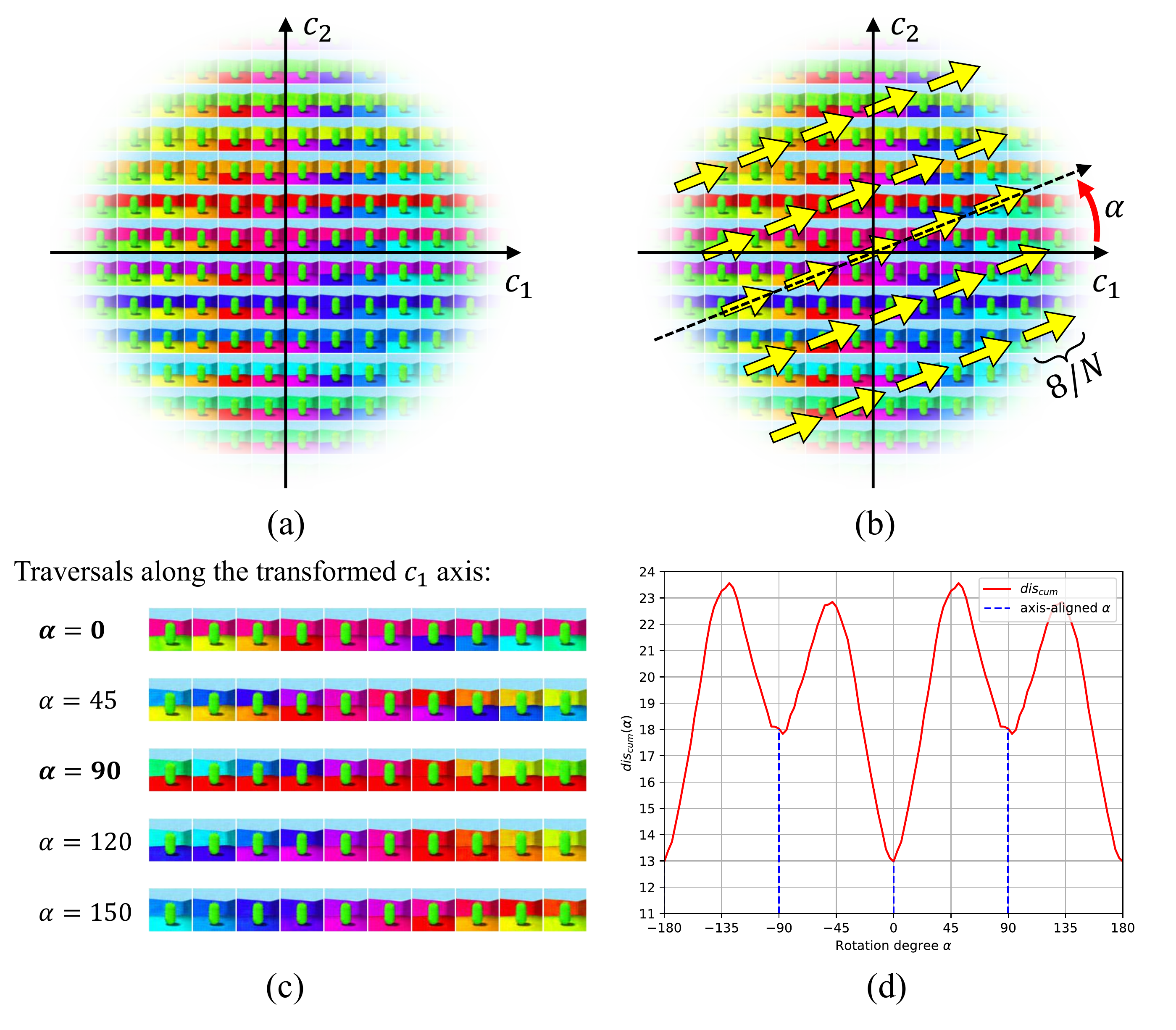}
    \vspace{-20pt}
\end{center}
    \caption{(a) A 2D disentangled representation capturing
    semantics of pure wall color and floor color.
    (b) Illustration of how the $\text{dis}_{cum}$
    (Eq. \ref{eq:cum_c1_c2}) is quantified in the $(c_1, c_2)$ space.
    Perceptual distance scores along yellow arrows are accumulated.
    %We accumulate the
    %perceptual distances computed by image pairs that are generated by
    %latent codes differing in the direction $\alpha$ with a small distance
    %(indicated by yellow arrows).
    (c) Latent traversals for different $\alpha$'s along axis $c_1$.
    (d) The plot showing how the
    $\text{dis}_{cum}$ value changes with rotation degree $\alpha$.}
    %\vspace{-10pt}
\label{fig:tpl_explain}
\end{figure}

\subsection{Traversal Perceptual Length}
\label{sec:tpl}
In this section, we first show a concrete example
of the Perceptual Simplicity assumption in interpretability,
%interpretability that
%\say{each latent code should correspond to \emph{simple} variations},
then we introduce an unsupervised model selection method.

Fig. \ref{fig:tpl_explain} (a) shows a disentangled representation
(with $c_1$ and $c_2$ dimensions)
capturing two data variations from 3DShapes dataset
\cite{Kim2018DisentanglingBF} (wall color vs floor color).
This representation is interpretable since we can tell that each
dimension encodes pure wall color and floor color respectively.
Then we apply a rotation on the coordinate system in this latent space
by $\alpha$ degrees as shown in Fig. \ref{fig:tpl_explain} (b).
Because standard Gaussian distribution is rotation-invariant, this
transformation does not break the statistical independence property of
the representation. However it breaks the interpretability property
since the traversals along individual axes are not controlling pure
variations (see traversals in Fig. \ref{fig:tpl_explain} (c)).
%It is not surprising for a human to notice the break of interpretability,
%but the interesting point is that a model can also feel the difference
%without referring to labels by accumulating perceptual distances
%\cite{zhang2018perceptual} along the axis in this
%latent space (see Fig. \ref{fig:tpl_explain} (b) for an
%illustration):
This absence of interpretability is noticeable by humans,
but can a model sense it without labels?
To show this is possible,
we define the accumulated perceptual distance ($\text{dis}_{cum}$)
by traversing along the transformed $c_1$ axis in the 2D space
(see Fig. \ref{fig:tpl_explain} (b) for an
illustration):
\begin{align}
    \text{dis}_{cum}(\alpha) = \sum_{\mathclap{\substack{(c_1, c_2) \in
    \text{grid}(-4, 4)}}} \text{dis} \Big(
    &G\big( \mathtt{R}(\alpha) (c_1, c_2) \big), \nonumber \\
    &G\big(\mathtt{R}(\alpha) (c_1 + \frac{8}{N}, c_2)\big) \Big),
    \label{eq:cum_c1_c2}
\end{align}
where $\text{grid}(-4, 4)$ is an $N \times N$ grid with coordinates ranging
from $-4$ to $4$, and $\text{dis}(...)$ denotes perceptual distance
computation using VGG16 \cite{Simonyan15}.
The $\mathtt{R}(\alpha)$ is a rotation matrix in 2D space parameterized by
degree $\alpha$.
The plot of $\text{dis}_{cum}$ vs $\alpha$
is in Fig. \ref{fig:tpl_explain} (d).
%The point $\alpha = 0$ is the current disentangled state.
As we can see, when the coordinate system is aligned
with the interpretable axes in Fig. \ref{fig:tpl_explain} (a)
($\alpha=-90,0,90,180$),
the $\text{dis}_{cum}$ scores become
local minima (indicated by the blue dash lines).
This experiment indicates that though a disentangled representation
is indeed isotropic in terms of statistical independence,
it is \emph{perceptually anisotropic}, with variations along latent axes
being simpler than other directions (more examples are shown in
the Appendix Sec. \ref{ap:tpl_explain_more}).
%Note that
%the direction $\alpha = -45$ controls pure floor color
%in images (axis $c'_1$ in Fig. \ref{fig:rot_impression} (b)),
%and direction $\alpha = 45$ controls pure wall color
%(axis $c'_2$ in Fig. \ref{fig:rot_impression} (b)).
%This \say{interpretable directions are perceptually local minima}
%phenomenon is commonly observed in other attribute-pairs and
%other datasets. More experiments can
%be found in the Appendix Sec. \ref{ap:tpl_explain_more}.

%This observation shows that interpretable variations are perceptually
%simpler than other kinds of variations.
This perceptual-anisotropy phenomenon inspires us to
develop an unsupervised model selection
method, since we can assume the overall
accumulated perceptual distance scores along all latent axes to
be generally small in disentangled representations.
The method, namely Traversal Perceptual Length (TPL), is defined as follows:
%in which it aggregates
%the traversal perceptual distances along all activated latent dimensions:
\begin{align}
    \text{tpl}&_{i}(G)
    = \mathbb{E}_{\bsb{c}} \sum_{\mathclap{c_i \in \text{lin}(-4, 4)}}
    \text{dis}\big(
    G(\bsb{c}_{\backslash i}, c_{i}),
    G(\bsb{c}_{\backslash i}, c_{i}+\frac{8}{N})\big), \label{eq:tpl_i} \\
    \text{tpl}&(G) = \sum_{i=0}^{d-1}
    \text{act}_i\cdot\text{tpl}_{i}(G), \quad
    \text{act}_i =
        \begin{cases}
            1, & \text{tpl}_{i}(G) \geq S \\
            0, & \text{tpl}_{i}(G) < S \\
        \end{cases},
\end{align}
where $\text{lin}(-4, 4)$ denotes linearly spaced $N$ values in interval
$(-4, 4)$, and $S$ is a threshold to determine if a dimension
encodes enough information to be activated.
Our method is different from existing model selection methods
\cite{Duan2020AHF} and unsupervised metrics \cite{Karras2020ASG,VPdis_eccv20}
in ways like: 1) it does not rely on comparing
a herd of models;
2) it does not rely on training a classifier;
3) it approximately evaluates interpretability along axes.
More Pros and Cons are shown in the
Appendix Sec. \ref{ap:tpl_pros_cons}.

%-------------------------------------------------------------------------
\section{Experiments}
In this section we evaluate the proposed TPL model selection method,
and the effectiveness of our interpretability-oriented
models for learning disentangled representations.
The introductions of the used datasets and implementations are
in the Appendix Sec. \ref{ap:datasets} and
Sec. \ref{ap:implementations} respectively.
Code is available at \url{https://github.com/zhuxinqimac/PS-SC}.

\subsection{Effectiveness of TPL}
We first conduct experiments to evaluate the effectiveness of the proposed
TPL model selection method.
An intuitive way to do so is by examining its agreement
with existing supervised metrics
applied on existing disentanglement learning models.
%By using existing models, we can eliminate any biases introduced by our
%proposed constraints that may favor the TPL scores.
Specifically, we compute the TPL scores (the threshold $S$ is set to be 0.01
and the number of segments $N$ is set to be 50) on
1,800 pretrained checkpoints
from \cite{Locatello2018ChallengingCA}
on DSprites dataset and then compute the correlation
coefficients against four
supervised metrics: $\beta$-VAE metric (BVM) \cite{Higgins2017betaVAELB},
FactorVAE metric (FVM) \cite{Kim2018DisentanglingBF},
DCI disentanglement score \cite{Eastwood2018AFF},
and Mutual Information Gap (MIG) \cite{chen2018isolating}.
The pretrained checkpoints include 6 different models
($\beta$-VAE \cite{Higgins2017betaVAELB},
FactorVAE \cite{Kim2018DisentanglingBF},
DIP-VAE I and II \cite{Kumar2017VariationalIO},
$\beta$-TC-VAE \cite{chen2018isolating}, and
Annealed VAE \cite{Burgess2018UnderstandingDI}), covering
6 hyper-parameter configurations each and 50 random seeds each.
These configurations form an extensive
coverage from good models to bad models.

\begin{table}
    \small
    \begin{center}
        \begin{tabular}{llcccc}
            \toprule
            %\hline
            %\hline
            Range & Methods & BVM & DCI & FVM & MIG \\
            \midrule
            %\hline
            \multirow{3}{*}{All} & TPL (act$>$0) & 0.15 & 0.44 & 0.21 & 0.49 \\
            %& TPL (act$>$3) & 0.43 & 0.70 & 0.56 & 0.65 \\
            & TPL (act$>$4) & \textbf{0.45} & \textbf{0.72} & \textbf{0.60} & \textbf{0.66} \\
            & FVM & 0.82 & 0.77 & 1.00 & 0.72 \\
            %\midrule
            %\multirow{2}{*}{DIP-VAE-I} & UDR & \textbf{0.69} & \textbf{0.50} & \textbf{0.50} & \textbf{0.65} \\
            %%& UDR-All & \textbf{0.76} & 0.33 & \textbf{0.61} & \textbf{0.72} \\
            %& TPL (act$>0$) & 0.25 & 0.28 & 0.03 & 0.18 \\
            \midrule
            \multirow{2}{*}{TC-VAE} & UDR & \textbf{0.42} & 0.55 & 0.30 & 0.37 \\
            %& UDR-All & \textbf{0.59} & 0.47 & \textbf{0.51} & 0.60 \\
            & TPL (act$>$0) & 0.39 & \textbf{0.79} & \textbf{0.39} & \textbf{0.73} \\
            \bottomrule
            %\hline
            %\hline
        \end{tabular}
    \end{center}
    \caption{Spearman's rank correlation between unsupervised
    model selection methods and supervised disentanglement metrics.}
    \vspace{-10pt}
    \label{table:tpl_spearman}
\end{table}

\begin{figure}[t]
    \begin{center}
        \includegraphics[width=\linewidth]{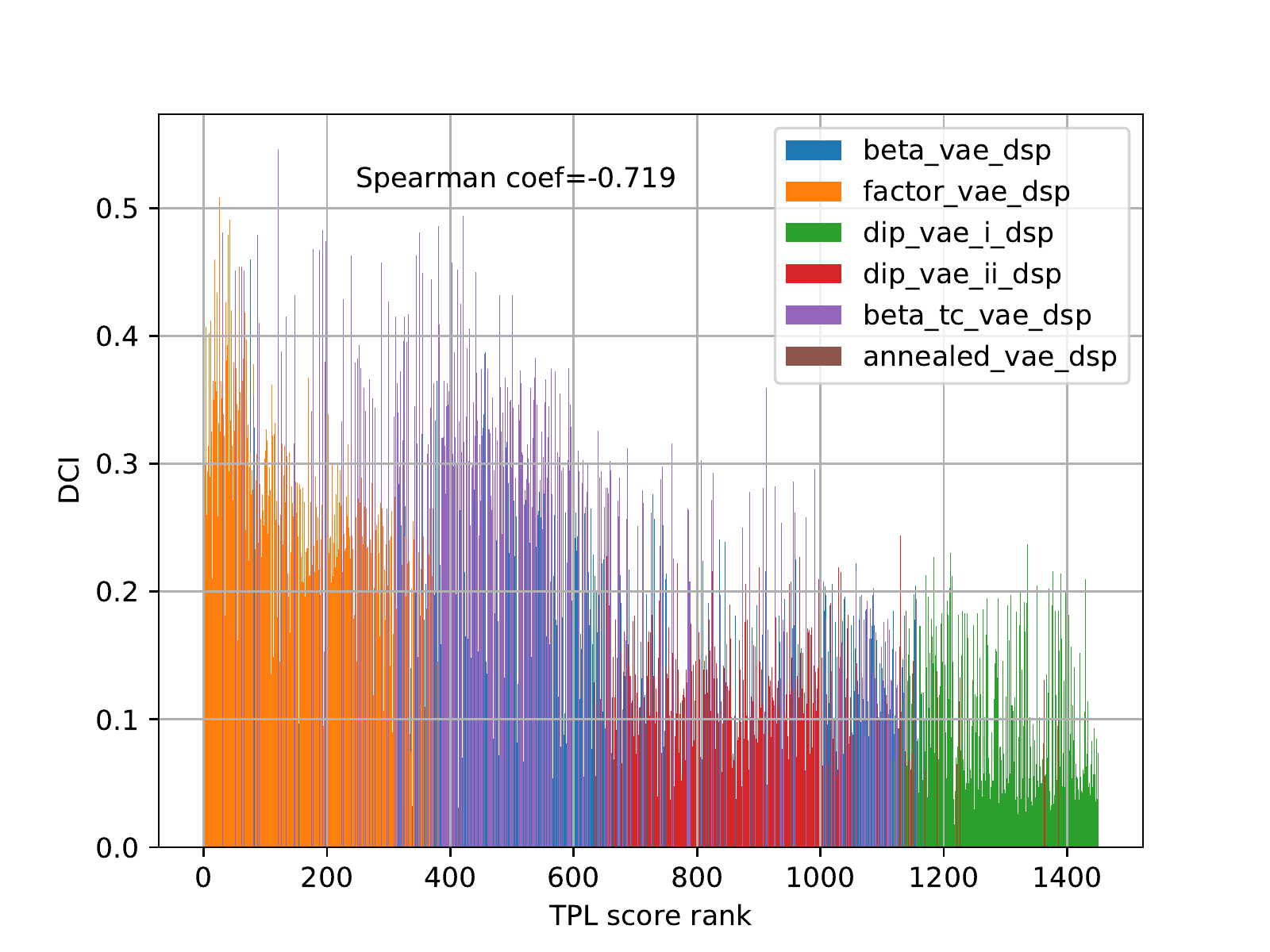}
    \end{center}
    \vspace{-10pt}
    \caption{TPL (act$>$4) vs DCI disentanglement
    on DSprites dataset across various configurations. Ranked by TPL scores.}
    \vspace{-10pt}
    \label{fig:tpl_vs_dci}
\end{figure}

In Table \ref{table:tpl_spearman} upper part we show the
Spearman's rank correlation scores computed over all models thresholded
by the number of active latent dimensions
(the active dimensions are determined
by TPL without supervision). The reason we threshold the models by
active latent dimensions is that the TPL can be misled by \emph{cheating}
models which achieve disentanglement by only encoding
a subset of generative factors.
These models \emph{are} indeed disentangled if they are evaluated based only
on this subset of factors, but will not be ranked high if compared
against all the ground-truth factors as done by supervised metrics.
However, our TPL is an unsupervised method and has no access to the
ground-truth factors, thus may wrongly rank those cheating models high,
leading to lower correlation with supervised metrics
as shown by the entry act$>$0 in Table \ref{table:tpl_spearman}
(there are 5 ground-truth factors).
Fortunately these models can be directly filtered out by the number of
active dimensions computed by TPL,
and in real-world applications they can also be filtered
out by unsupervised generative quality metrics like FID
\cite{Heusel2017GANsTB}, ensuring TPL to work in its more effective zone.
The TPL (act$>$4) works reasonably well for ranking
the pretrained models.
The supervised FVM row works as an upper bound, and the
largest difference between TPL and FVM is the correlation with BVM,
indicating the TPL is more similar to DCI and MIG metrics while
different from BVM.
In the lower part of Table \ref{table:tpl_spearman} we compare our model
against UDR unsupervised metric on the model TC-VAE
(UDR scores calculated on all models are not available).
Our method correlates better with supervised metrics,
especially DCI and MIG.
The plot of TPL (act$>$4) vs DCI metric is shown in Fig.
\ref{fig:tpl_vs_dci}, where the models are ranked by TPL.
An obvious descending trend can be observed, indicating the TPL
roughly sorts the models in a correct way.
As we plot different models with different colors, we see that
among the 6 models,
FactorVAE and TC-VAE are most promising for disentanglement
learning, which agrees with our common sense in this field.
More correlation results and more plots for other metrics and
other numbers of active dimensions are shown in the Appendix Sec.
\ref{ap:tpl_other_experiments}.
%the TPL is not aware
%of the true number of generative factors in dataset, and thus may
%assign a high rank to representations which ignore some variations
%in data. This causes the
%This is a \emph{cheating} strategy for a model to achieve
%high disentanglement with low generative capability, and is commonly
%observed in these models.
This experiment also implies that our hypothesis of Perceptual Simplicity
holds in general disentangled representations learned by existing models.

\subsection{Shoes and Clevr}
\begin{table}
    \small
    \begin{center}
        \begin{tabular}{lcccccc}
            \toprule
            %\hline
            %\hline
            \multirow{2}{*}{Methods} & \multicolumn{2}{c}{Shoes+Edges} &
                \multicolumn{2}{c}{Clevr-Simple} & \multicolumn{2}{c}{Clevr-Comp} \\
                %\multicolumn{2}{c}{Clevr-U} & \multicolumn{2}{c}{Clevr-1FOV} \\
            & PPL & FID & PPL & FID & PPL & FID \\
            \midrule
            %\hline
            InfoGAN & 2952.2 & 10.4 & 56.2 & \textbf{2.9} & 83.9 & \textbf{4.2} \\
            HP & 1301.3 & 21.2 & 45.7 & 25.0 & 73.1 & 21.1 \\
            HP+FT & 554.1 & 17.3 & 39.7 & 6.1 & 74.7 & 7.1 \\
            %\hline
            \midrule
            Ours & \textbf{246.2} & \textbf{9.7} & \textbf{9.2} & 9.2 & \textbf{17.4} & 10.8 \\
            \bottomrule
            %\hline
            %\hline
        \end{tabular}
    \end{center}
    \caption{Comparing Perceptual Path Length (PPL) and Fr\'echet
    Inception Distance (FID) on Edge+Shoes and Clevr datasets. Lower is better
    for both metrics.}
    \label{table:exp_shoes_clevr}
\end{table}
\begin{figure}[t]
    \begin{center}
        \includegraphics[width=\linewidth]{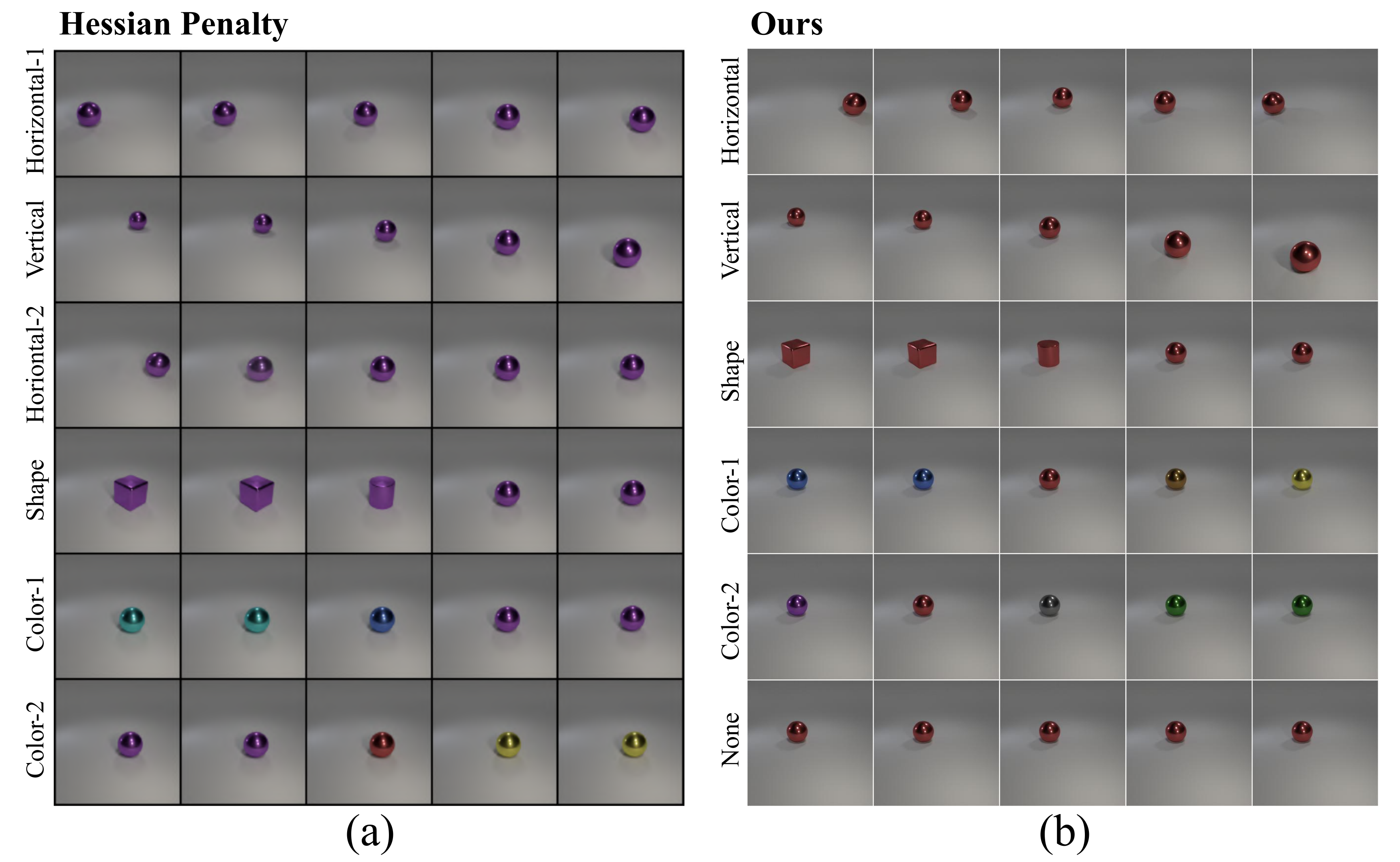}
    \end{center}
    \vspace{-10pt}
    \caption{Qualitative comparison between Hessian Penalty model
    \cite{peebles2020hessian} and ours on Clevr-Simple dataset.}
    \vspace{-10pt}
    \label{fig:clevr_compare}
\end{figure}
We follow the setups in \cite{peebles2020hessian} to conduct experiments
on the Shoes+Edges dataset (created by mixing 50,000 edges and 50,000 shoes)
\cite{fine-grained}, and variants of Clevr dataset \cite{8099698}:
Clevr-Simple contains four factors of variation: object color,
shape, and location (10,000 images);
Clevr-Complex contains two objects of Clevr-Simple
in multiple sizes (10,000 images).
Table \ref{table:exp_shoes_clevr} shows the quantitative comparison
between our model and multiple baselines provided by
\cite{peebles2020hessian}, using the same metrics
Perceptual Path Length (PPL) \cite{Karras2020ASG}
and Fr\'echet Inception Distance (FID) \cite{Heusel2017GANsTB}.
It is clear our model outperforms the baselines significantly.
Note that HP+FT is a fine-tuned model based on a pretrained
ProGAN \cite{karras2018progressive}, thus the direct end-to-end
trained baseline should be the HP version.
Our models work best in terms of disentanglement on all datasets,
and on Shoes+Edges ours can even achieve the best FID score.
On Clevr datasets, our models have worse FID, which may be caused
by the smaller size of the datasets, which consist of more factors of
variations than Shoes+Edges but have fewer data samples to train.

\begin{figure}[t]
    \begin{center}
        \includegraphics[width=\linewidth]{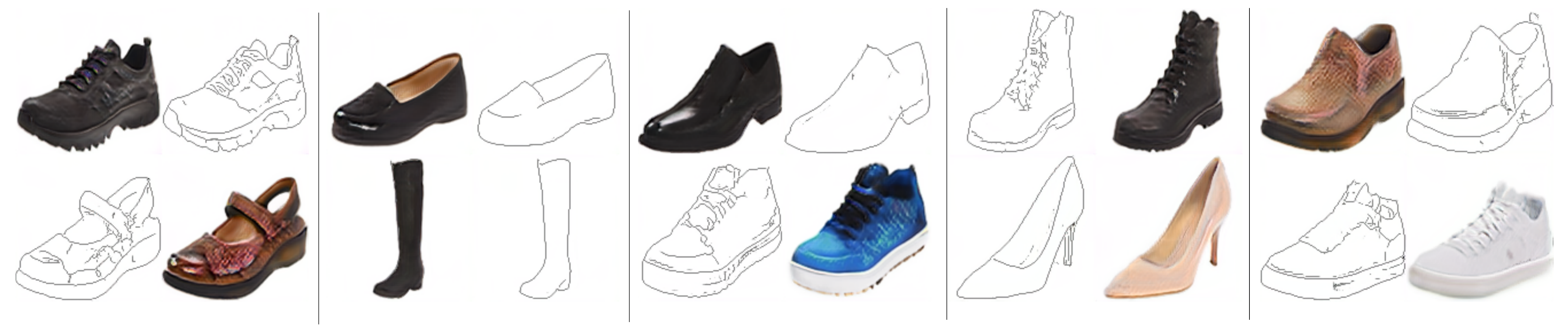}
    \end{center}
    \vspace{-10pt}
    \caption{Shoes $\leftrightarrow$ Edges translation by altering the
    dimension corresponding to the \emph{domain} concept.}
    %\vspace{-10pt}
    \label{fig:shoes_translation}
\end{figure}
In Fig. \ref{fig:clevr_compare} we qualitatively compare the
representations learned by a HP baseline and our model on
Clevr-Simple dataset. We show the latent traversals of
the learned latent codes
(baseline images are taken from \cite{peebles2020hessian}),
ordered from high to low
by our defined $\text{tpl}_i$ in Eq. \ref{eq:tpl_i}.
We can see our model encodes the vertical and horizontal position
variations into two clear separate latent codes, while the baseline
encodes these two factors into three codes.
In Fig. \ref{fig:shoes_translation} we show that our model
encode the domain concept in a single latent dimension on the
Shoes+Edges dataset. We achieve such a domain shift by just
reversing the sign of this single dimension in the representation.

\subsection{CelebA and FFHQ}
\label{sec:faces_exp}
We conduct experiments on human-face datasets CelebA \cite{Liu2014DeepLF}
and FFHQ \cite{Karras2020ASG}.
For CelebA we crop the center $128\times128$ area, and for FFHQ we use the
$512\times512$ version.
\begin{figure*}[t!]
    \begin{center}
        \includegraphics[width=\linewidth]{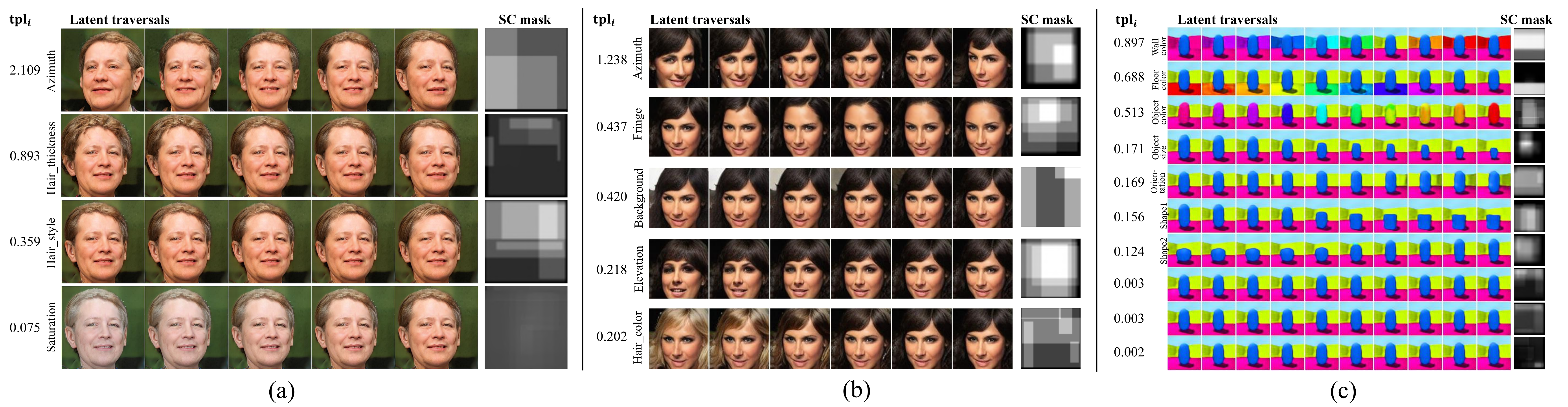}
    \end{center}
    \vspace{-10pt}
    \caption{(a) Latent traversals on FFHQ dataset
    ordered by $\text{tpl}_i$ scores.
    %SC masks are shown at the right end.
    %The $\text{tpl}_i$ scores agree with our common sense
    %about the significance of different semantics, and
    The masks coarsely highlight the corresponded components in images.
    (b) Same results on CelebA dataset. (c) Same results on 3DShapes dataset.}
    \vspace{-10pt}
    \label{fig:traversals_tpl}
\end{figure*}

\begin{figure}[t]
    \begin{center}
        \includegraphics[width=\linewidth]{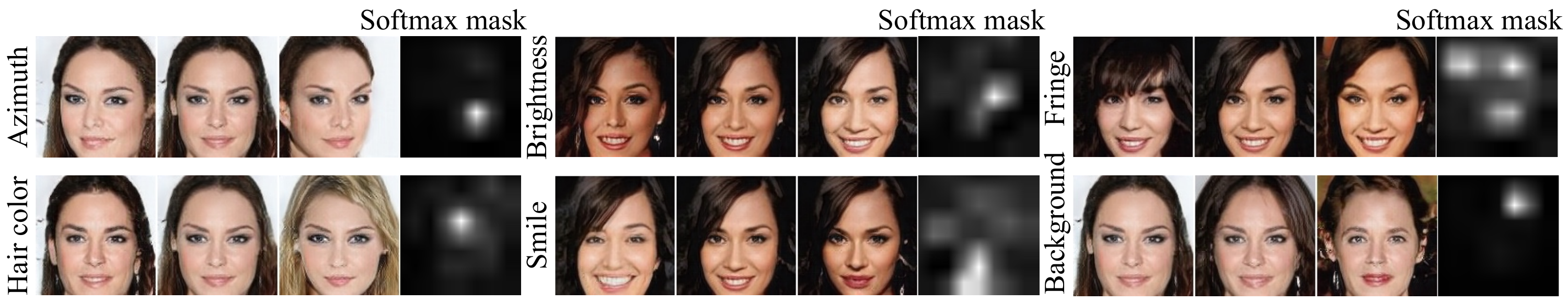}
    \end{center}
    \vspace{-10pt}
    \caption{Latent traversals by models trained with softmax masks.}
    \vspace{-5pt}
    \label{fig:softmax_masks}
\end{figure}

\begin{table}
    \small
    \begin{center}
        \begin{tabular}{lcccccc}
            \toprule
            %\hline
            %\hline
            \multirow{2}{*}{Methods} & \multicolumn{3}{c}{CelebA} & \multicolumn{3}{c}{FFHQ} \\
            & TPL & PPL & FID & TPL & PPL & FID \\
            \midrule
            %\hline
            %StyleGAN $\mathcal{W}$ & - & - & - & - & 195.9 & \textbf{4.4} \\
            InfoGAN & 10.9 & 43.6 & 6.0 & 33.4 & 142.7 & \textbf{11.0} \\
            +PS Loss & 8.5 & \textbf{34.3} & 6.2 & 30.7 & 139.6 & 13.8 \\
            +SC Module & 9.9 & 44.1 & \textbf{5.9} & 20.4 & 136.5 & 16.4 \\
            +Both & \textbf{8.1} & 38.9 & 6.0 & \textbf{18.7} & \textbf{120.1} & 13.5 \\
            \bottomrule
            %\hline
            %\hline
        \end{tabular}
    \end{center}
    \vspace{-5pt}
    \caption{Ablation study about different modules on CelebA and FFHQ
    datasets.}
    \vspace{-5pt}
    \label{table:ablation_modules}
\end{table}

\begin{table}
    \small
    \begin{center}
        \begin{tabular}{lcccccc}
            \toprule
            %\hline
            %\hline
            \multirow{2}{*}{Methods} & \multicolumn{3}{c}{CelebA} & \multicolumn{3}{c}{FFHQ} \\
            & TPL & PPL & FID & TPL & PPL & FID \\
            \midrule
            %\hline
            softmax-mask & 14.8 & 49.3 & 18.4 & 28.3 & 145.4 & 57.1 \\
            $\lambda=$ 0.001 & 10.2 & 47.5 & \textbf{5.3} & 20.3 & 135.4 & 23.7 \\
            $\lambda=$ 0.01 & \textbf{8.1} & \textbf{38.9} & 6.0 & \textbf{18.7} & \textbf{120.1} & \textbf{13.5} \\
            $\lambda=$ 0.1 & 9.1 & 42.0 & 7.2 & 20.0 & 123.2 & 16.0 \\
            \bottomrule
            %\hline
            %\hline
        \end{tabular}
    \end{center}
    \vspace{-5pt}
    \caption{Ablation study about softmax mask and $\lambda$ on
    CelebA and FFHQ datasets.}
    \vspace{-15pt}
    \label{table:softmax_lambda}
\end{table}

For quantitative evaluation, the FID metric is used to evaluate
generative quality, and the
PPL and TPL (the threshold $S$ is set to be 0.3 on CelebA and 0.5 on FFHQ,
the number of segments $N$ is set to be 48)
are used for rough measurement of disentanglement.
On CelebA we train models for around 19 epochs, and on FFHQ
we train models for around 28 epochs where FID starts to
saturate. In Table \ref{table:ablation_modules} we evaluate the effectiveness
of the proposed Perceptual Simplicity loss and the Spatial Constriction
module. Both modules improve the baseline model
in terms of disentanglement, validated by both the PPL metric and
our TPL scores. Note that the PPL measures the perceptual smoothness
of the latent space but cannot detect if a latent axis captures
simple variations, while TPL can perform a rough evaluation on this
property of interpretability.
This is why the PPL and TPL disagree on the +PS and +Both rows,
where the additional SC module slightly sacrifices
the overall smoothness in the latent space
to achieve an alignment between simple variations and latent axes.
%We also list the StyleGAN baseline from \cite{Karras2020ASG}
%in the first row to show that modeling interpretability is
%much more effective than mapping
%latent variables to a second latent space to achieve disentanglement.
On both datasets, the two contributions can be combined to achieve
the best performance, indicating their complimentarity.
Then we evaluate how the strength of PS loss impact the
learning of models. We denote the trade-off between the GAN loss and
PS loss by a hyper-parameter $\lambda$
: $\mathcal{L}_{total} = \mathcal{L}_{GAN} + \lambda \mathcal{L}_{PS}$.
Table \ref{table:softmax_lambda} shows the ablation study.
We see setting $\lambda=0.01$ is generally a good choice for both
datasets, while when it increases to 0.1 both models appear to
degenerate in generation ability.
For CelebA, setting $\lambda=0.001$ is beneficial to the image synthesis,
but on FFHQ this harms the image quality. It indicates this $\lambda$
is too small for the model to maintain good generation ability on this
high-resolution dataset with less training samples,
probably due to that the effect of the
latent reconstruction loss as a regularization
has been wiped out.
The experiment of switching the SC masks to softmax masks
is also shown in the Table \ref{table:softmax_lambda}.
We see using softmax masks is not an ideal choice since it harms
both the disentanglement and the generation quality.
This is also qualitatively verified by comparing
the latent traversals shown in
Fig. \ref{fig:softmax_masks} and Fig. \ref{fig:traversals_tpl} (b).
The ablation study on the number of rectangles $J$ used in SC modules
is shown in the Appendix Sec. \ref{ap:n_rectangles}.

\begin{figure}[t]
    \begin{center}
        \includegraphics[width=\linewidth]{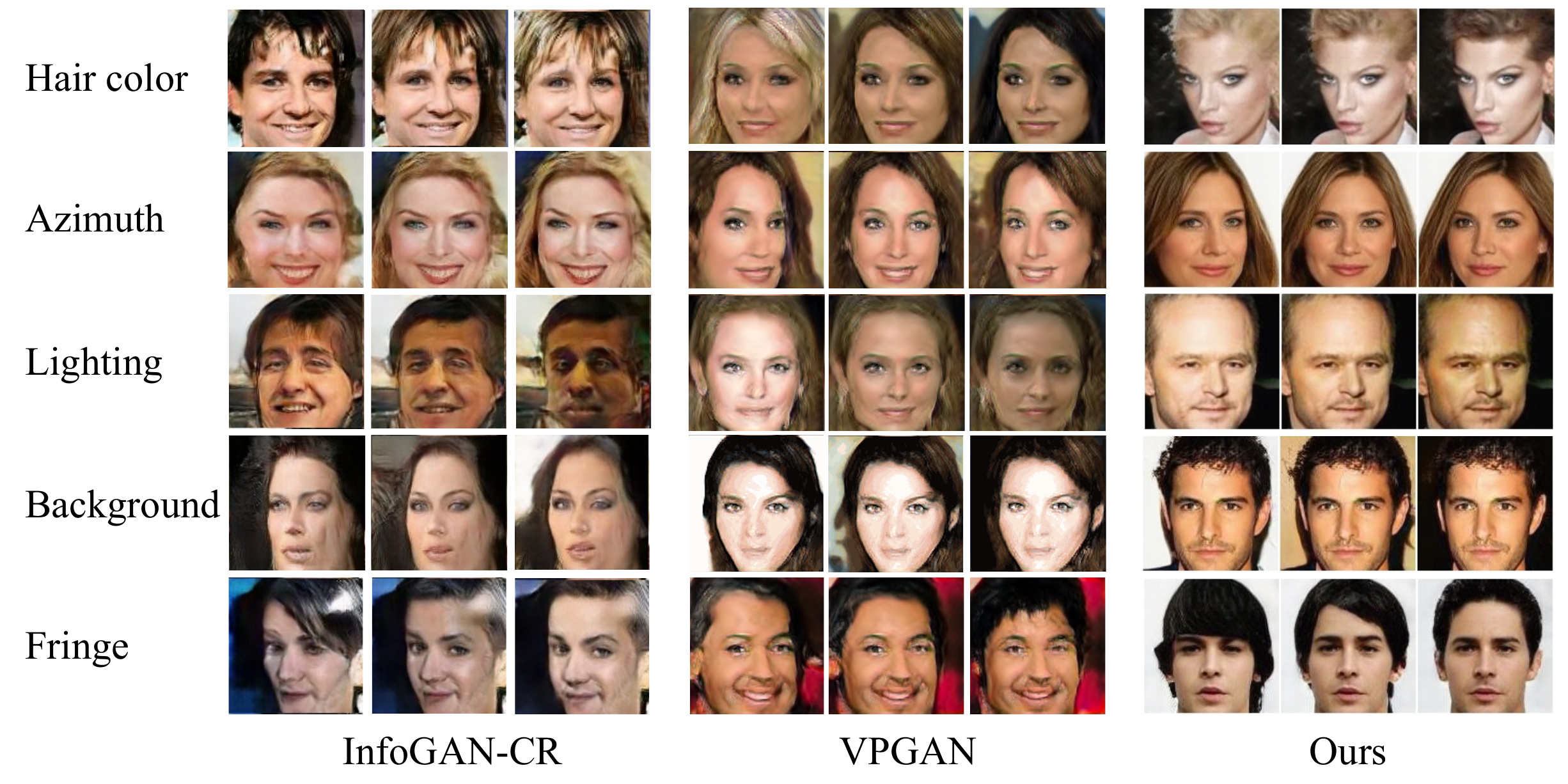}
    \end{center}
    \vspace{-10pt}
    \caption{Latent traversal comparison on CelebA dataset between InfoGAN-CR
    \cite{Lin2019InfoGANCRDG}, VPGAN \cite{VPdis_eccv20}, and our model.}
    %\vspace{-5pt}
    \label{fig:celeba_travs_comp}
\end{figure}

For qualitative evaluation, we show the latent traversals and
the learned SC masks on both datasets in
Fig. \ref{fig:traversals_tpl} (a) (b).
More traversals are shown
in the Appendix Sec. \ref{ap:more_qualitative_results}
and \texttt{traversals.gif} in the supplementary.
%for an impression, and refer \texttt{traversals.gif} in the
%supplementary material to view more learned semantics with animation).
We observe that 1) many semantics in these two datasets are
successfully captured by individual latent dimensions, including some
subtle variations like the hair thickness, hair style on FFHQ, and
the elevation on CelebA; 2) the dim-wise TPL score ($\text{tpl}_i$) agrees
with human's common sense about the significance of the discovered
semantics, indicating it can be used to filter out non-significant
noise information, or automatically detect important variations;
3) the SC masks coarsely align with the components controlled by
each latent code, \eg masks for azimuth and elevation covering main
areas in the images, masks for hair related information covering upper part
of the images, \etc. Compared to the masks learned by softmax shown in
Fig. \ref{fig:softmax_masks}, the SC masks are more informative
and interpretable, and the softmax masks usually consist of
point-like heatmaps. Though the softmax masks are
sometimes meaningful like the
ones corresponding to smile and fringe, masks for most other semantics
are not interpretable, resulting in worse
disentanglement quality than using SC masks.
Notice that in Fig. \ref{fig:traversals_tpl} (a) the dimension capturing
saturation is assigned with a very low
$\text{tpl}_i$ score. This is due to the bias of
perceptual distance used in the score computation,
which is more sensitive to high-level semantic variations.
In Fig. \ref{fig:celeba_travs_comp}, we compare the latent traversals
between our model and two existing GAN-based models on CelebA
($128 \times 128$ version). Though all three models seem to
discover the shown semantics,
the InfoGAN-CR has the worst disentanglement quality (e.g.
lighting entangled with smile).
Moreover, both baselines cannot maintain a high generative quality
when constrained to disentangle the underlying semantic factors.
%the InfoGAN-CR was trained with a
%latent code of exactly 5 dimensions to achieve this,
%which is a very strong constraint that forces each code to capture
%significant variations and is not generalizable to learn other
%less significant variations on datasets in higher resolution.
%On the other hand, the VPGAN was trained with over-parameterized latent
%dimensions like ours, but the generative quality
%has been significantly sacrificed
%to achieve this level of disentanglement.
Unlike either of them,
our model can achieve high-quality disentanglement while maintaining
much better generative quality.

\begin{table}
    \small
    \begin{center}
        \begin{tabular}{llll}
        \toprule
            \multicolumn{2}{c}{Model} & DSprites & 3DShapes \\
        \midrule
            \multirow{3}{*}{VAE} & VAE & 63 (6) & - \\
            & $\beta$-VAE & 74.41 (7.68) & 91 (from \cite{Kim2018DisentanglingBF}) \\
            %& CascadeVAE & \bf{90.49 (5.28)} & 93.51 (3.25) \\
            & CascadeVAE & 81.74 (2.97) & - \\
            & FactorVAE & 82.15 (0.88) & 89 (from \cite{Kim2018DisentanglingBF}) \\
        \midrule
            \multirow{5}{*}{GAN} & InfoGAN & 65.41 (7.03) & 83.65 (9.49) \\
            & IB-GAN & 80 (7) & - \\
            & InfoGAN-CR & \textbf{88 (1)} & - \\
            & Ours & 83.54 (6.91) & 92.04 (6.49) \\
            & Ours+TPL & 84.22 (4.21) & \textbf{93.41 (3.34)} \\
        \bottomrule
        \end{tabular}
    \end{center}
    \vspace{-5pt}
    \caption{State-of-the-art comparison using FactorVAE metric on DSprites
    and 3DShapes datasets.}
    \vspace{-10pt}
    \label{table:sota_dsp_3ds}
\end{table}

\begin{figure}[t]
    \begin{center}
        \includegraphics[width=\linewidth]{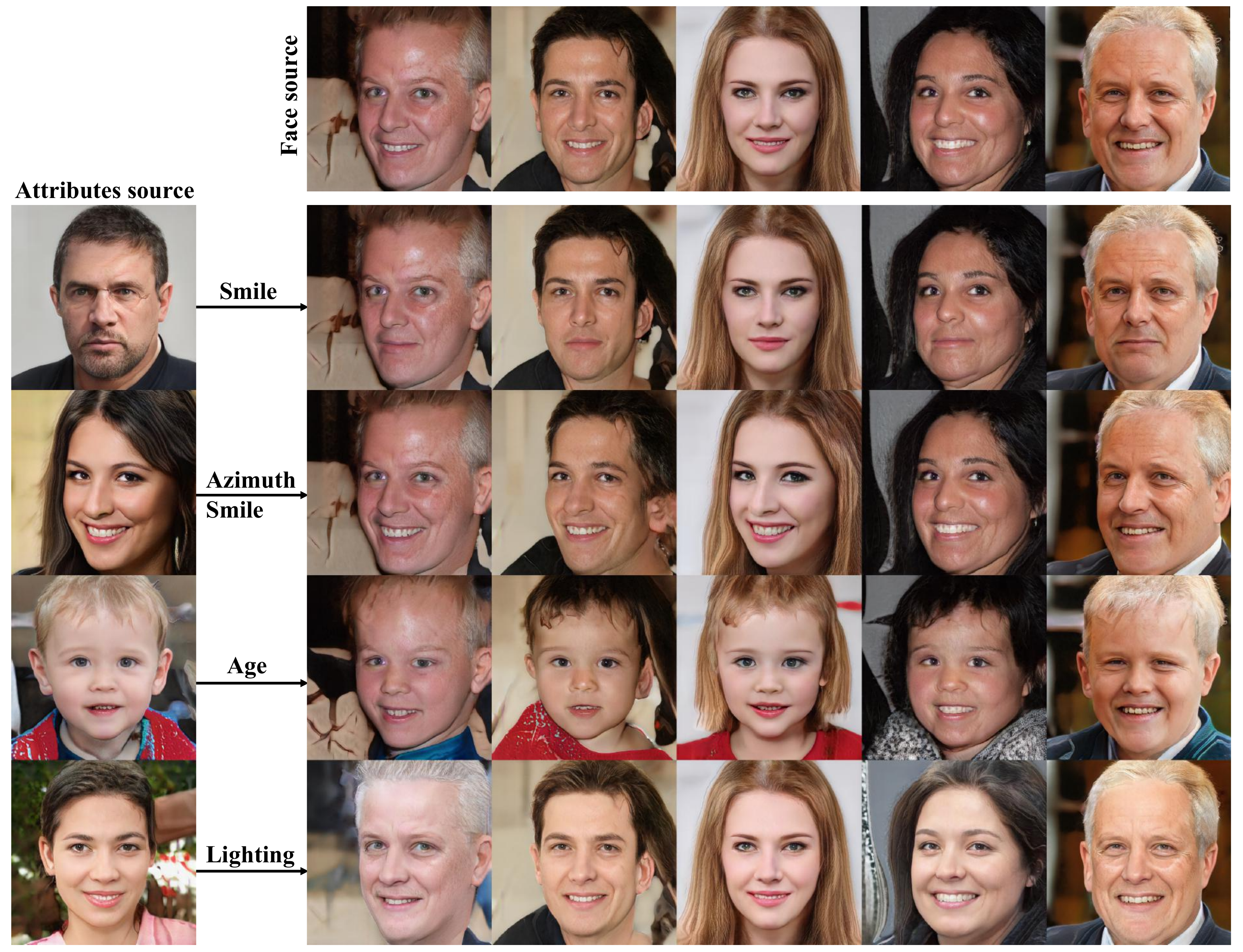}
    \end{center}
    \vspace{-10pt}
    \caption{
    %The image editing experiment conducted using our unsupervised model.
    Top row: source images. Left column: images providing attributes.
    Main section: transformed images using the
    provided attributes from the left ones.}
    \vspace{-10pt}
    \label{fig:image_editing}
\end{figure}

We then conduct an image editing experiment with our trained model on FFHQ.
Specifically, we generated a set of source images to provide main attributes,
and another set of images to provide the new attributes.
Afterwards we copy the latent code dimensions representing new attributes to
the corresponding positions in the latent code of the source images.
The results are shown in Fig. \ref{fig:image_editing}. We see
the source images are naturally adapted to the given attributes,
while keeping the other attributes intact.
More image editing examples are in the Appendix Sec. \ref{ap:more_editing}.
%Note that the StyleGAN uses a more advanced architecture and
%is trained for a much longer time to achieve a better FID score.
%\begin{table}
    %\begin{center}
        %\begin{tabular}{l|c|c|c}
            %\hline
            %\hline
            %Methods & VP Scores & TPL & FID \\
            %\hline
            %InfoGAN & 34.5 & - & 24.1 \\
            %FactorVAE & 75.0 & - & 73.9 \\
            %VPGAN-flat & 64.5 & - & 32.8 \\
            %VPGAN-hierar & 70.3 & - & 56.9 \\
            %\hline
            %InfoGAN-our-impl & - & 6.6 & \textbf{5.8} \\
            %Ours & - & \textbf{5.6} & 6.0 \\
            %\hline
            %\hline
        %\end{tabular}
    %\end{center}
    %\caption{Comparing Variation Predictability score (VP)on
    %CelebA-128 dataset. Higher is better for VP score.}
    %\label{table:exp_celeba_vpgan}
%\end{table}

\subsection{DSprites and 3DShapes}
In Table \ref{table:sota_dsp_3ds} we compare the state-of-the-art models
trained with continuous latent codes on DSprites and 3DShapes synthetic
datasets. 10 random seeds are used to train the models, and for the +TPL
version we use 30 seeds to train and report results with the
top 10 seeds ranked by our unsupervised TPL model selection method.
On DSprites the InfoGAN-CR achieves the best performance,
but it was trained by keeping the number of latent dimensions equal
to the ground-truth factors, which is different from the
general over-parameterized setting.
Our model shown on DSprites does not use SC module since the
Spatial Constriction assumption does not hold on this synthetic dataset
and using SC module harms the performance (obtaining a score of 81.47 (5.39)).
On 3DShapes our model achieves the best performance, and a latent traversal
is shown in Fig. \ref{fig:traversals_tpl} (c).

%-------------------------------------------------------------------------
\section{Conclusion}
Based on the observation that interpretability usually comes from
localized and simple variations, we proposed to learn disentangled
representations by directly modeling interpretability
from these two perspectives as a proxy.
We adopted two hypotheses, Spatial Constriction and
Perceptual Simplicity, to construct our models.
We designed a module to constrain the impact of each latent dimension
into constricted subareas, and a loss to enforce the encoding
of simple variations along latent axes.
We also introduced a simple unsupervised model selection method
by quantifying the perceptual variations accumulated along latent axes.
Experiments on various datasets validated the effectiveness of
our proposed modules and the model selection method.
Although our work was proposed as a standalone approach for learning
disentangled representations, it should work together with
other assumptions like statistical
independence to achieve a boosted performance, and we left this
exploration for future work.

\section*{Acknowledgement\vspace{-1.5ex}}
{\footnotesize This work is supported by Australian
Research Council under Projects FL-170100117, DP-180103424,
DE180101438 and DP210101859.
}

\clearpage

{\small
\bibliographystyle{ieee_fullname}
\bibliography{egbib}
}

\clearpage

\section*{Appendix}

\section{GAN and InfoGAN}
\label{ap:gan_infogan}
The Generative Adversarial Network \cite{Goodfellow2014GenerativeAN}
is a generative model trained via a minimax game performed between a
generator $G$ and a discriminator $D$:
\begin{align}
    \underset{G}{\text{min}}\, \underset{D}{\text{max}}\, V(G, D) &=
    \mathbb{E}_{\bsb{x} \sim p_{\text{data}}}[\text{log}\,
    D(\bsb{x})] \nonumber \\
    &+ \mathbb{E}_{\bsb{z} \sim p_{\text{noise}}}[\text{log}\,
    (1 - D(G(\bsb{z})))
    ] \label{eq:gan},
\end{align}
where $\bsb{z}$ is a noise variable sampled from a prior distribution
$p_{\text{noise}}(\bsb{z})$, the generator $G$
maps $\bsb{z}$ to the pixel space to
synthesize images, and the discriminator $D$ predictes
if an images is sampled from the real
distribution or not. After convergence, the $G$ should be able to synthesize
realistic images and the $D$ cannot tell if an image is fake or not.

The InfoGAN \cite{Chen2016InfoGANIR} augments the GAN loss (Eq. \ref{eq:gan})
with a regularization term:
\begin{align}
    \underset{G,Q}{\text{min}}\, \underset{D}{\text{max}}\, V_{\text{INFO}}(G, D) =
    V(G, D) - \beta I(\bsb{c}; G(\bsb{c}, \bsb{z})) \label{eq:infogan},
\end{align}
where $I(\bsb{c}; G(\bsb{c}, \bsb{z}))$ is the mutual information between
a subset of latent codes $\bsb{c} \in \mathbb{R}^{d}$
and the generated samples $G(\bsb{c}, \bsb{z})$.
By maximizing the mutual information, the latent codes $\bsb{c}$ are able to
represent a set of interpretable salient variations in data.
In practice, the mutual information is approximated by a variational lower
bound and is implemented with an auxiliary network $Q$, trained together with
$G$ by regression loss for predicting the latent codes $\bsb{c}$.

\section{Datasets Introduction}
\label{ap:datasets}
\paragraph{CelebA}
This is a dataset of 202,599 images of cropped real-world
human faces, containing various poses, backgrounds and
facial expressions. We use the cropped center
$128 \times 128$ area in this paper.
\paragraph{Shoes+Edges}
This dataset contains the commonly used image-to-image translation datasets
Shoes and Edges,
but mixes the 50,000 Shoes images and 50,000 Edges images together
to form a 100,000-image dataset in 128$\times$128.
\paragraph{Clevr-Simple}
This dataset is a variant of the Clevr dataset,
which contains an object featuring four factors of variation:
object color, shape, and location (both horizontal
and vertical). It contains 10,000 256$\times$256 images.
\paragraph{Clevr-Complex}
This dataset retains all variations from Clevr-Simple but adds a
second object and multiple sizes for a total of 10 factors of variation
(5 per object). It contains 10,000 256$\times$256 images.
\paragraph{FFHQ} This is a dataset of aligned images of human faces crawled
from Flickr (70,000 in tatal). We use the $512\times512$ version in our paper.
Link: https://github.com/NVlabs/ffhq-dataset.
\paragraph{DSprites} This is a dataset of 2D shapes generated from
5 independent factors, which are \emph{shape} (3 values),
\emph{scale} (6 values), \emph{orientation} (40 values),
\emph{x position} (32 values), and \emph{y position} (32 values).
All combinations are present exactly once, with the total number
of binary $64\times 64$ images 737,280.
Link: https://github.com/deepmind/dsprites-dataset.
\paragraph{3DShapes} This is a dataset of 3D shapes generated from
6 independent factors, which are \emph{floor color} (10 values),
\emph{wall color} (10 values), \emph{object color} (10 values),
\emph{scale} (8 values), \emph{shape} (4 values),
\emph{orientation} (15 values).
All combinations are present exactly once, with the total number
of $64\times 64$ images 480,000. Link: https://github.com/deepmind/3d-shapes.

\section{More Experiments of Rotating Latent Space}
\label{ap:tpl_explain_more}
\begin{figure}[t]
\begin{center}
   \includegraphics[width=\linewidth]{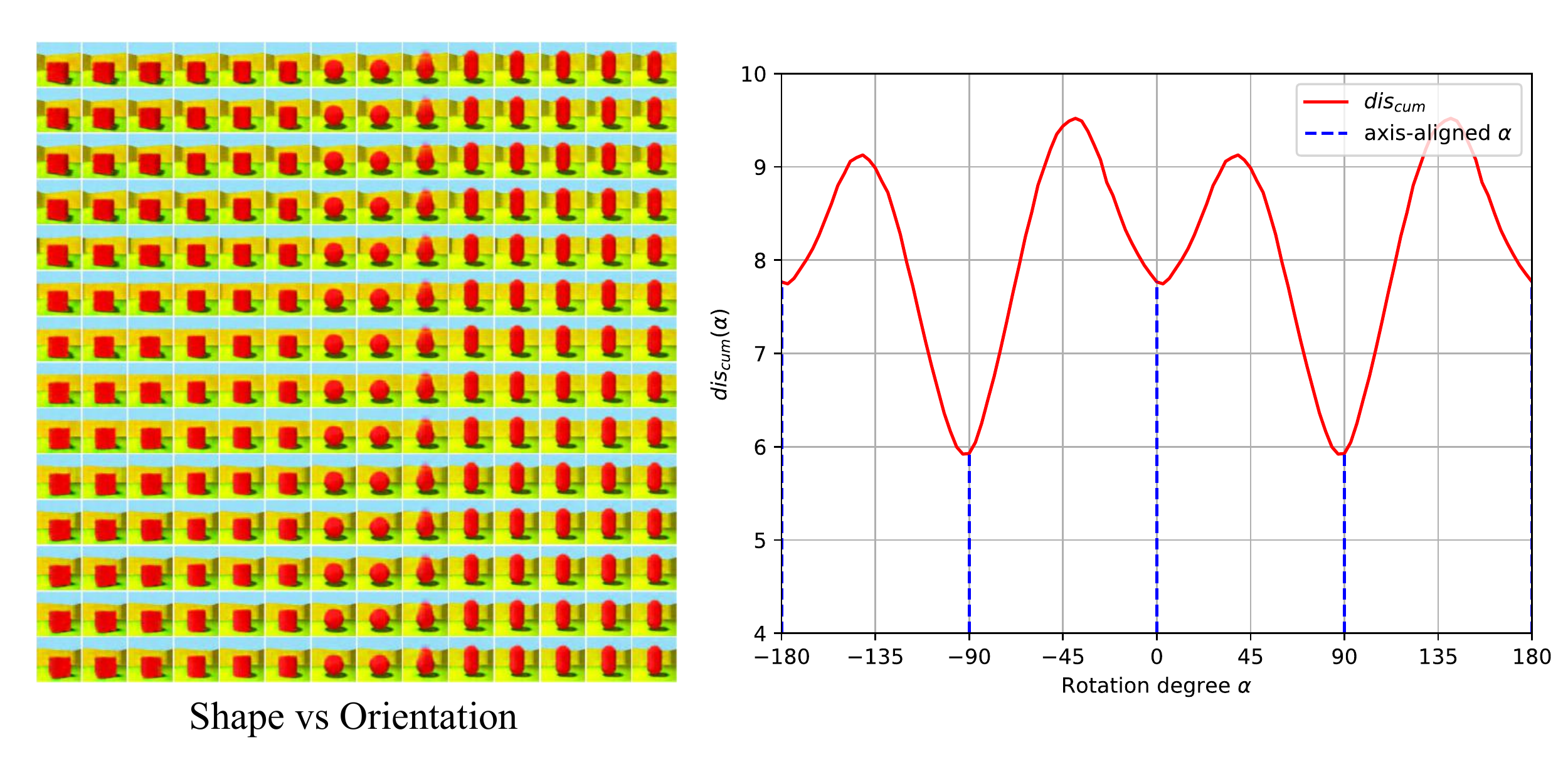}
\end{center}
    %\vspace{-8pt}
    \caption{Shape vs Orientation.}
    %\vspace{-8pt}
\label{fig:shape_vs_orientation}
\end{figure}
\begin{figure}[t]
\begin{center}
   \includegraphics[width=\linewidth]{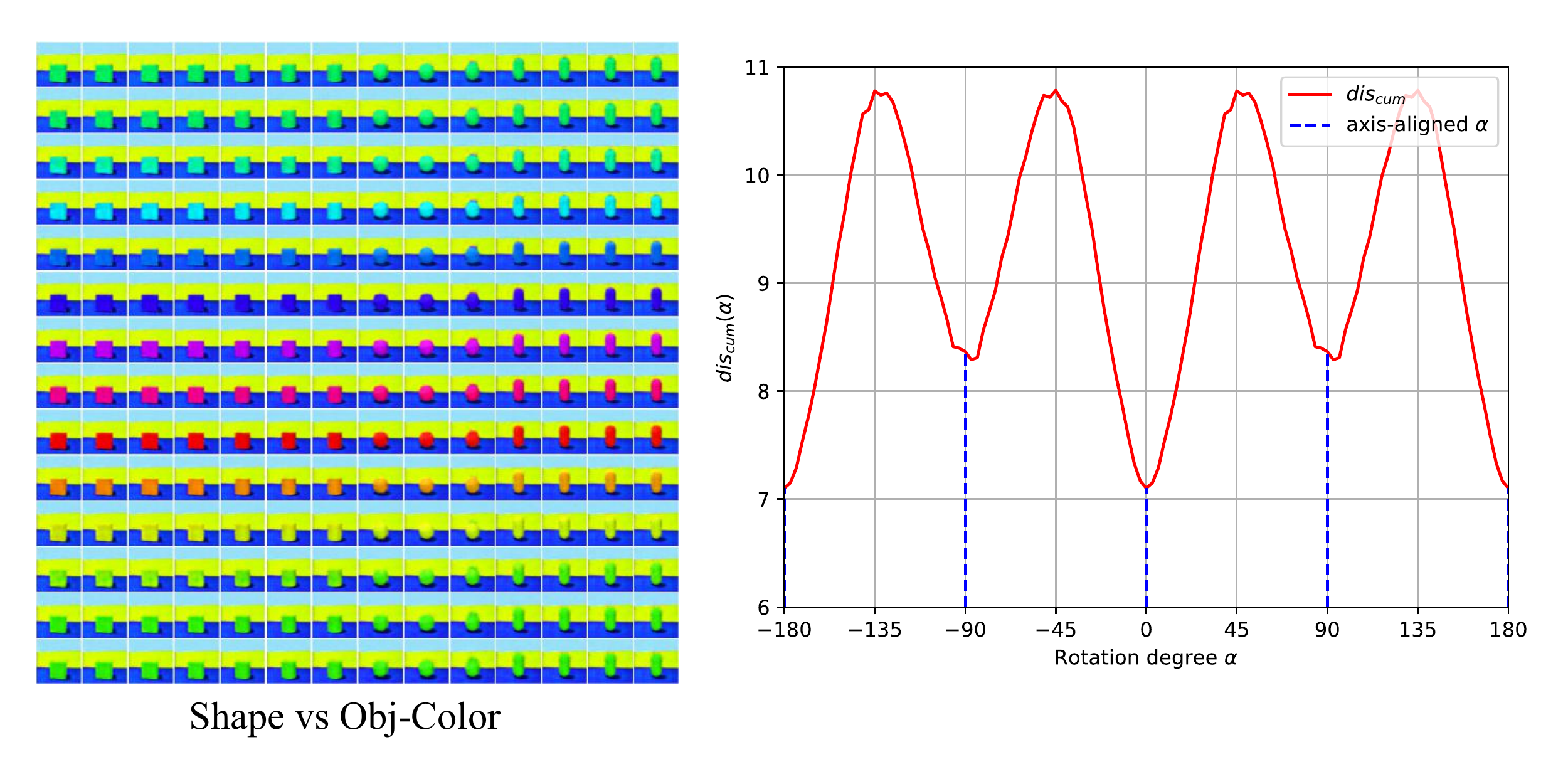}
\end{center}
    %\vspace{-8pt}
    \caption{Shape vs Obj-Color.}
    %\vspace{-8pt}
\label{fig:shape_vs_objcolor}
\end{figure}
\begin{figure}[t]
\begin{center}
   \includegraphics[width=\linewidth]{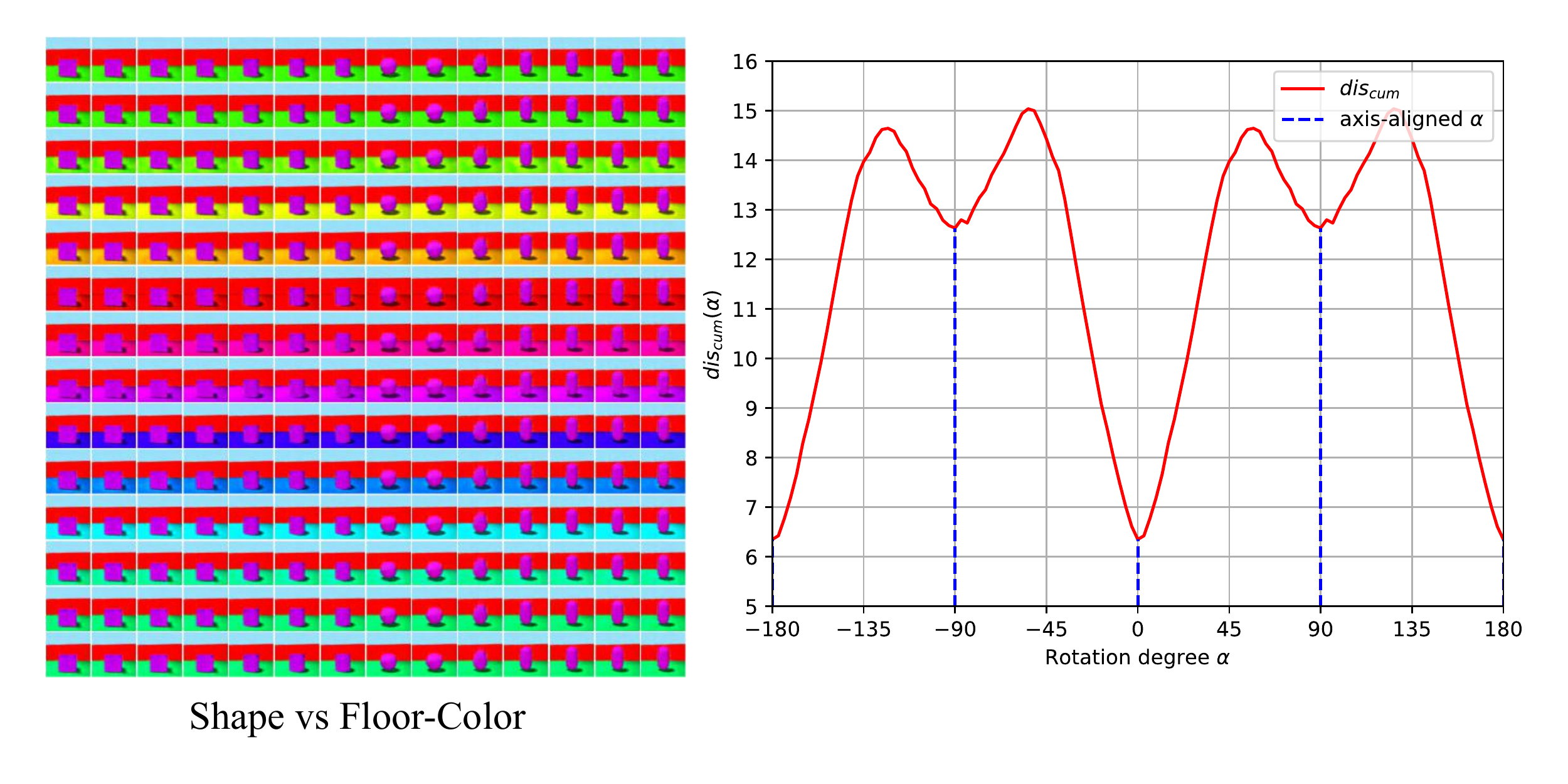}
\end{center}
    %\vspace{-8pt}
    \caption{Shape vs Floor-Color.}
    %\vspace{-8pt}
\label{fig:shape_vs_floorcolor}
\end{figure}
\begin{figure}[t]
\begin{center}
   \includegraphics[width=\linewidth]{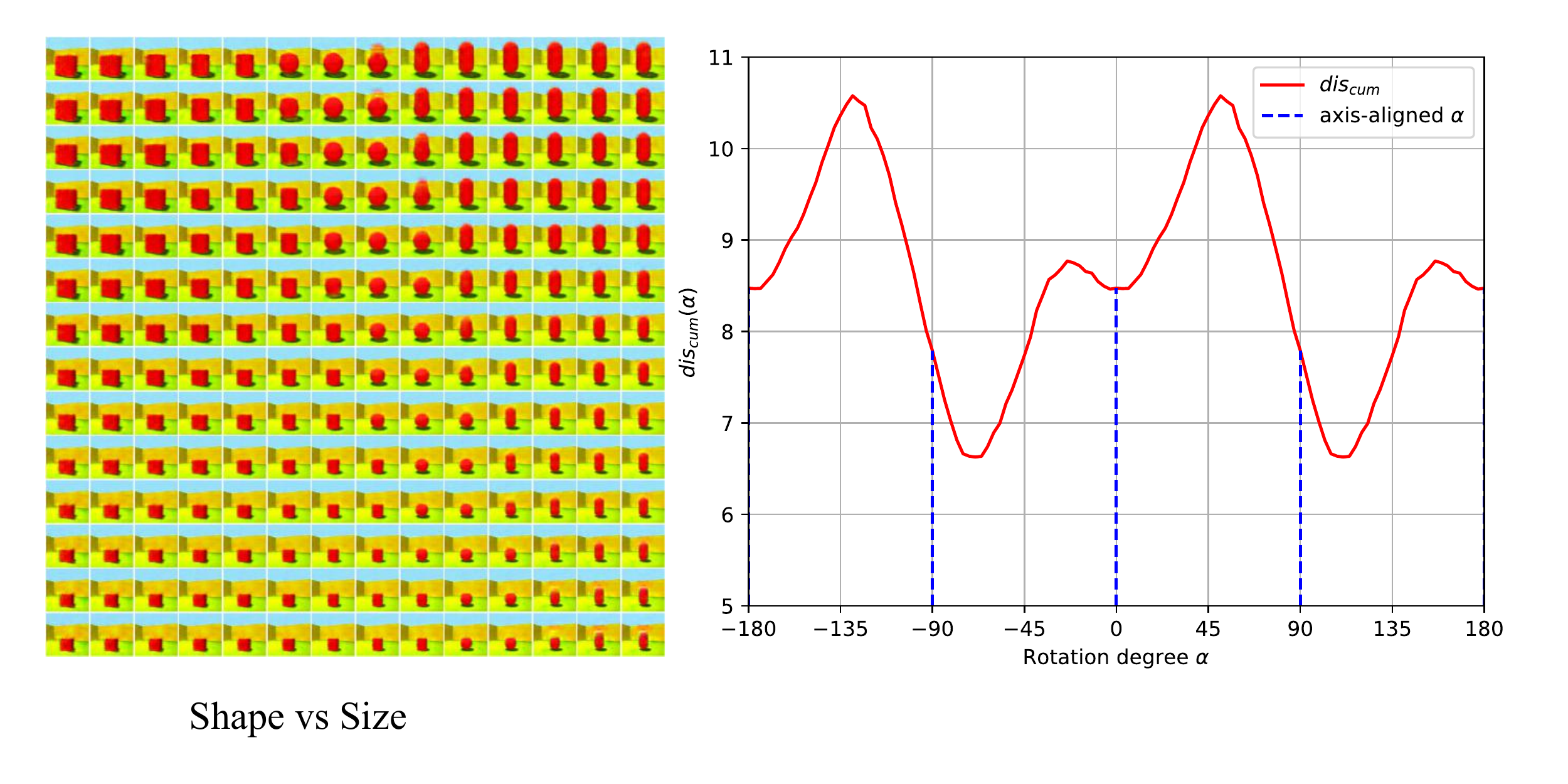}
\end{center}
    %\vspace{-8pt}
    \caption{Shape vs Size.}
    %\vspace{-8pt}
\label{fig:shape_vs_size}
\end{figure}
\begin{figure}[t]
\begin{center}
   \includegraphics[width=\linewidth]{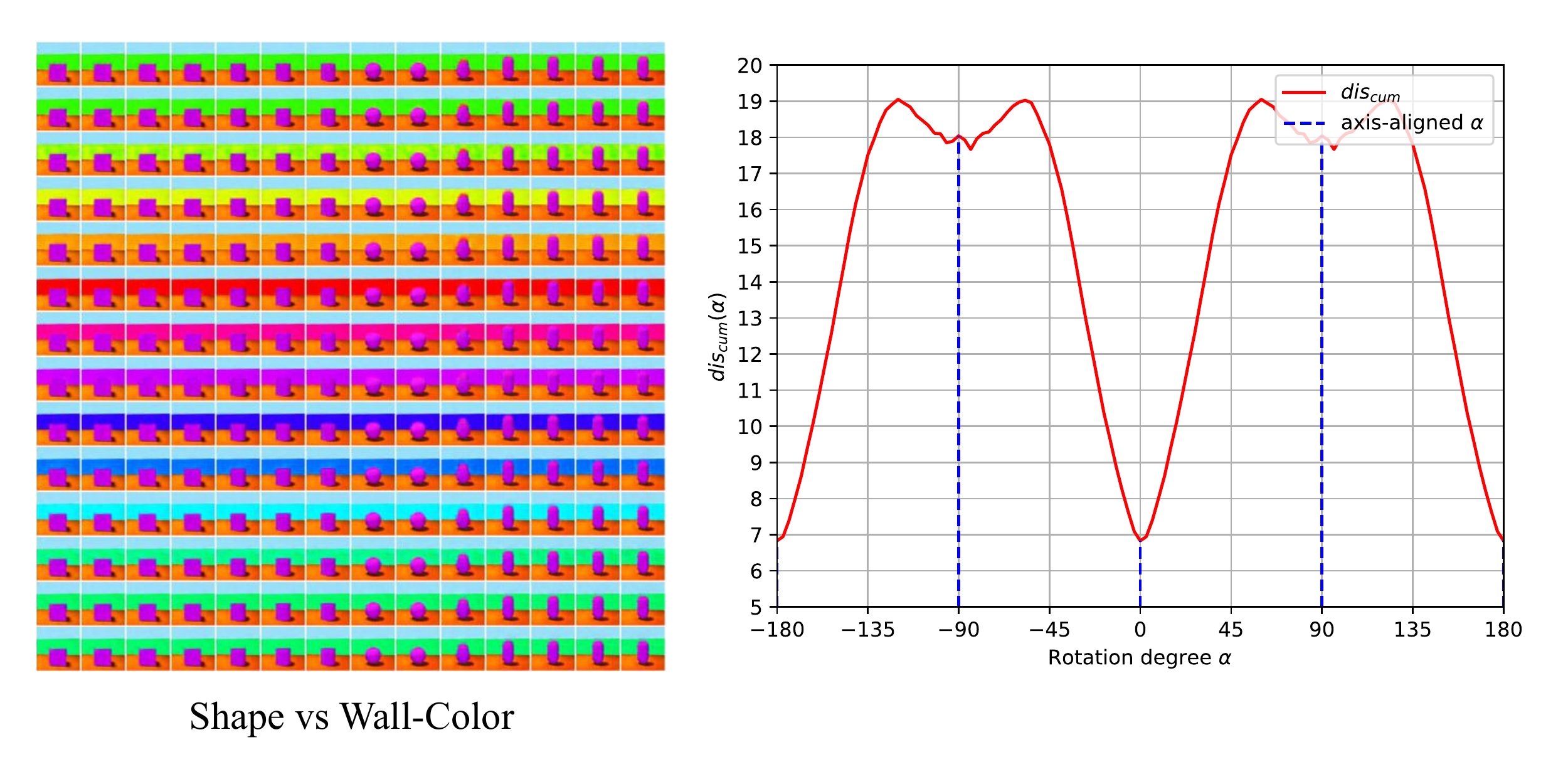}
\end{center}
    %\vspace{-8pt}
    \caption{Shape vs Wall-Color.}
    %\vspace{-8pt}
\label{fig:shape_vs_wallcolor}
\end{figure}
\begin{figure}[t]
\begin{center}
   \includegraphics[width=\linewidth]{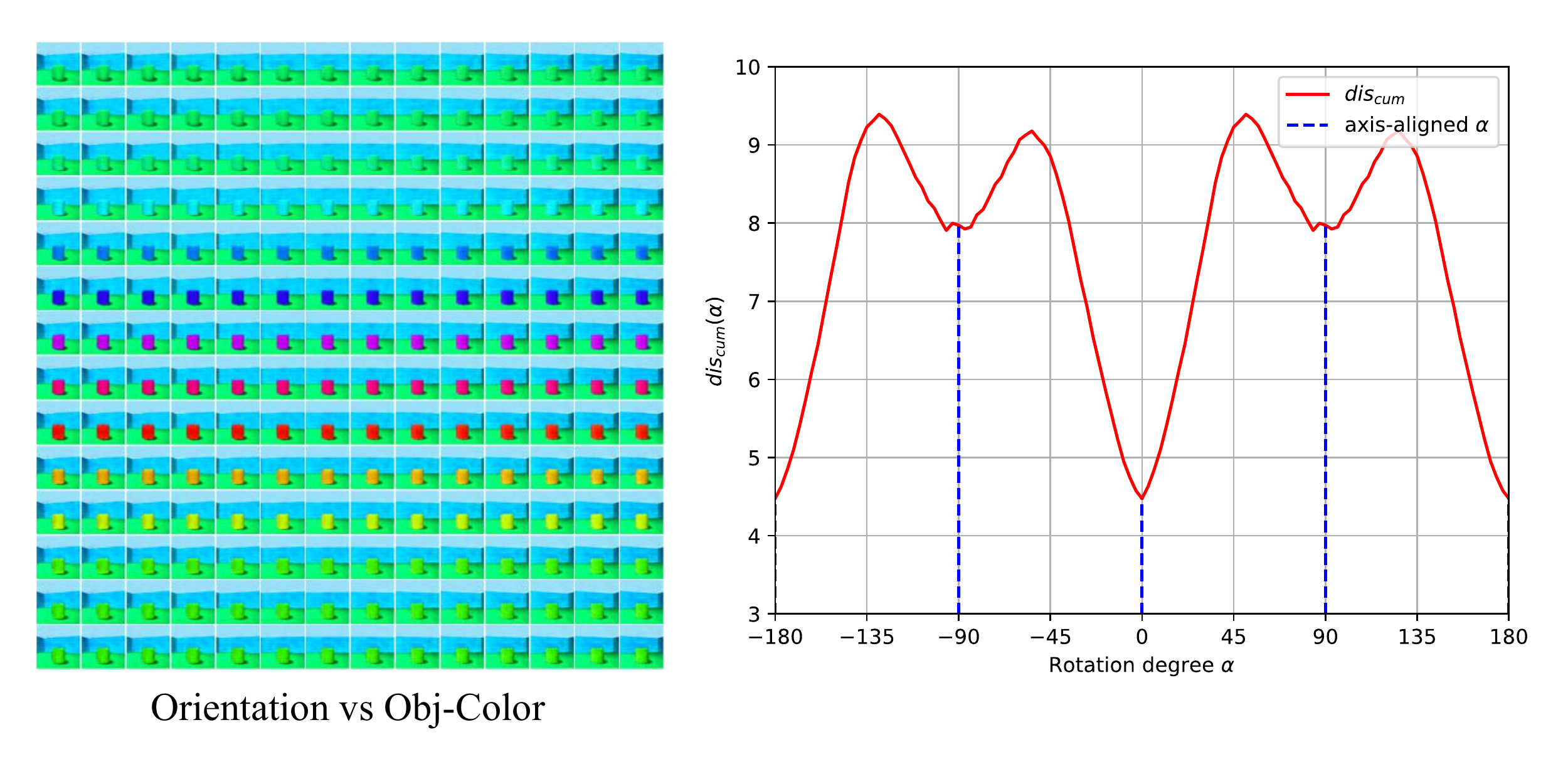}
\end{center}
    %\vspace{-8pt}
    \caption{Orientation vs Obj-Color.}
    %\vspace{-8pt}
\label{fig:orientation_vs_objcolor}
\end{figure}
\begin{figure}[t]
\begin{center}
   \includegraphics[width=\linewidth]{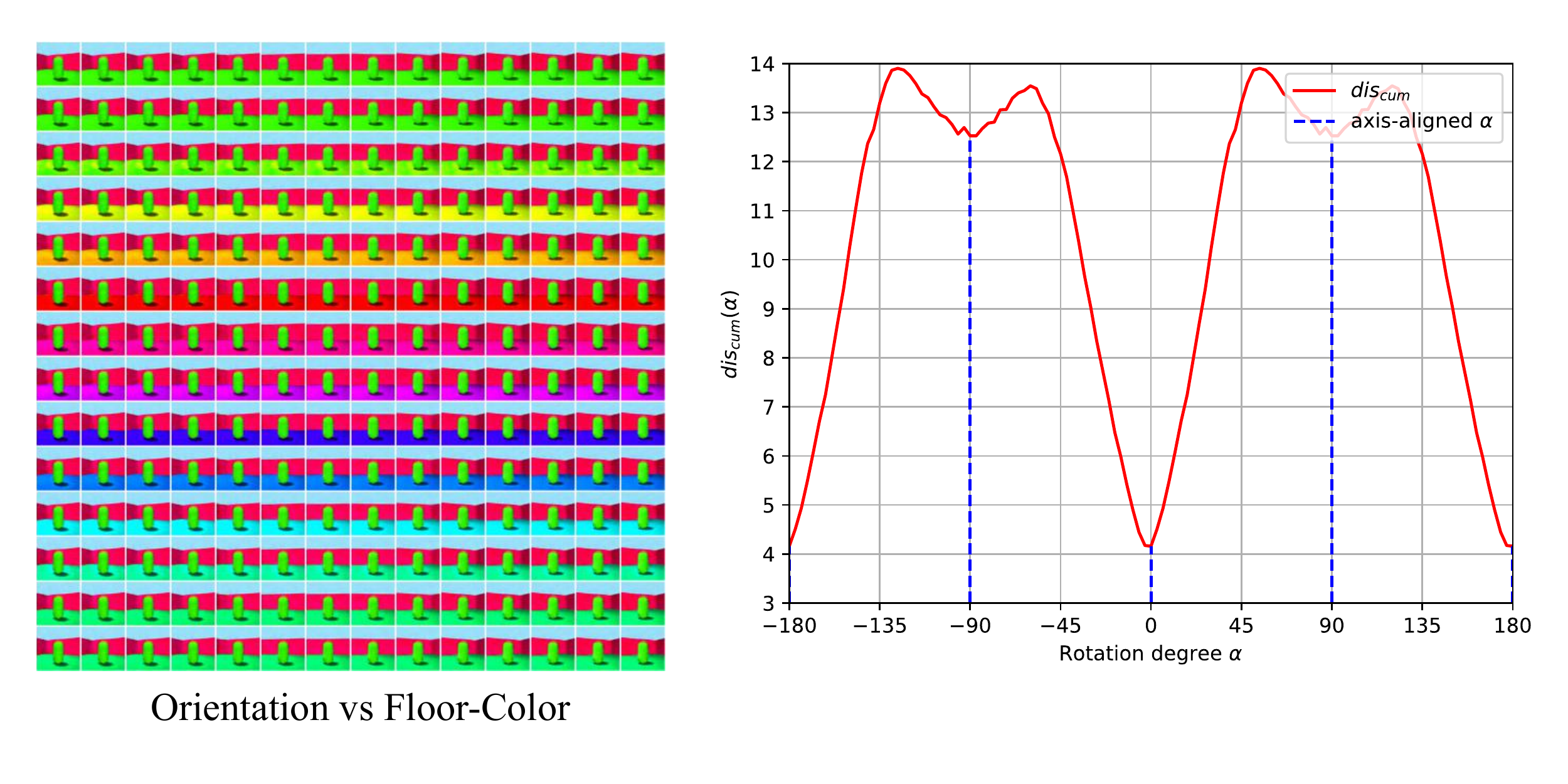}
\end{center}
    %\vspace{-8pt}
    \caption{Orientation vs Floor-Color.}
    %\vspace{-8pt}
\label{fig:orientation_vs_floorcolor}
\end{figure}
\begin{figure}[t]
\begin{center}
   \includegraphics[width=\linewidth]{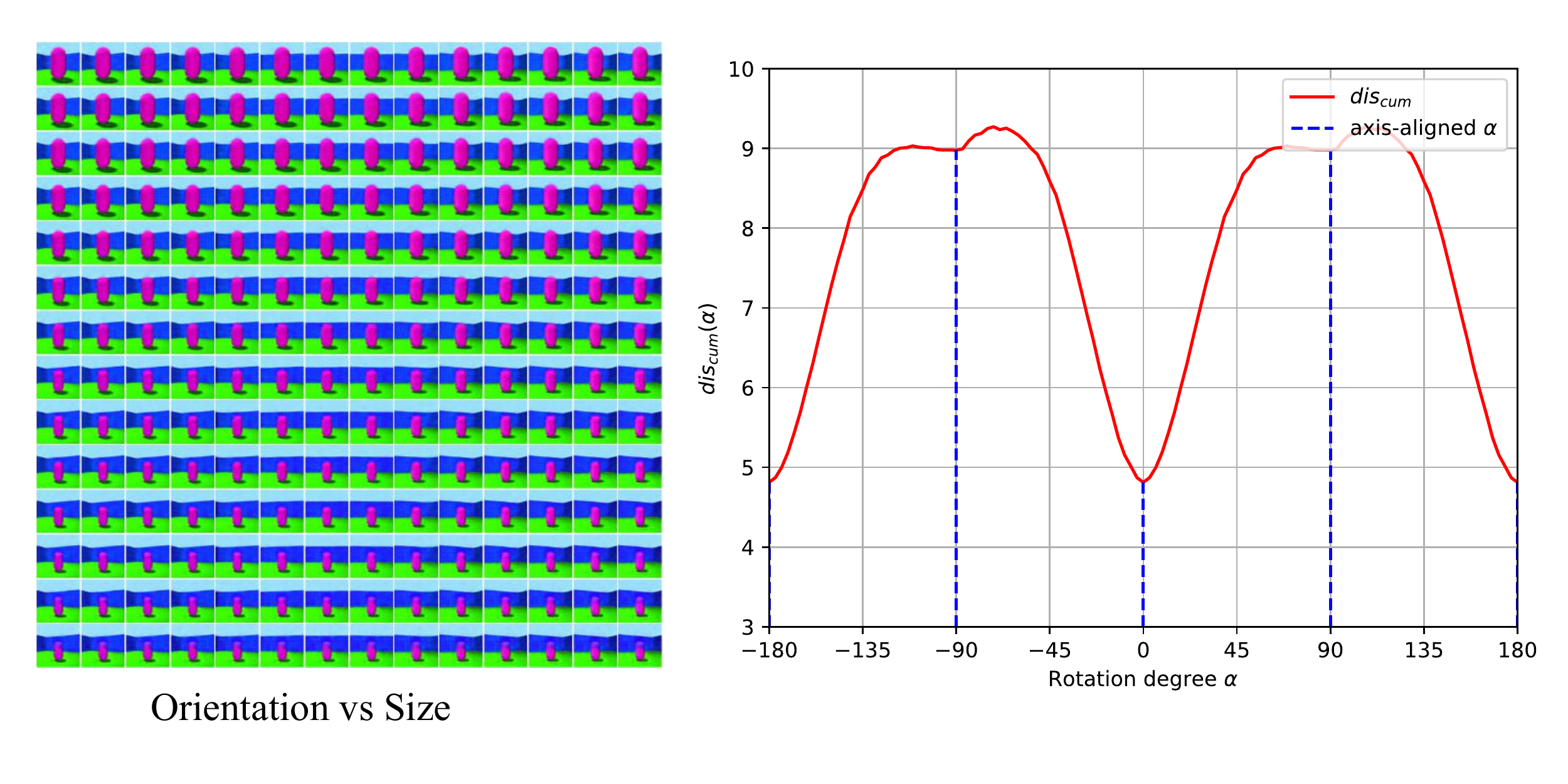}
\end{center}
    %\vspace{-8pt}
    \caption{Orientation vs Size.}
    %\vspace{-8pt}
\label{fig:orientation_vs_size}
\end{figure}
\begin{figure}[t]
\begin{center}
   \includegraphics[width=\linewidth]{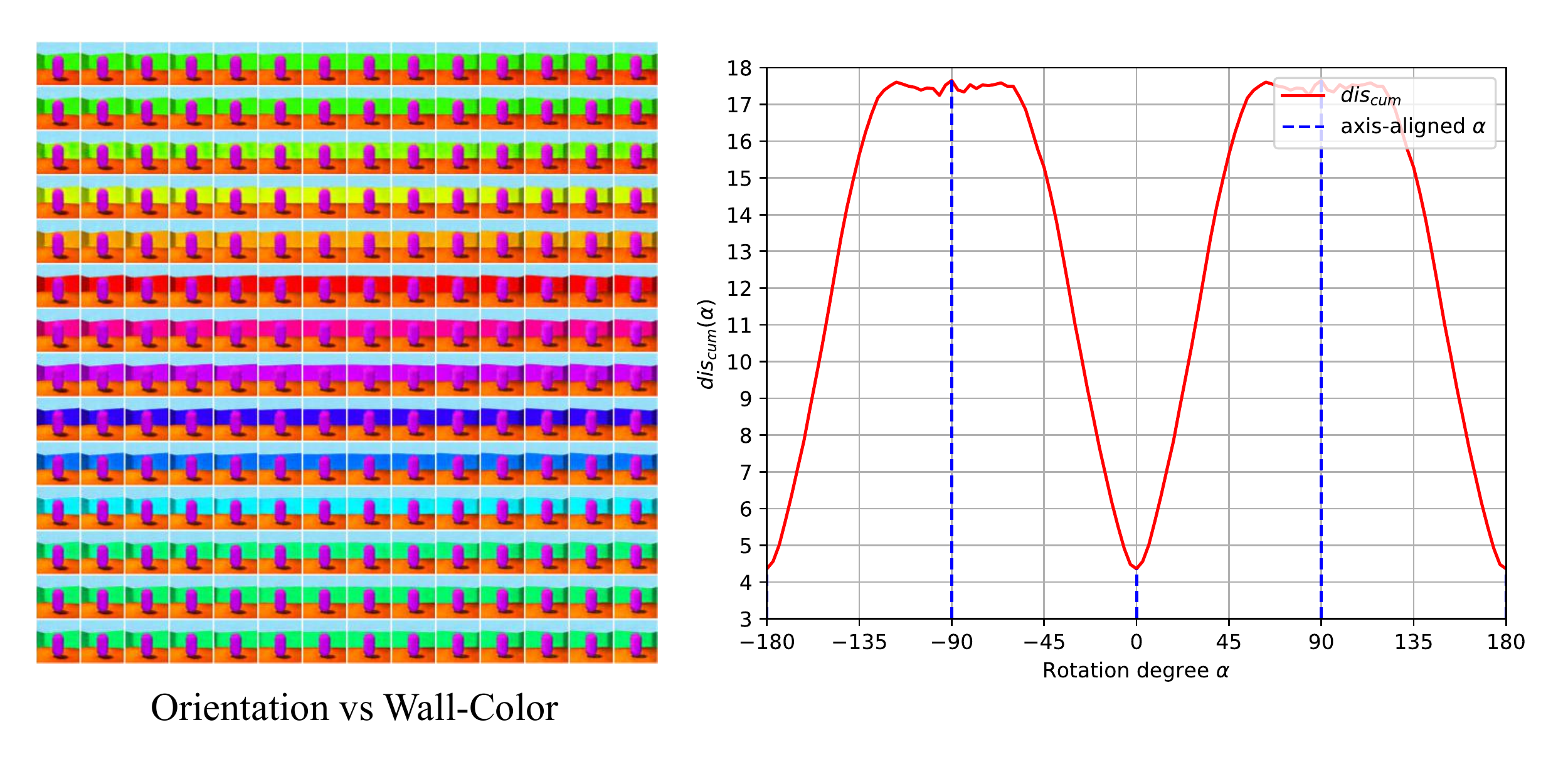}
\end{center}
    %\vspace{-8pt}
    \caption{Orientation vs Wall-Color.}
    %\vspace{-8pt}
\label{fig:orientation_vs_wallcolor}
\end{figure}
\begin{figure}[t]
\begin{center}
   \includegraphics[width=\linewidth]{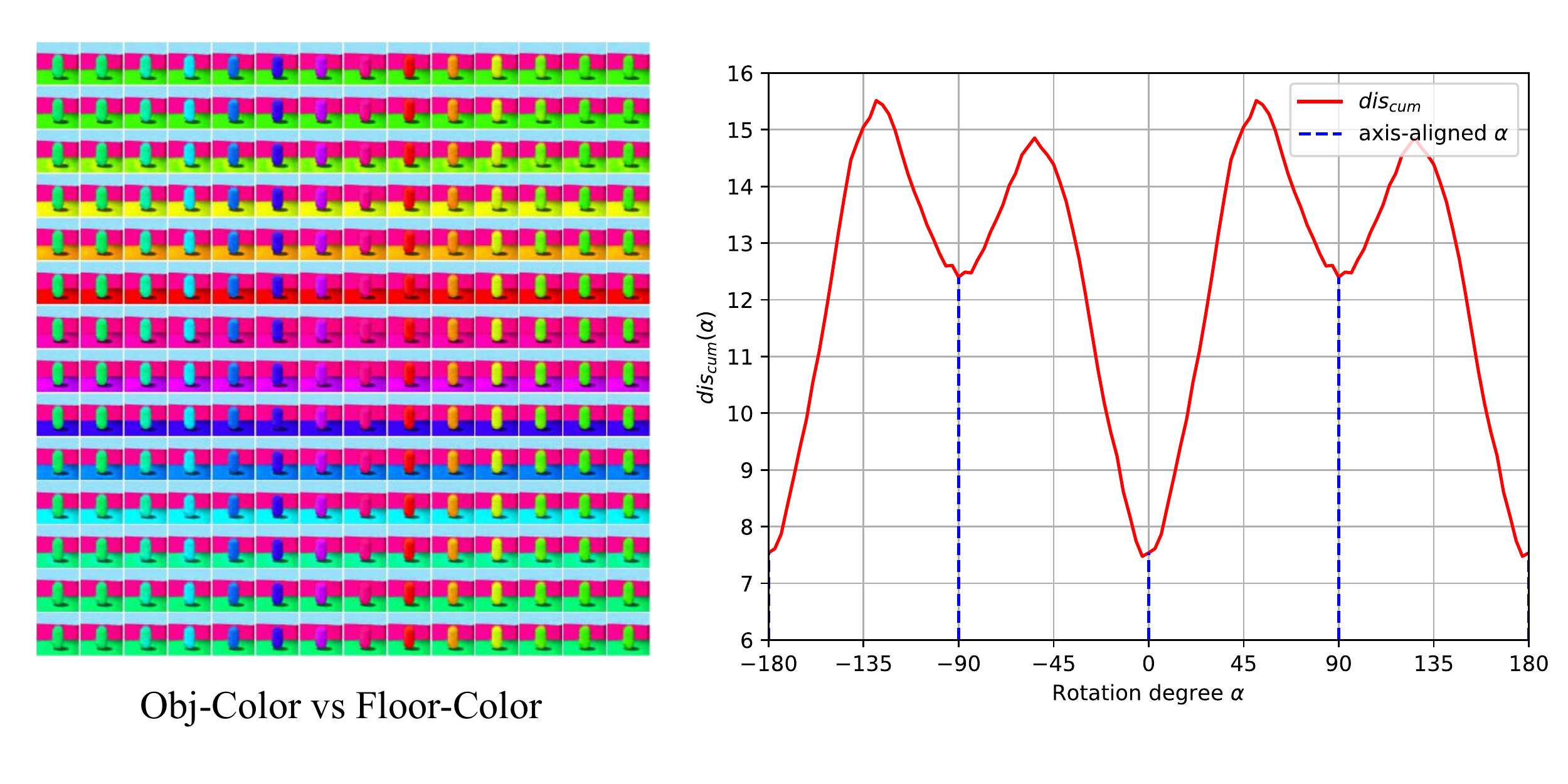}
\end{center}
    %\vspace{-8pt}
    \caption{Obj-Color vs Floor-Color.}
    %\vspace{-8pt}
\label{fig:objcolor_vs_floorcolor}
\end{figure}
\begin{figure}[t]
\begin{center}
   \includegraphics[width=\linewidth]{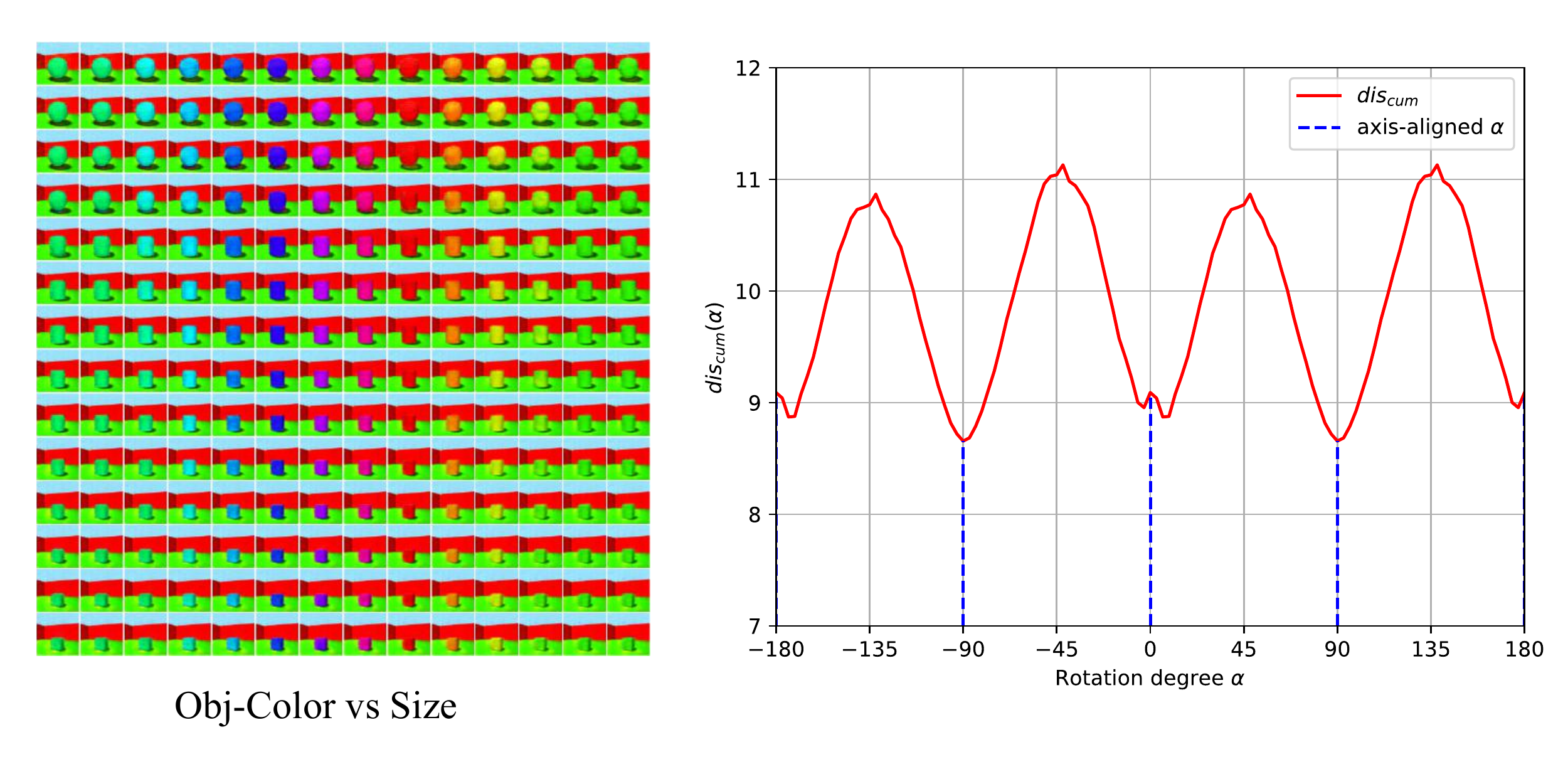}
\end{center}
    %\vspace{-8pt}
    \caption{Obj-Color vs Size.}
    %\vspace{-8pt}
\label{fig:objcolor_vs_size}
\end{figure}
\begin{figure}[t]
\begin{center}
   \includegraphics[width=\linewidth]{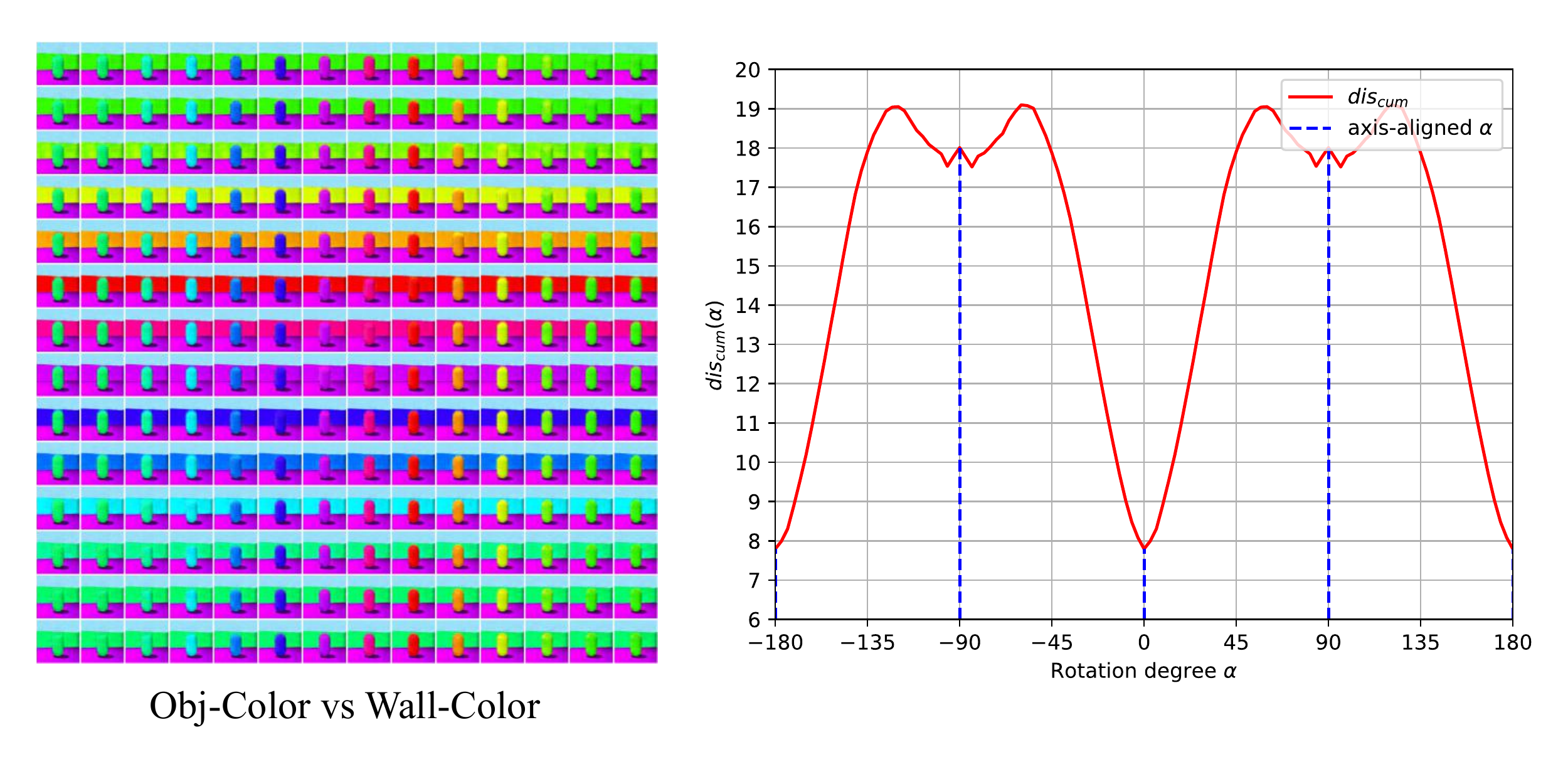}
\end{center}
    %\vspace{-8pt}
    \caption{Obj-Color vs Wall-Color.}
    %\vspace{-8pt}
\label{fig:objcolor_vs_wallcolor}
\end{figure}
\begin{figure}[t]
\begin{center}
   \includegraphics[width=\linewidth]{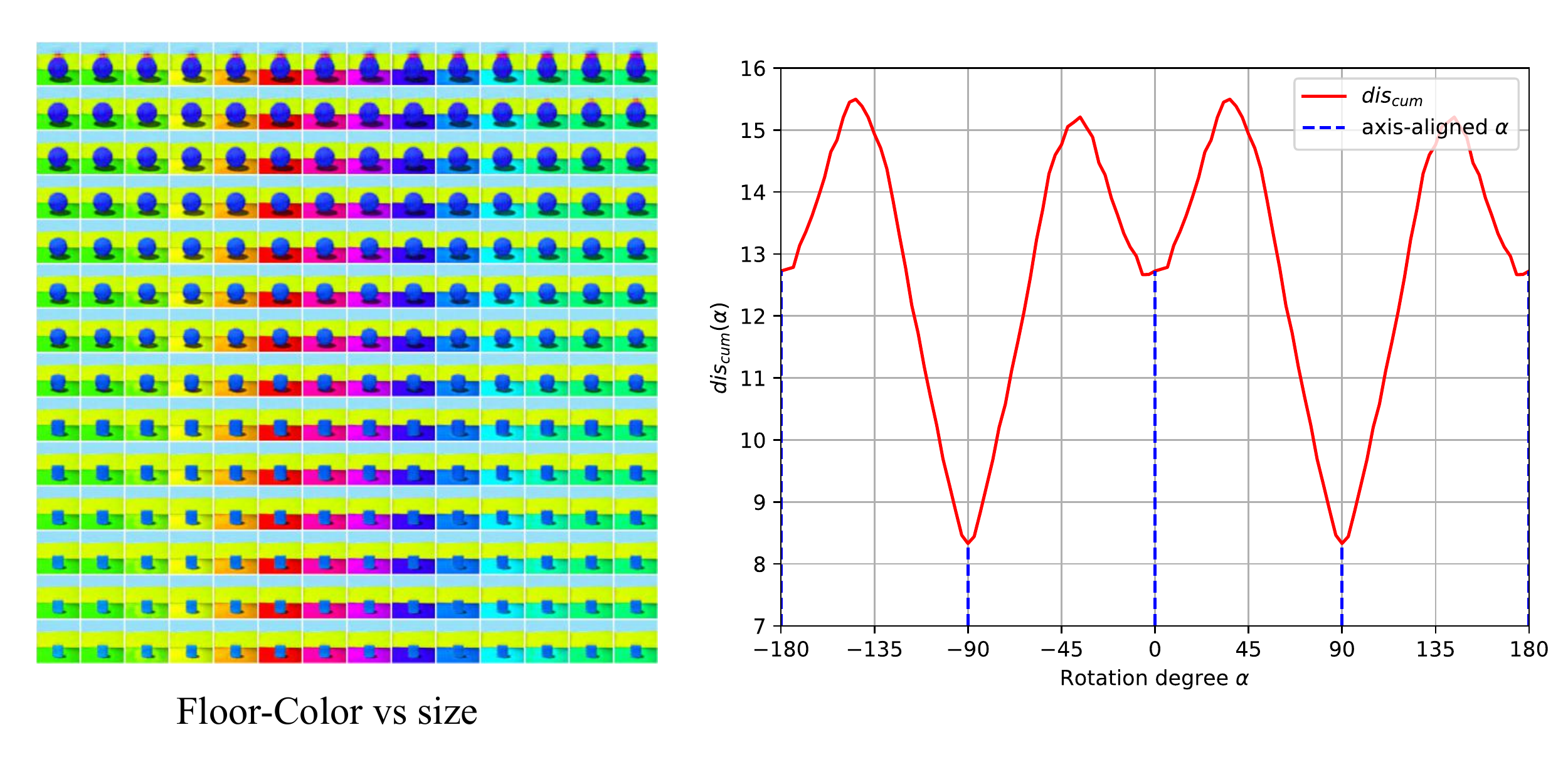}
\end{center}
    %\vspace{-8pt}
    \caption{Floor-Color vs Size.}
    %\vspace{-8pt}
\label{fig:floorcolor_vs_size}
\end{figure}
\begin{figure}[t]
\begin{center}
   \includegraphics[width=\linewidth]{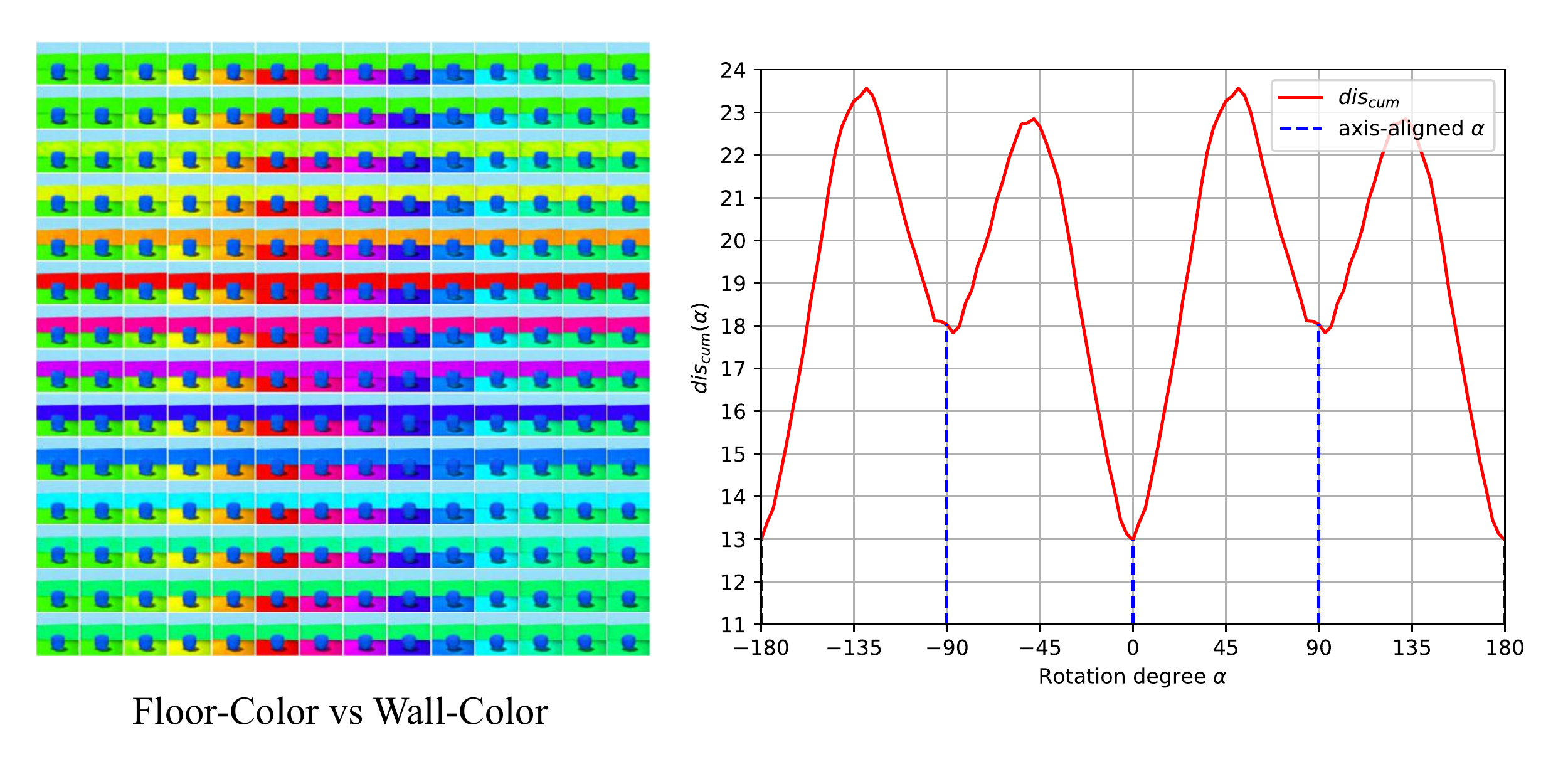}
\end{center}
    %\vspace{-8pt}
    \caption{Floor-Color vs Wall-Color.}
    %\vspace{-8pt}
\label{fig:floorcolor_vs_wallcolor}
\end{figure}
\begin{figure}[t]
\begin{center}
   \includegraphics[width=\linewidth]{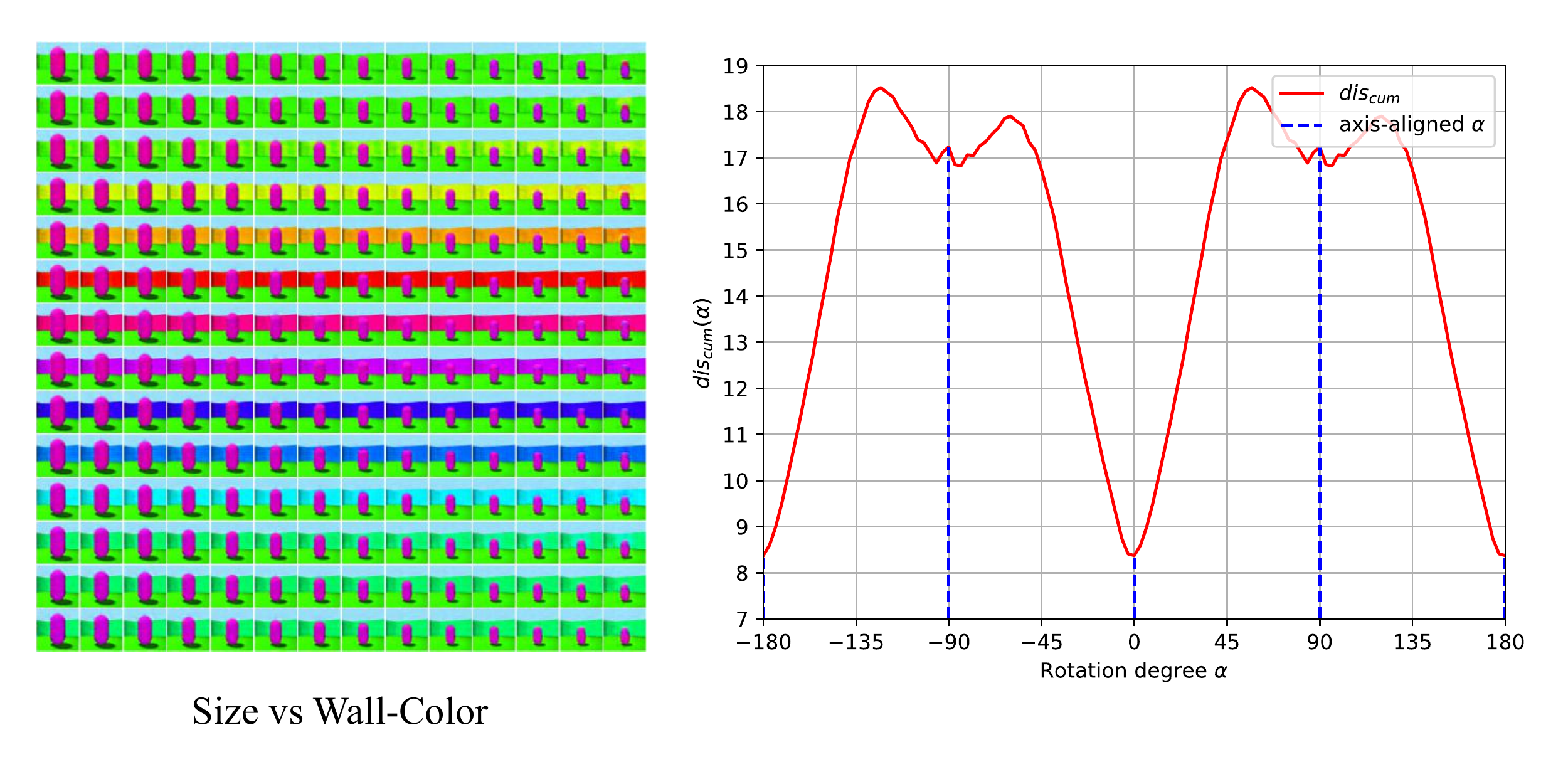}
\end{center}
    %\vspace{-8pt}
    \caption{Size vs Wall-Color.}
    %\vspace{-8pt}
\label{fig:size_vs_wallcolor}
\end{figure}

\begin{figure}[t]
\begin{center}
   \includegraphics[width=\linewidth]{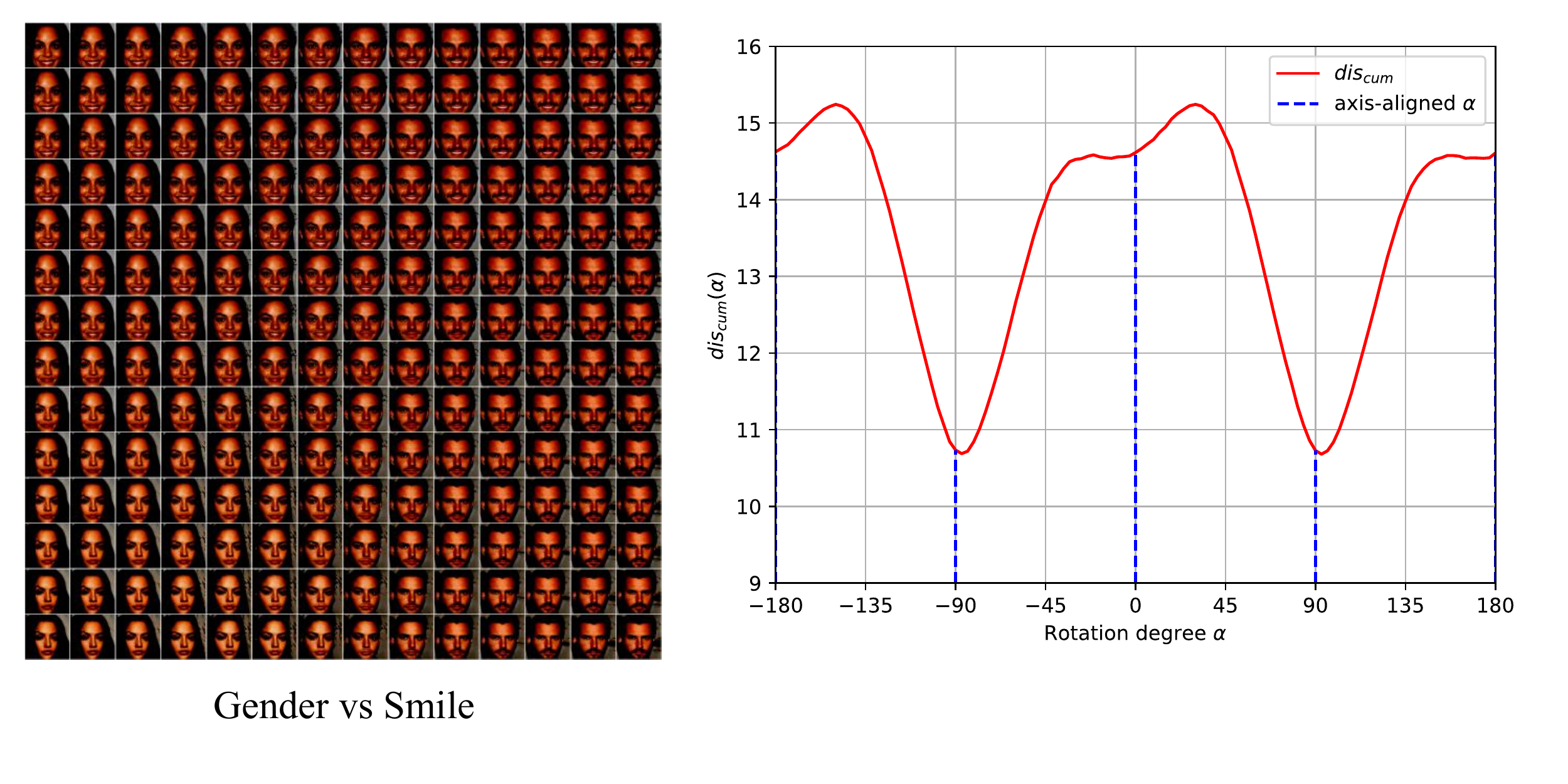}
\end{center}
    %\vspace{-8pt}
    \caption{Gender vs Smile.}
    %\vspace{-8pt}
\label{fig:gender_vs_smile}
\end{figure}
\begin{figure}[t]
\begin{center}
   \includegraphics[width=\linewidth]{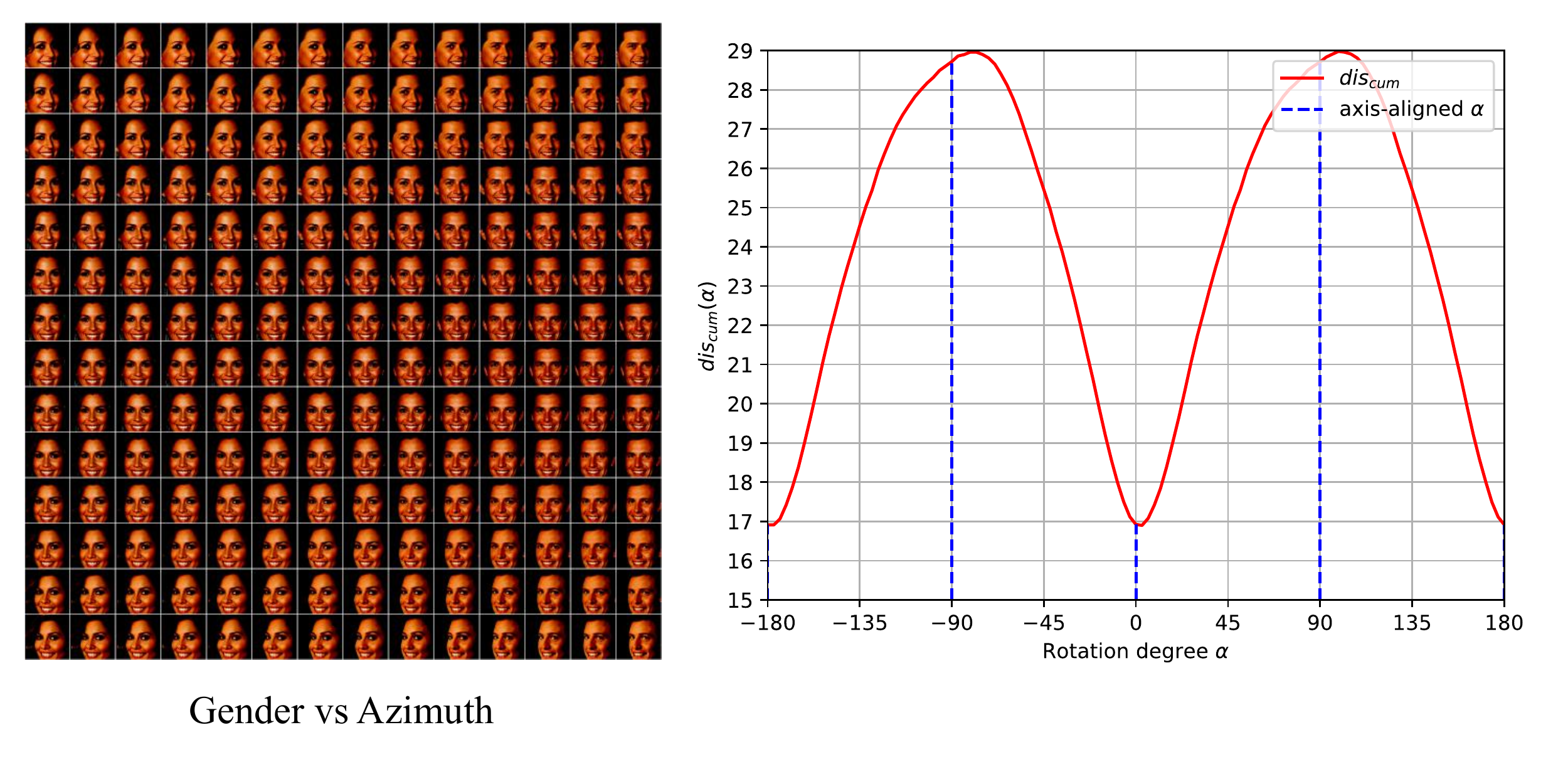}
\end{center}
    %\vspace{-8pt}
    \caption{Gender vs Azimuth.}
    %\vspace{-8pt}
\label{fig:gender_vs_azimuth}
\end{figure}
\begin{figure}[t]
\begin{center}
   \includegraphics[width=\linewidth]{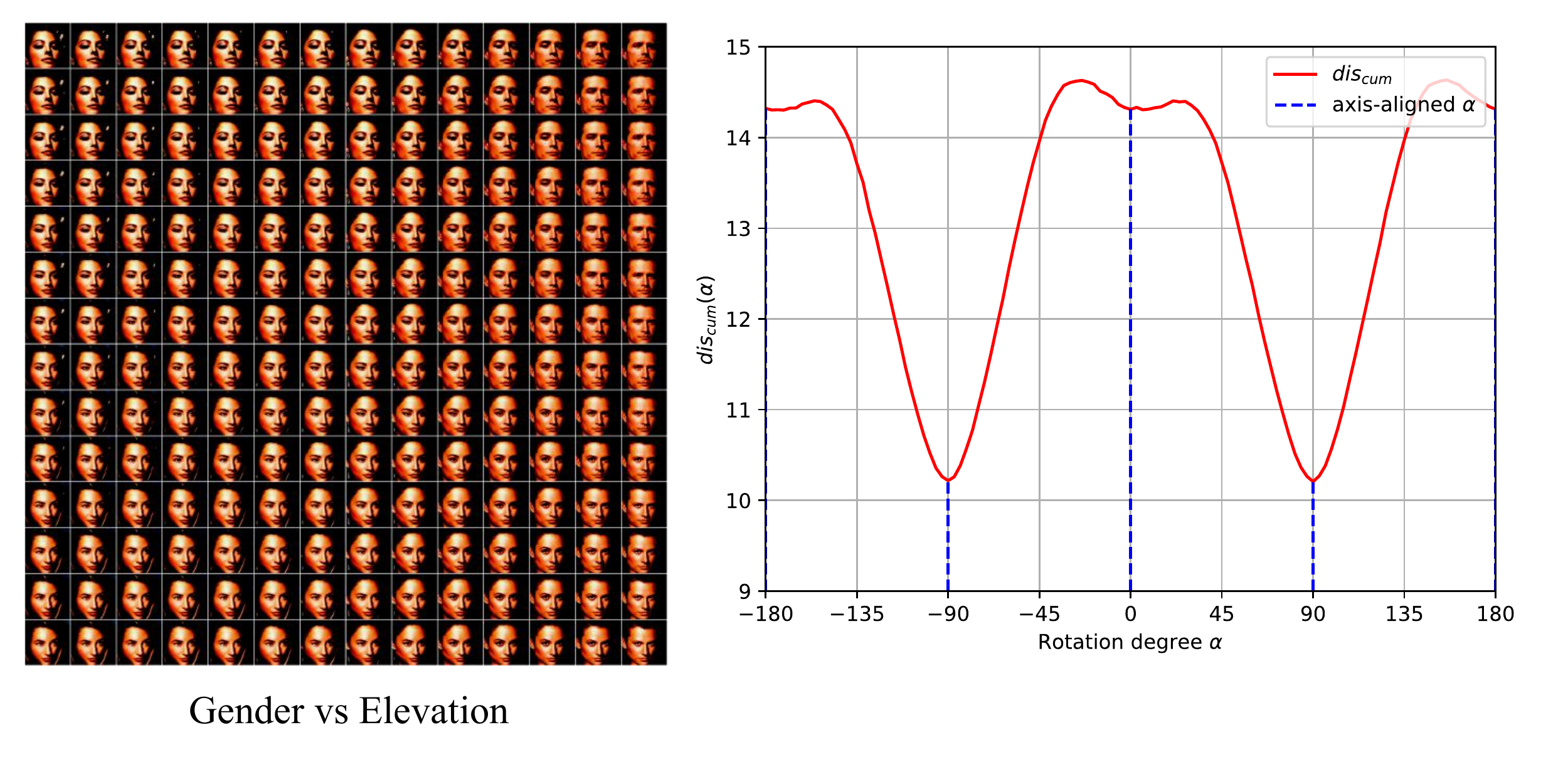}
\end{center}
    %\vspace{-8pt}
    \caption{Gender vs Elevation.}
    %\vspace{-8pt}
\label{fig:gender_vs_elevation}
\end{figure}
\begin{figure}[t]
\begin{center}
   \includegraphics[width=\linewidth]{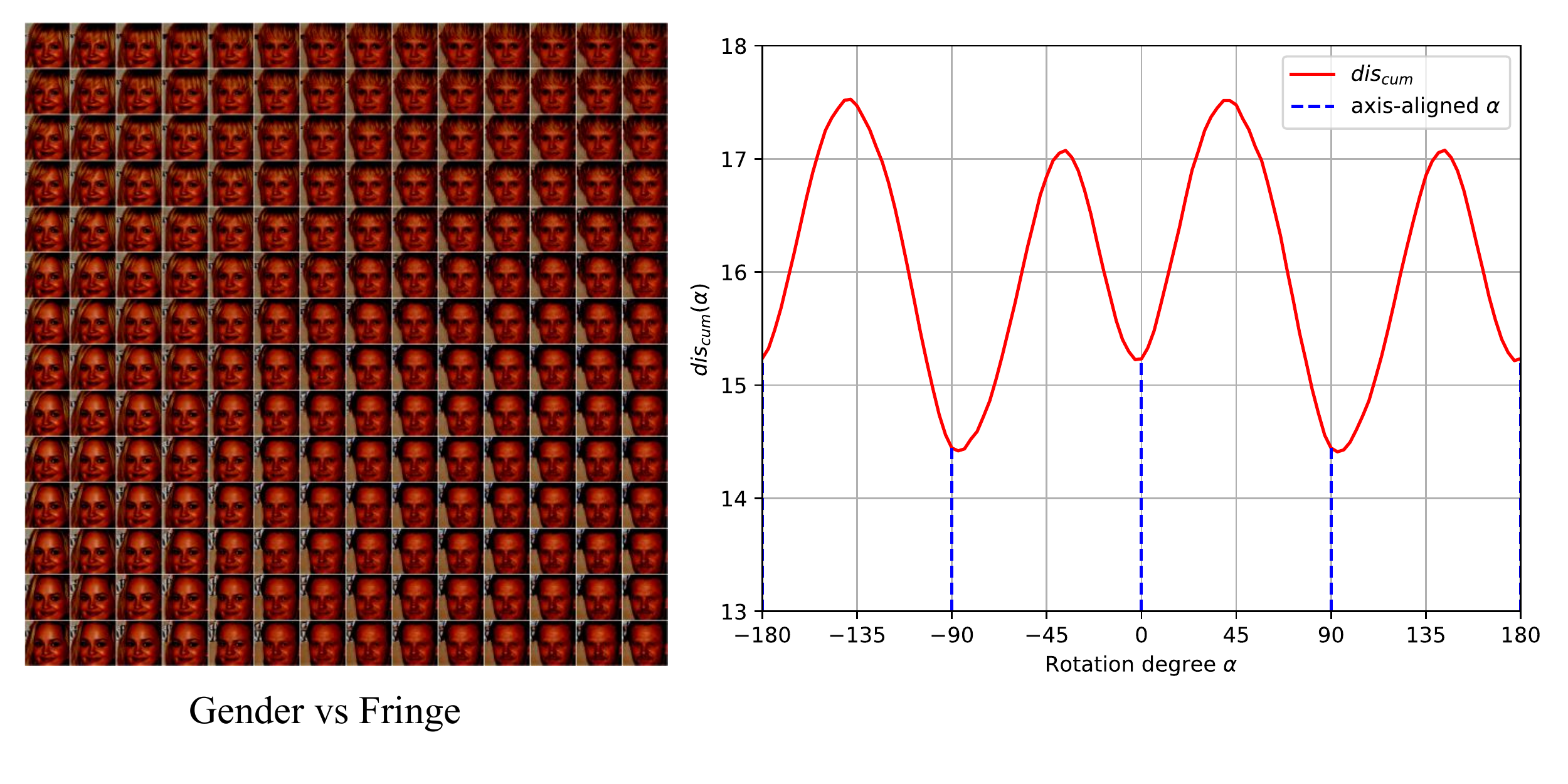}
\end{center}
    %\vspace{-8pt}
    \caption{Gender vs Fringe.}
    %\vspace{-8pt}
\label{fig:gender_vs_fringe}
\end{figure}
\begin{figure}[t]
\begin{center}
   \includegraphics[width=\linewidth]{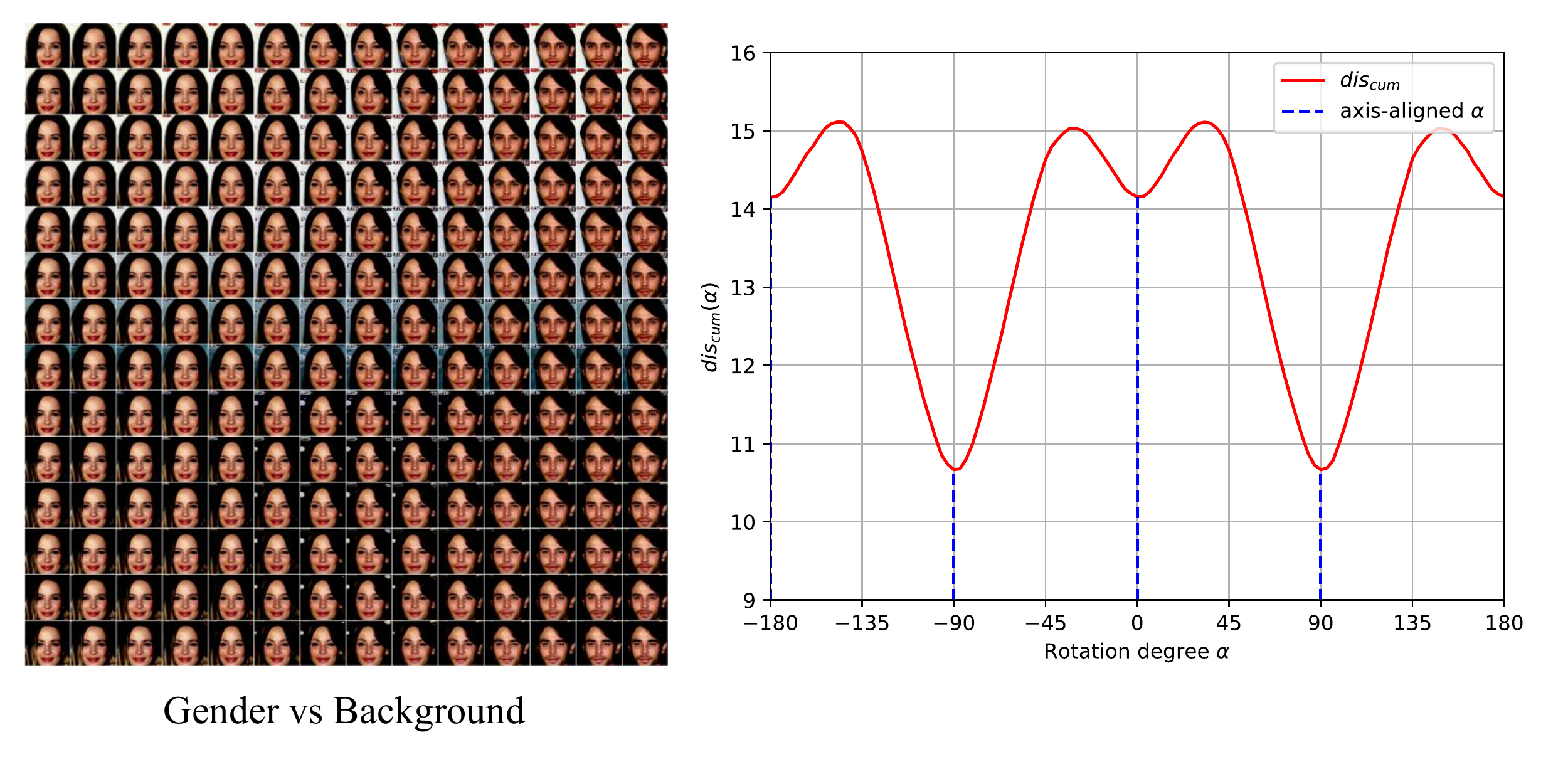}
\end{center}
    %\vspace{-8pt}
    \caption{Gender vs Background.}
    %\vspace{-8pt}
\label{fig:gender_vs_background}
\end{figure}
\begin{figure}[t]
\begin{center}
   \includegraphics[width=\linewidth]{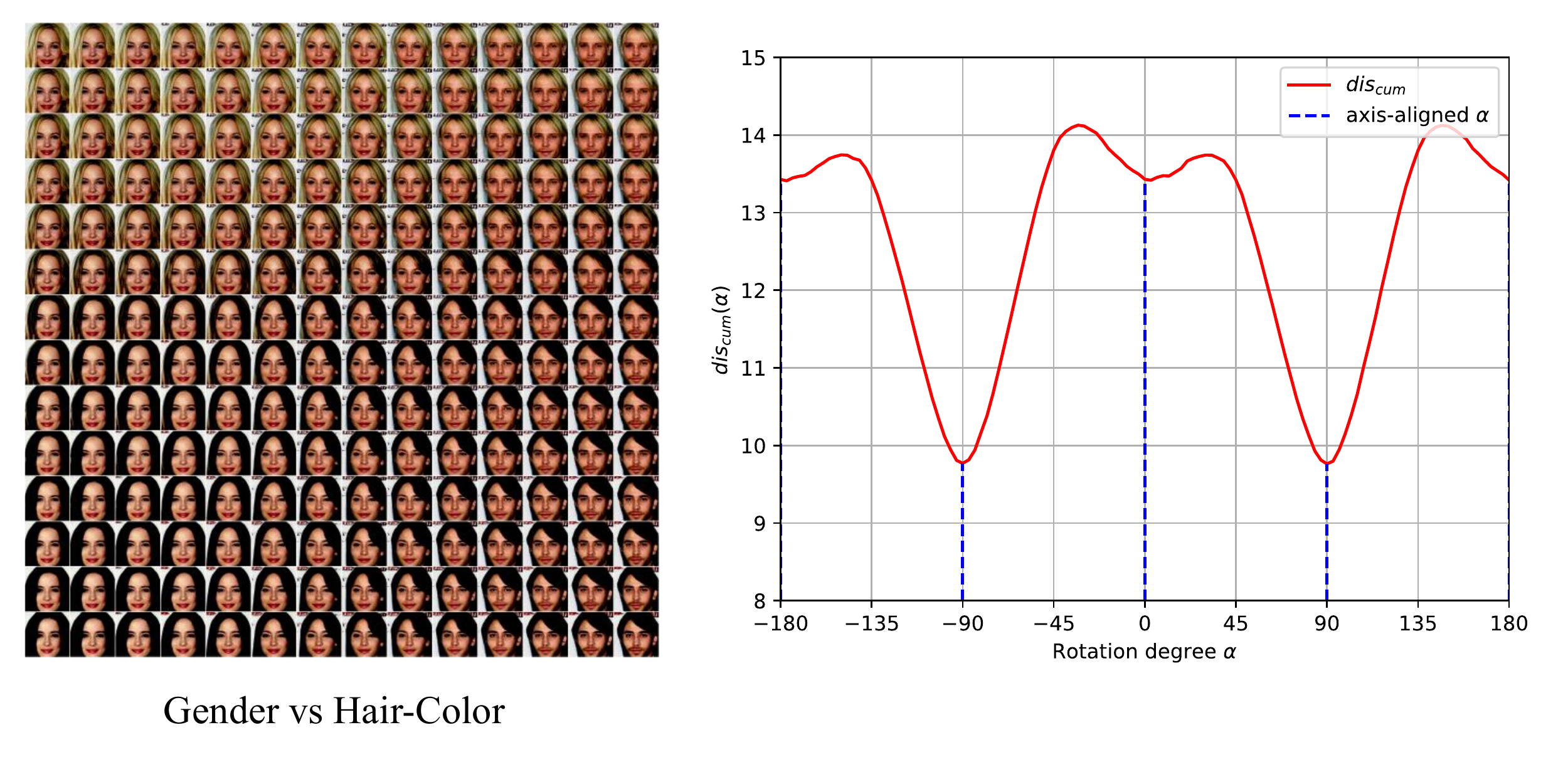}
\end{center}
    %\vspace{-8pt}
    \caption{Gender vs Hair-Color.}
    %\vspace{-8pt}
\label{fig:gender_vs_haircolor}
\end{figure}
\begin{figure}[t]
\begin{center}
   \includegraphics[width=\linewidth]{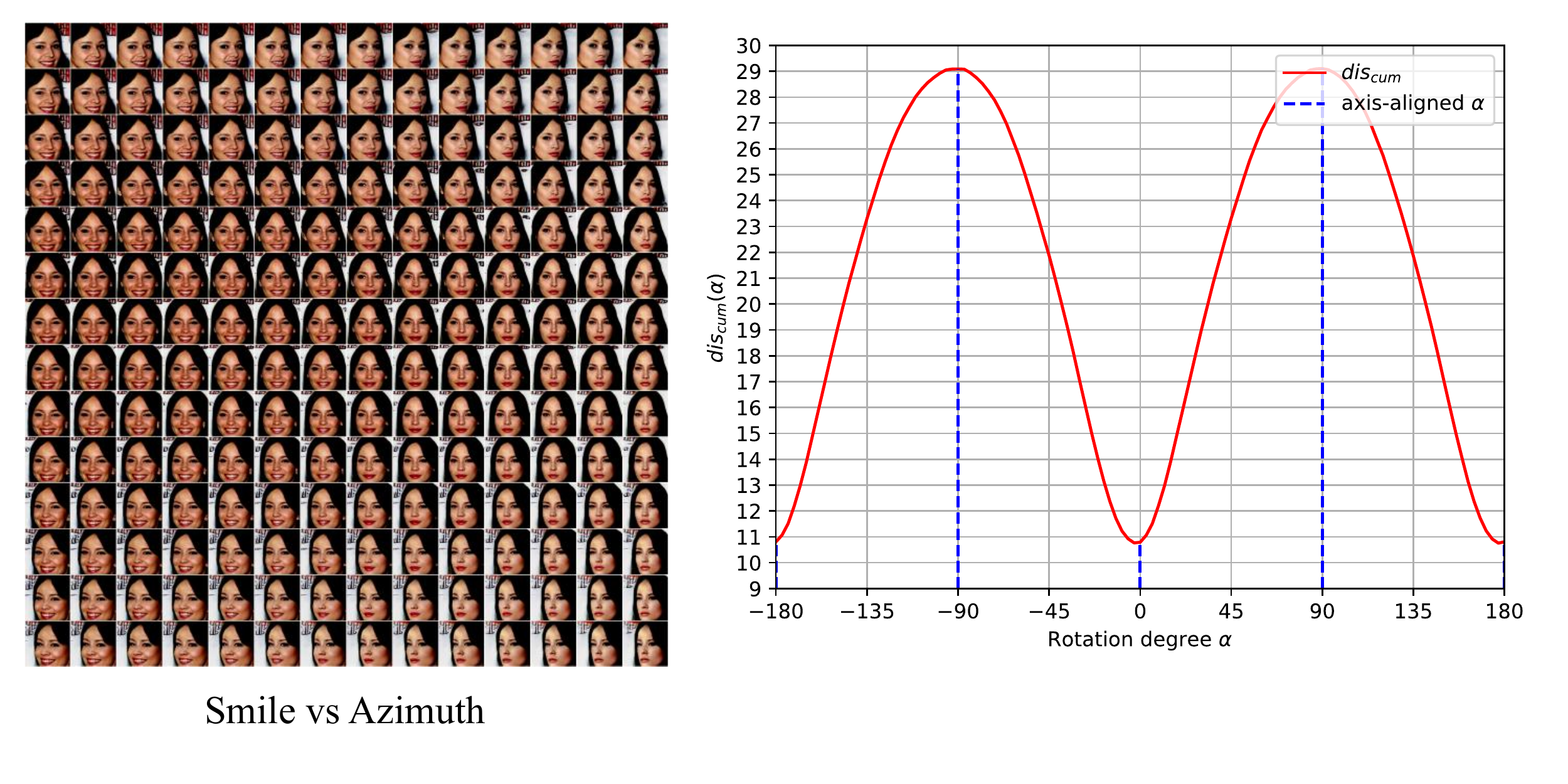}
\end{center}
    %\vspace{-8pt}
    \caption{Smile vs Azimuth.}
    %\vspace{-8pt}
\label{fig:smile_vs_azimuth}
\end{figure}
\begin{figure}[t]
\begin{center}
   \includegraphics[width=\linewidth]{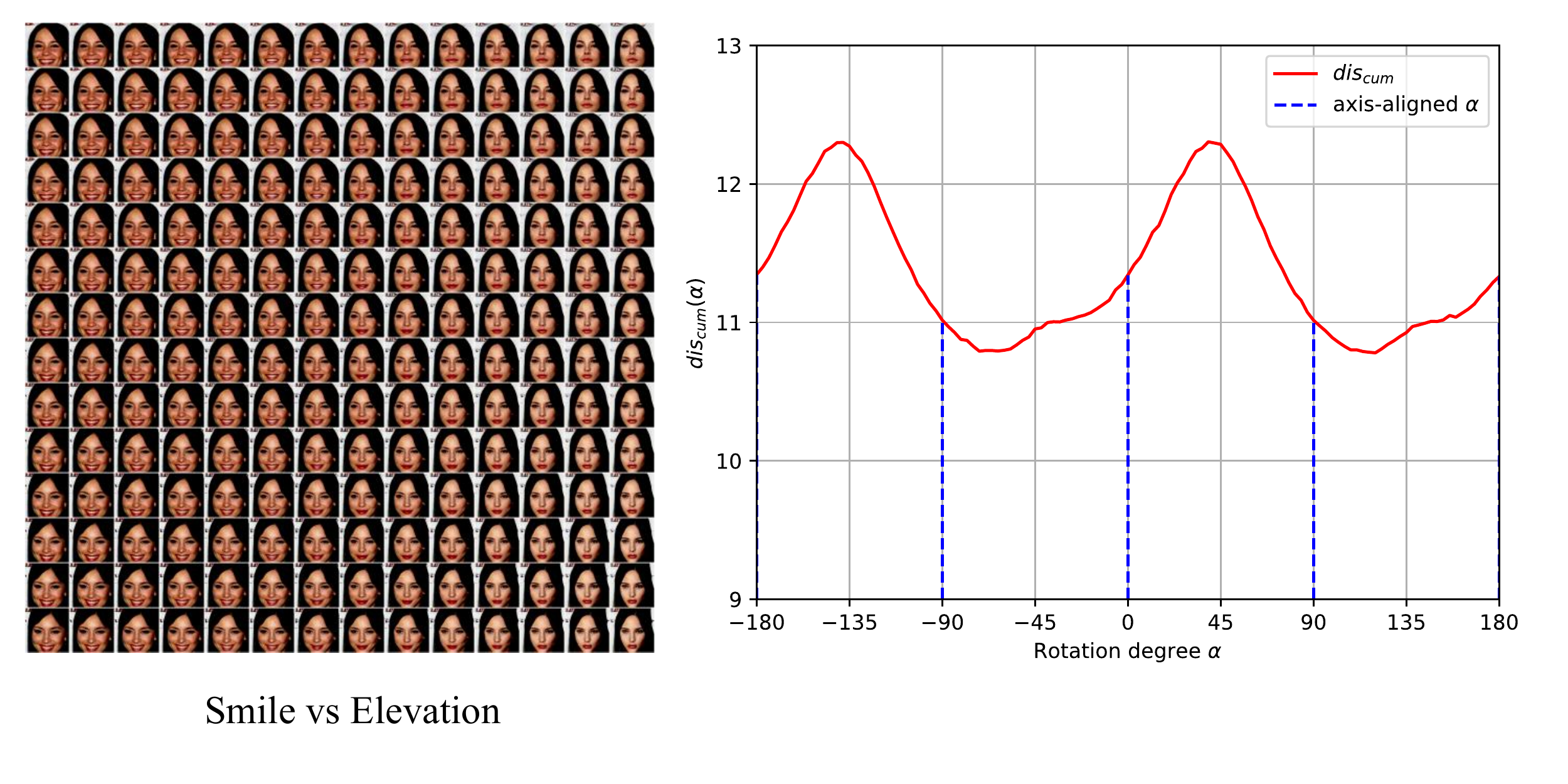}
\end{center}
    %\vspace{-8pt}
    \caption{Smile vs Elevation.}
    %\vspace{-8pt}
\label{fig:smile_vs_elevation}
\end{figure}
\begin{figure}[t]
\begin{center}
   \includegraphics[width=\linewidth]{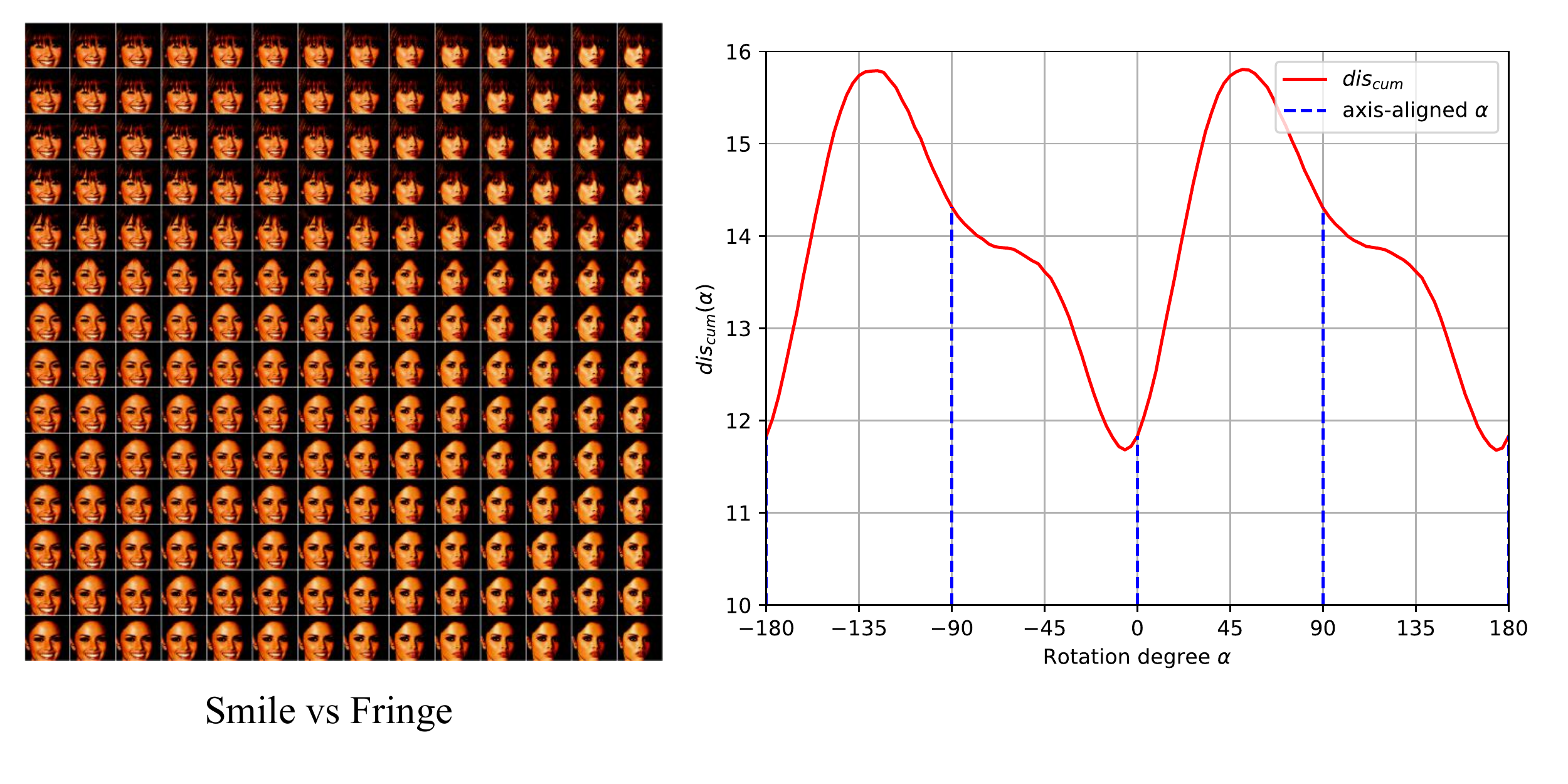}
\end{center}
    %\vspace{-8pt}
    \caption{Smile vs Fringe.}
    %\vspace{-8pt}
\label{fig:smile_vs_fringe}
\end{figure}
\begin{figure}[t]
\begin{center}
   \includegraphics[width=\linewidth]{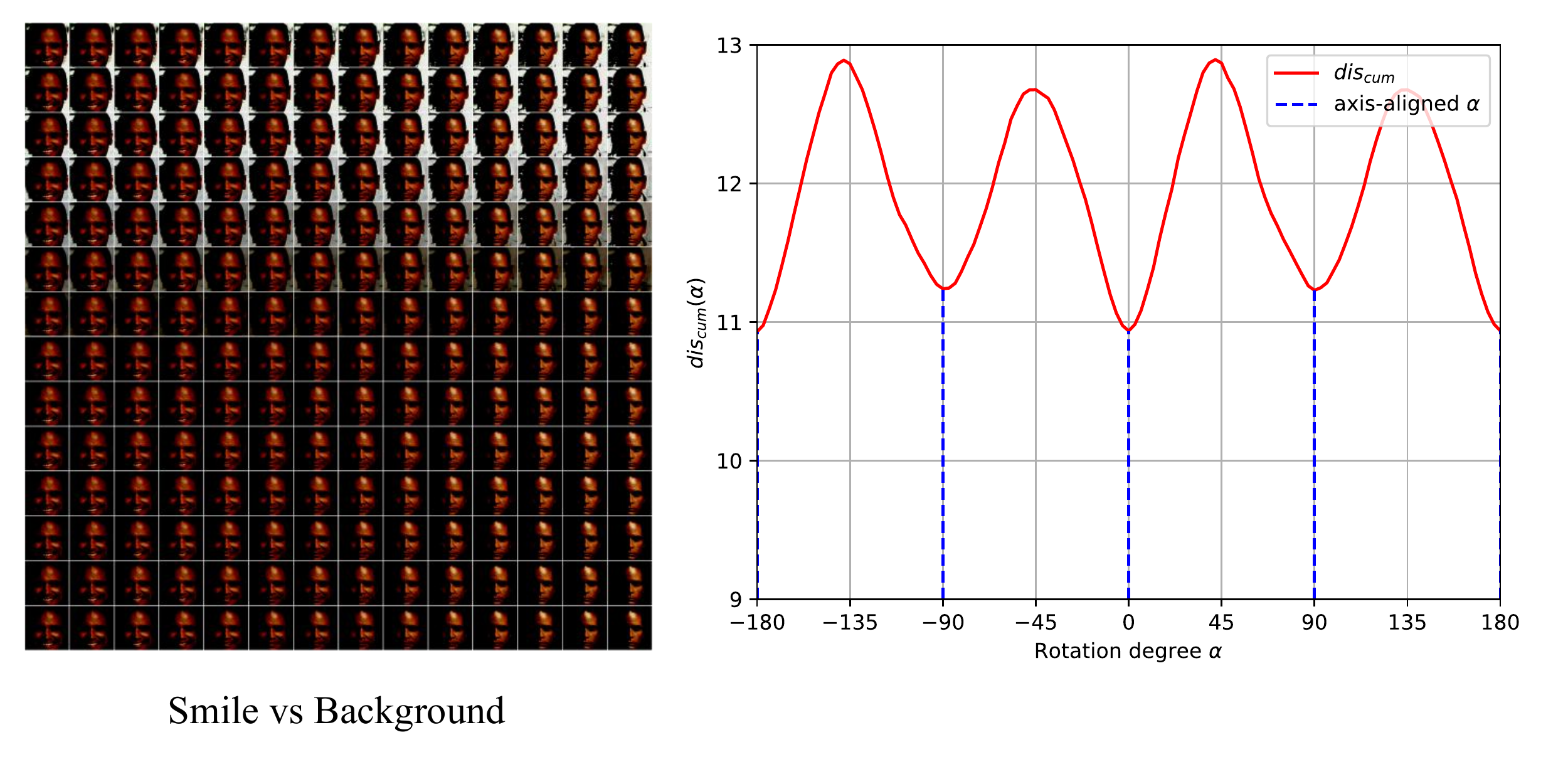}
\end{center}
    %\vspace{-8pt}
    \caption{Smile vs Background.}
    %\vspace{-8pt}
\label{fig:smile_vs_background}
\end{figure}
\begin{figure}[t]
\begin{center}
   \includegraphics[width=\linewidth]{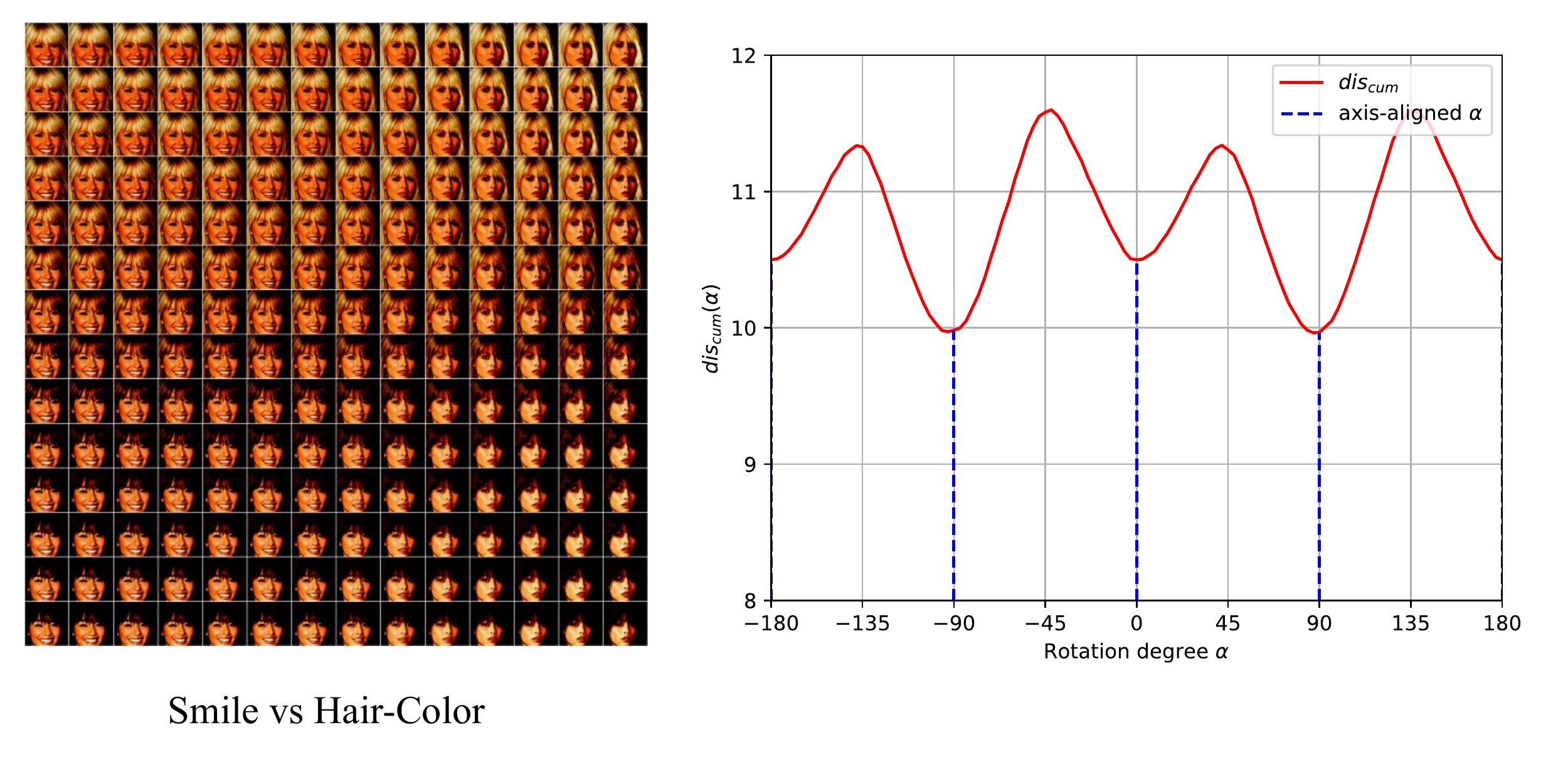}
\end{center}
    %\vspace{-8pt}
    \caption{Smile vs Hair-Color.}
    %\vspace{-8pt}
\label{fig:smile_vs_haircolor}
\end{figure}
\begin{figure}[t]
\begin{center}
   \includegraphics[width=\linewidth]{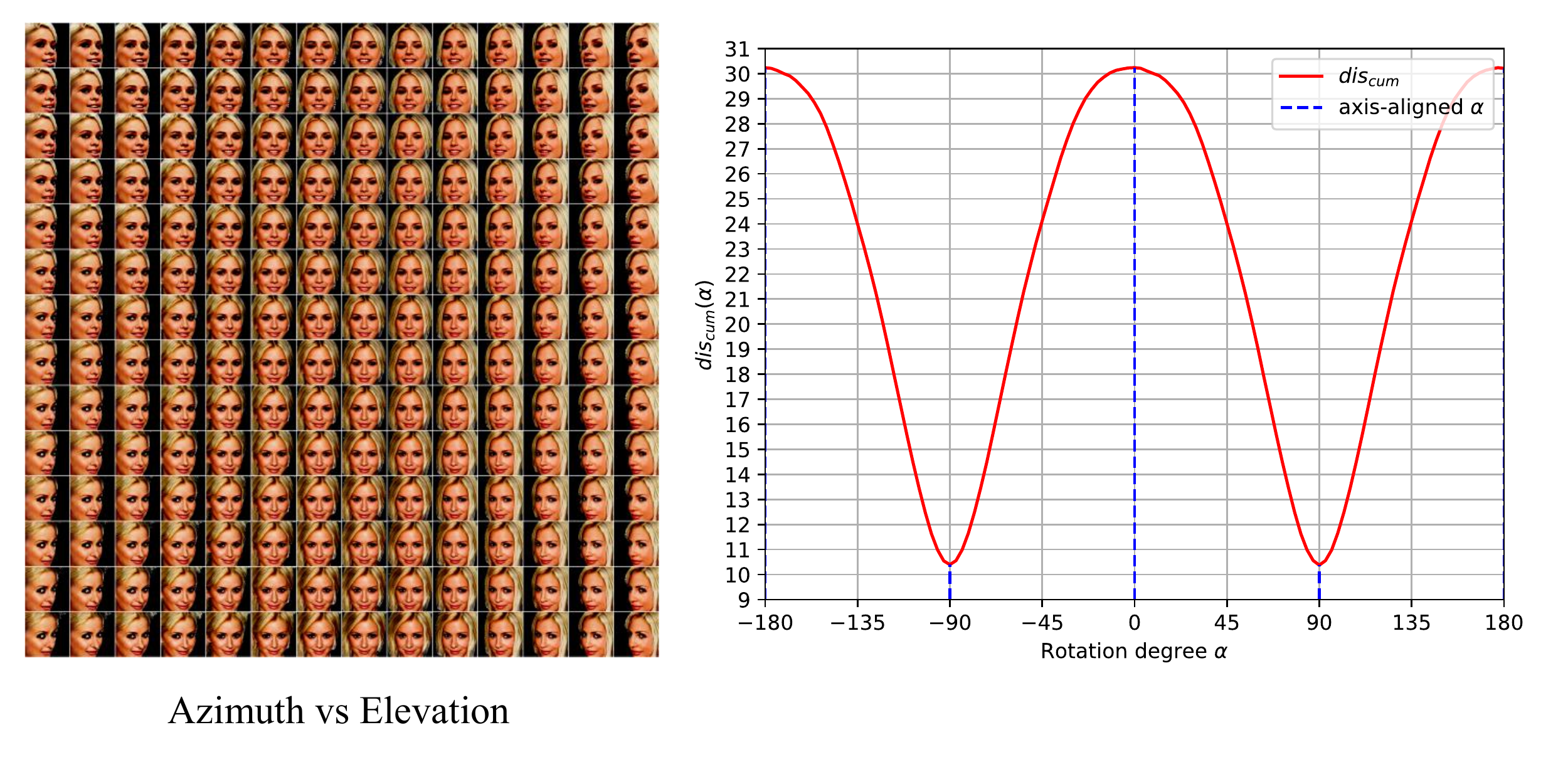}
\end{center}
    %\vspace{-8pt}
    \caption{Azimuth vs Elevation.}
    %\vspace{-8pt}
\label{fig:azimuth_vs_elevation}
\end{figure}
\begin{figure}[t]
\begin{center}
   \includegraphics[width=\linewidth]{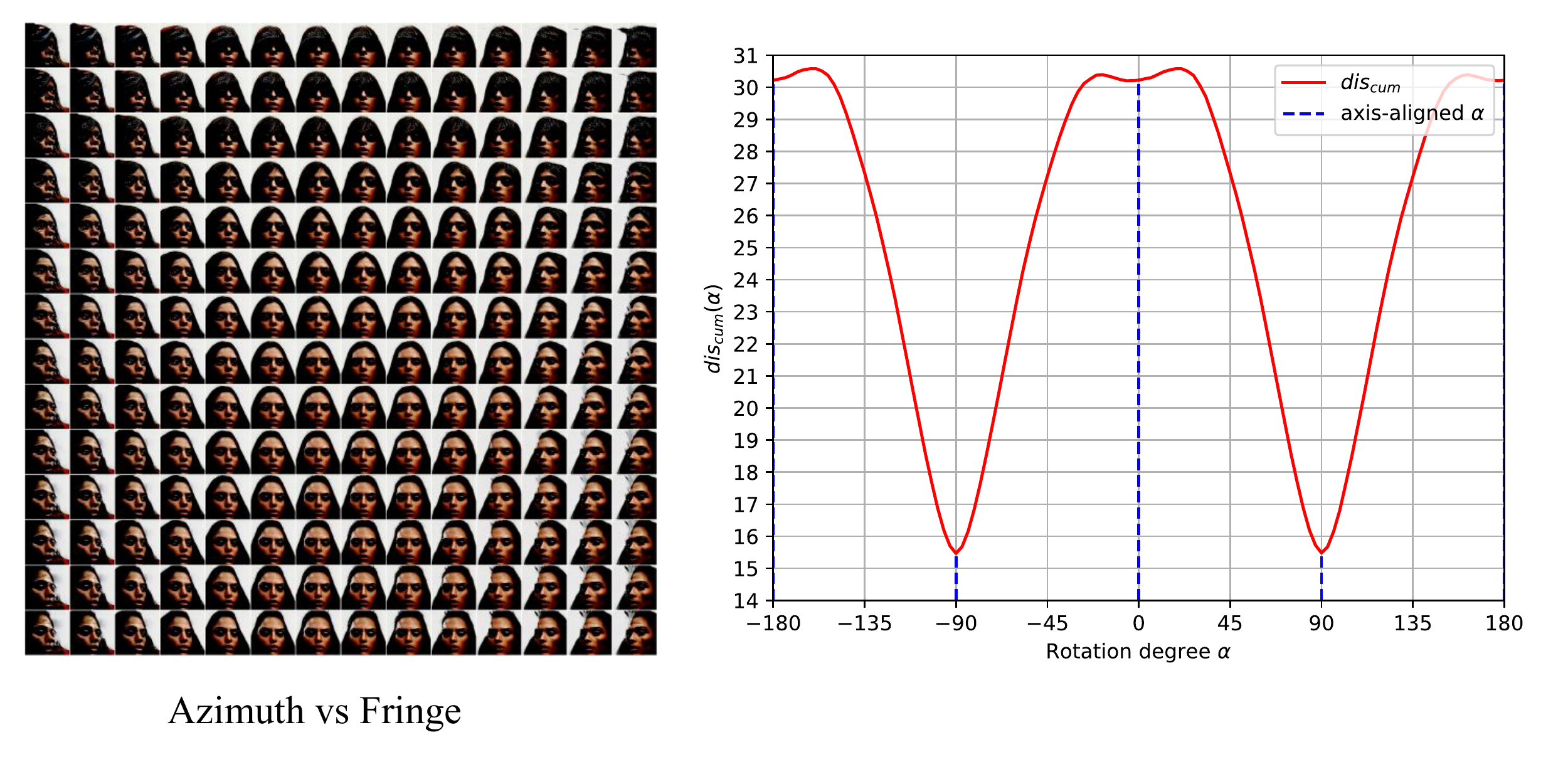}
\end{center}
    %\vspace{-8pt}
    \caption{Azimuth vs Fringe.}
    %\vspace{-8pt}
\label{fig:azimuth_vs_fringe}
\end{figure}
\begin{figure}[t]
\begin{center}
   \includegraphics[width=\linewidth]{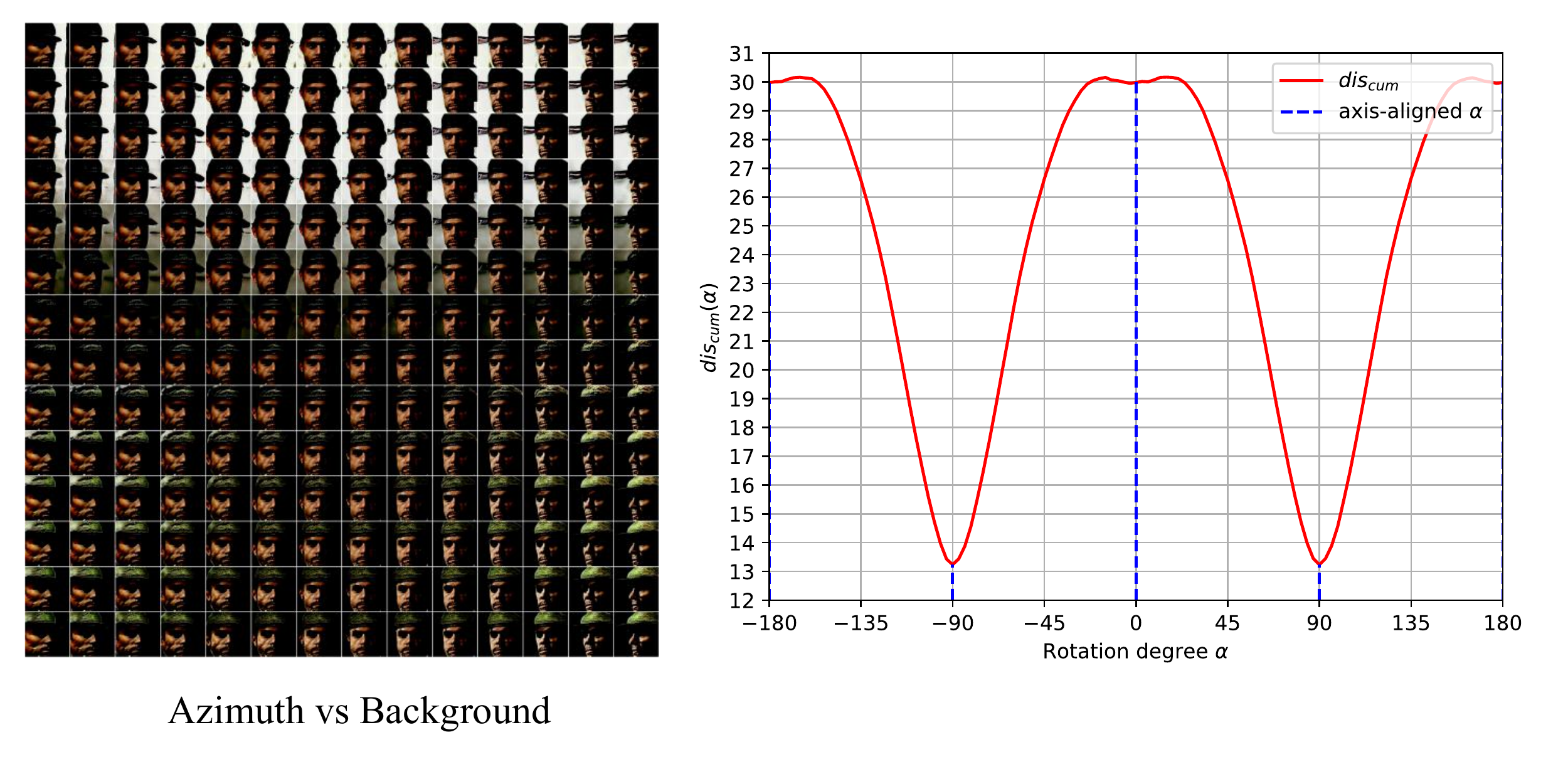}
\end{center}
    %\vspace{-8pt}
    \caption{Azimuth vs Background.}
    %\vspace{-8pt}
\label{fig:azimuth_vs_background}
\end{figure}
\begin{figure}[t]
\begin{center}
   \includegraphics[width=\linewidth]{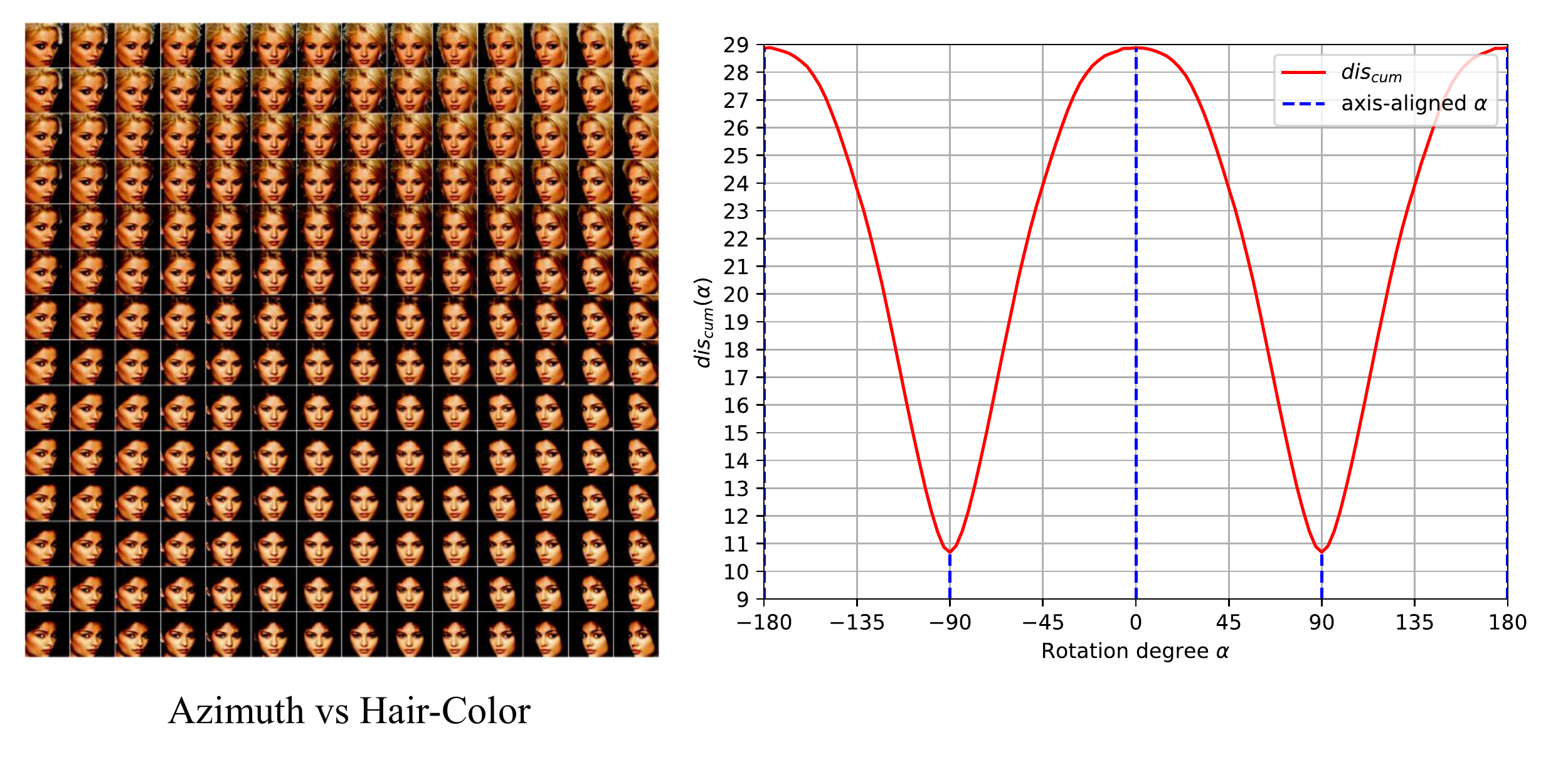}
\end{center}
    %\vspace{-8pt}
    \caption{Azimuth vs Hair-Color.}
    %\vspace{-8pt}
\label{fig:azimuth_vs_haircolor}
\end{figure}
\begin{figure}[t]
\begin{center}
   \includegraphics[width=\linewidth]{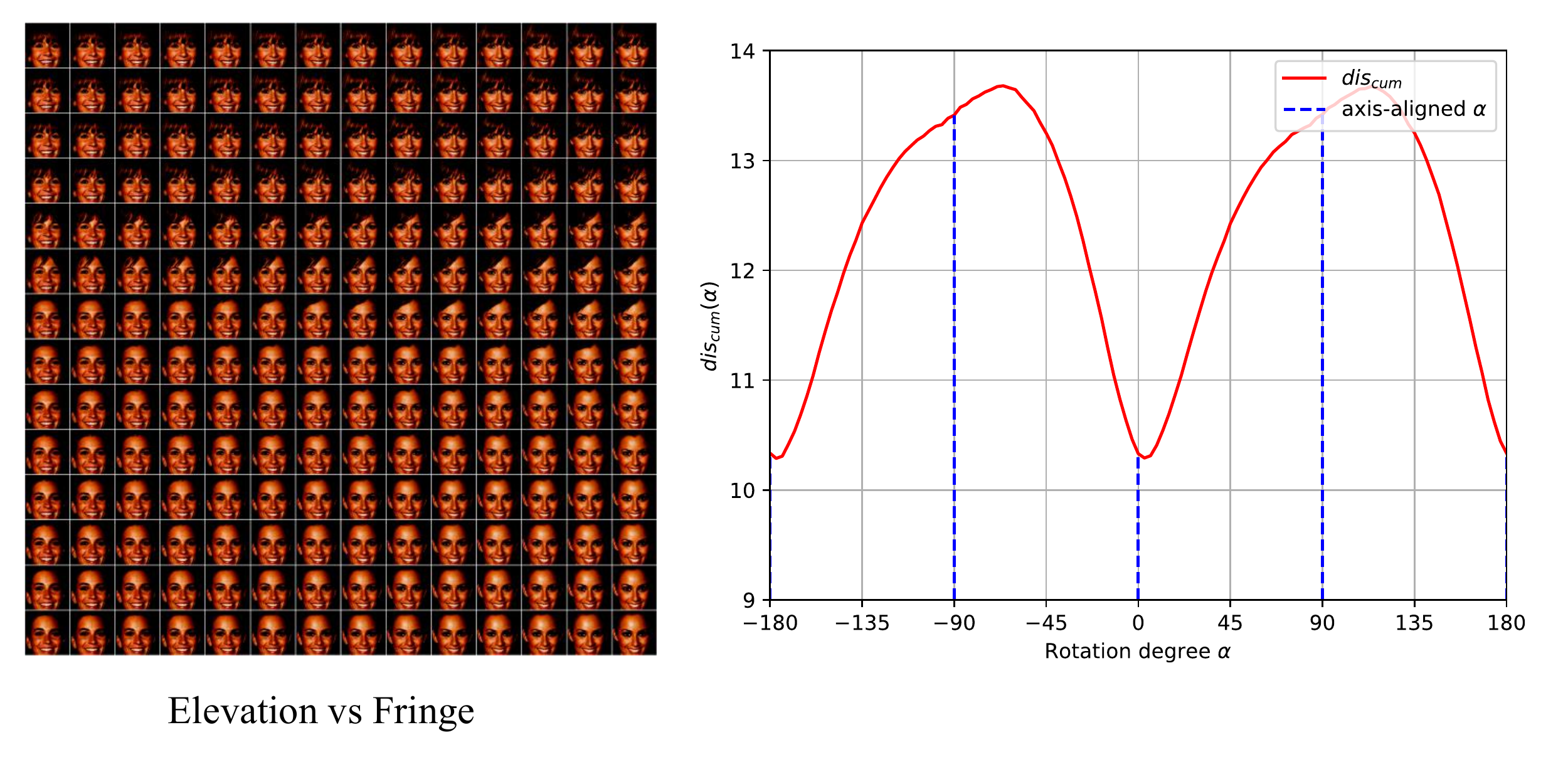}
\end{center}
    %\vspace{-8pt}
    \caption{Elevation vs Fringe.}
    %\vspace{-8pt}
\label{fig:elevation_vs_fringe}
\end{figure}
\begin{figure}[t]
\begin{center}
   \includegraphics[width=\linewidth]{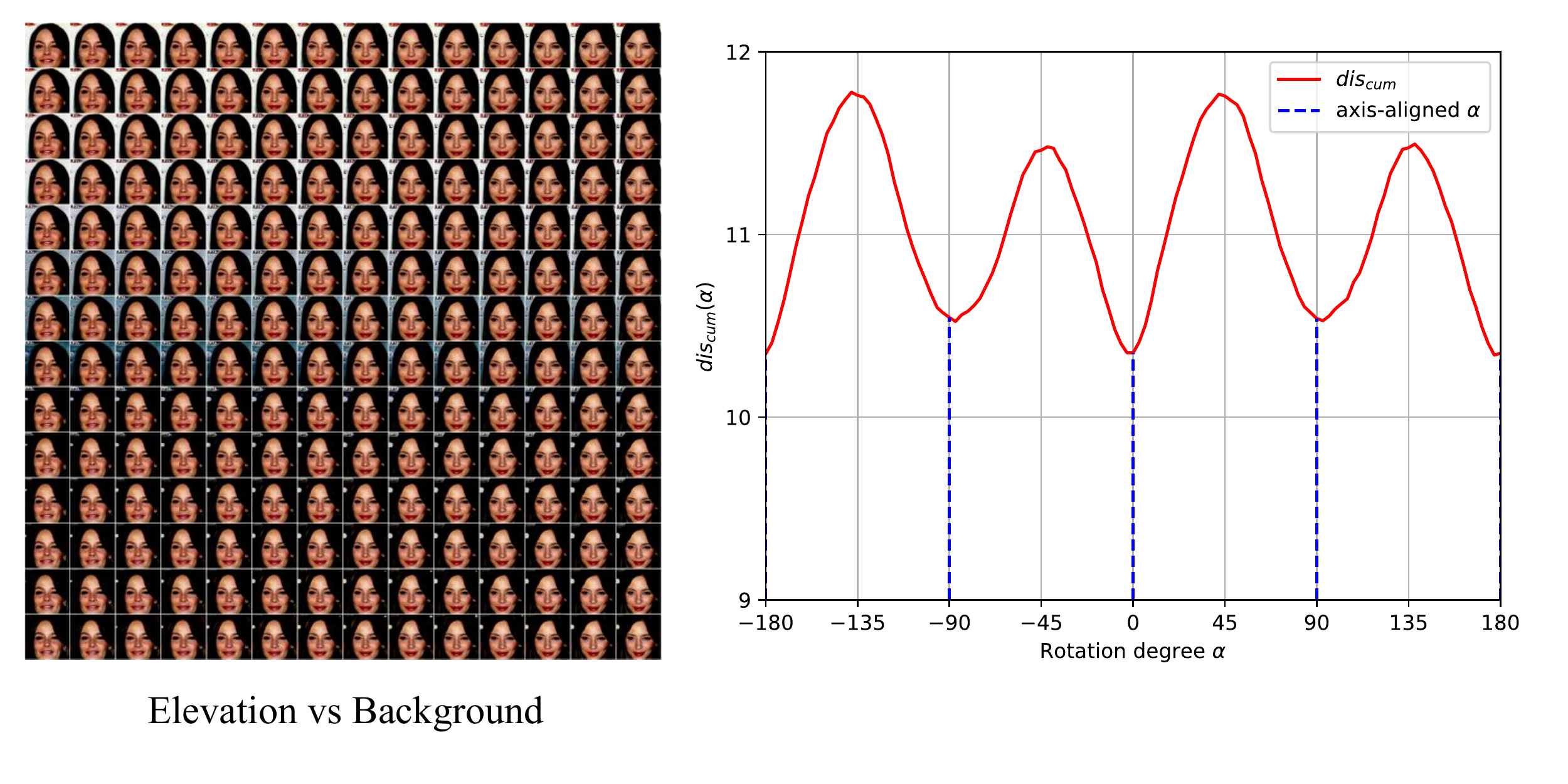}
\end{center}
    %\vspace{-8pt}
    \caption{Elevation vs Background.}
    %\vspace{-8pt}
\label{fig:elevation_vs_background}
\end{figure}
\begin{figure}[t]
\begin{center}
   \includegraphics[width=\linewidth]{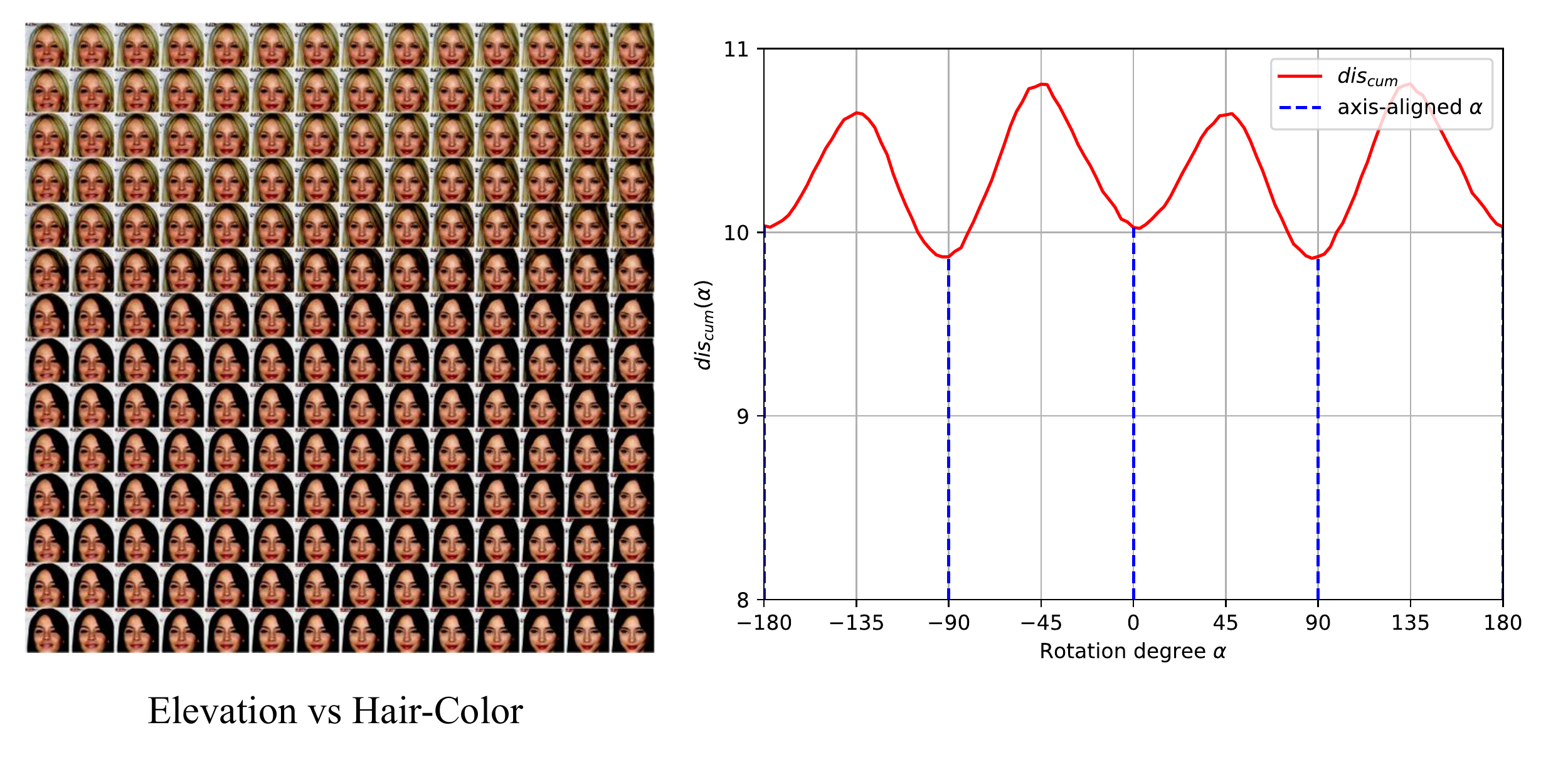}
\end{center}
    %\vspace{-8pt}
    \caption{Elevation vs Hair-Color.}
    %\vspace{-8pt}
\label{fig:elevation_vs_haircolor}
\end{figure}
\begin{figure}[t]
\begin{center}
   \includegraphics[width=\linewidth]{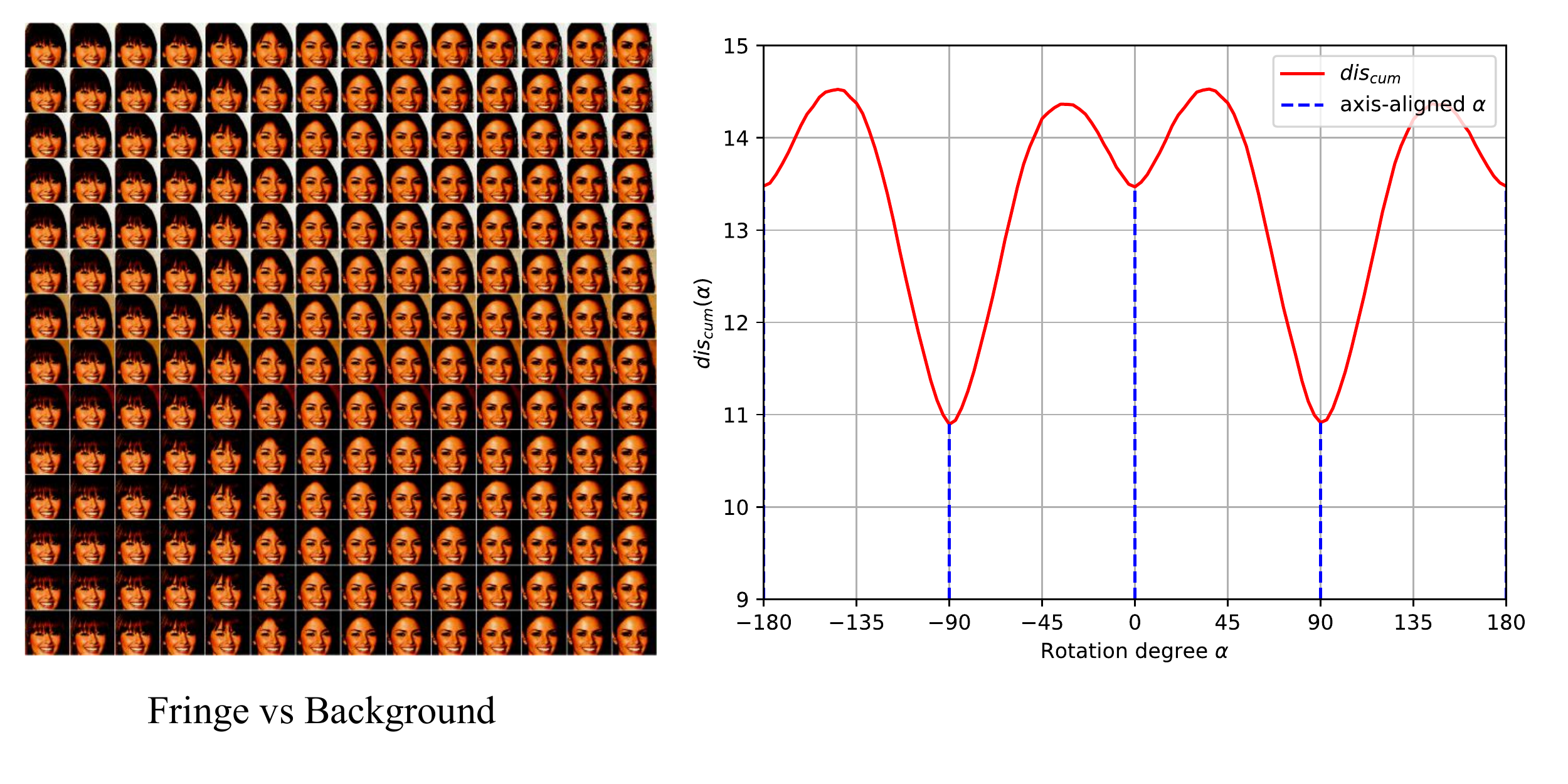}
\end{center}
    %\vspace{-8pt}
    \caption{Fringe vs Background.}
    %\vspace{-8pt}
\label{fig:fringe_vs_background}
\end{figure}
\begin{figure}[t]
\begin{center}
   \includegraphics[width=\linewidth]{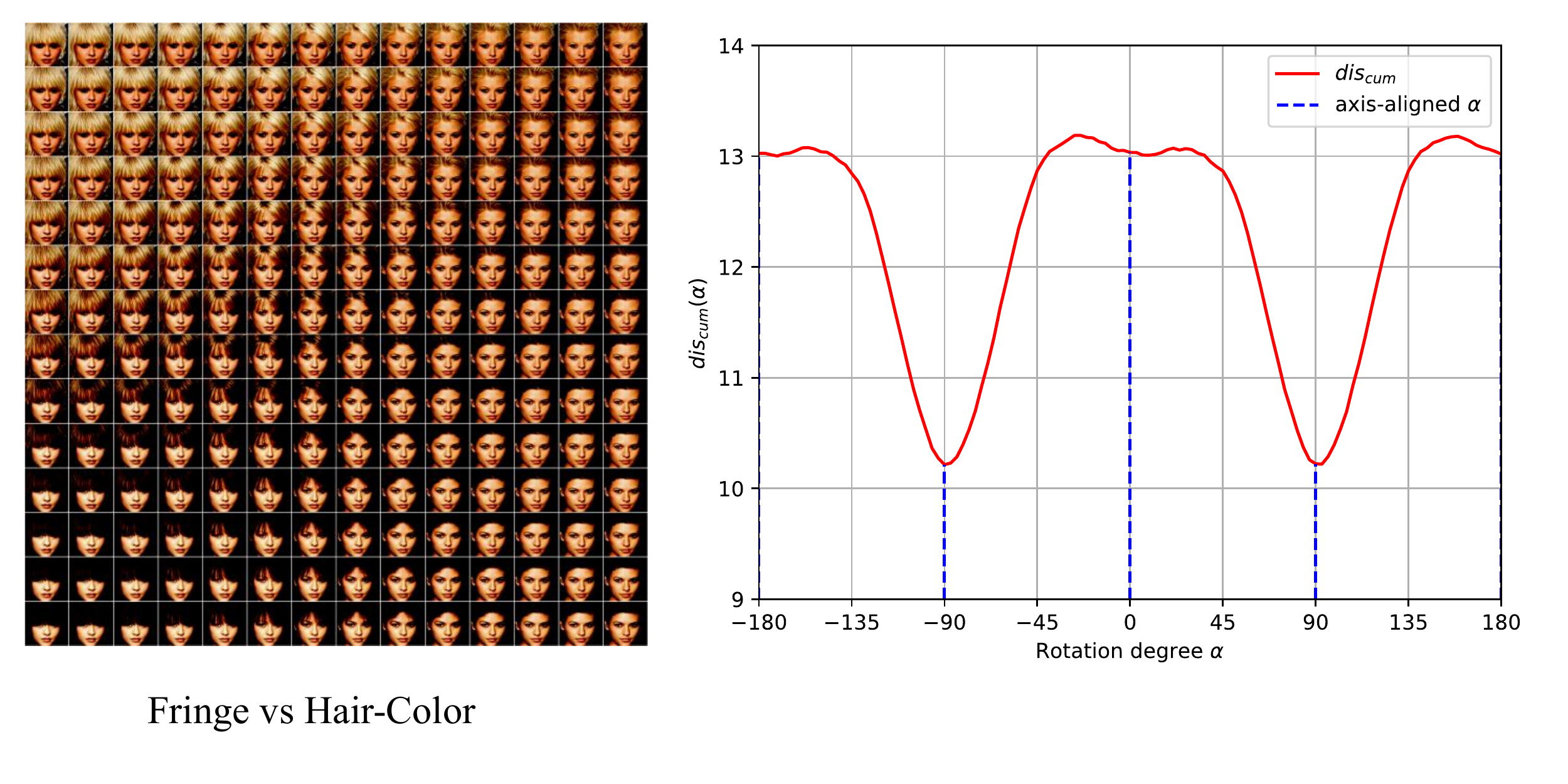}
\end{center}
    %\vspace{-8pt}
    \caption{Fringe vs Hair-Color.}
    %\vspace{-8pt}
\label{fig:fringe_vs_haircolor}
\end{figure}
\begin{figure}[t]
\begin{center}
   \includegraphics[width=\linewidth]{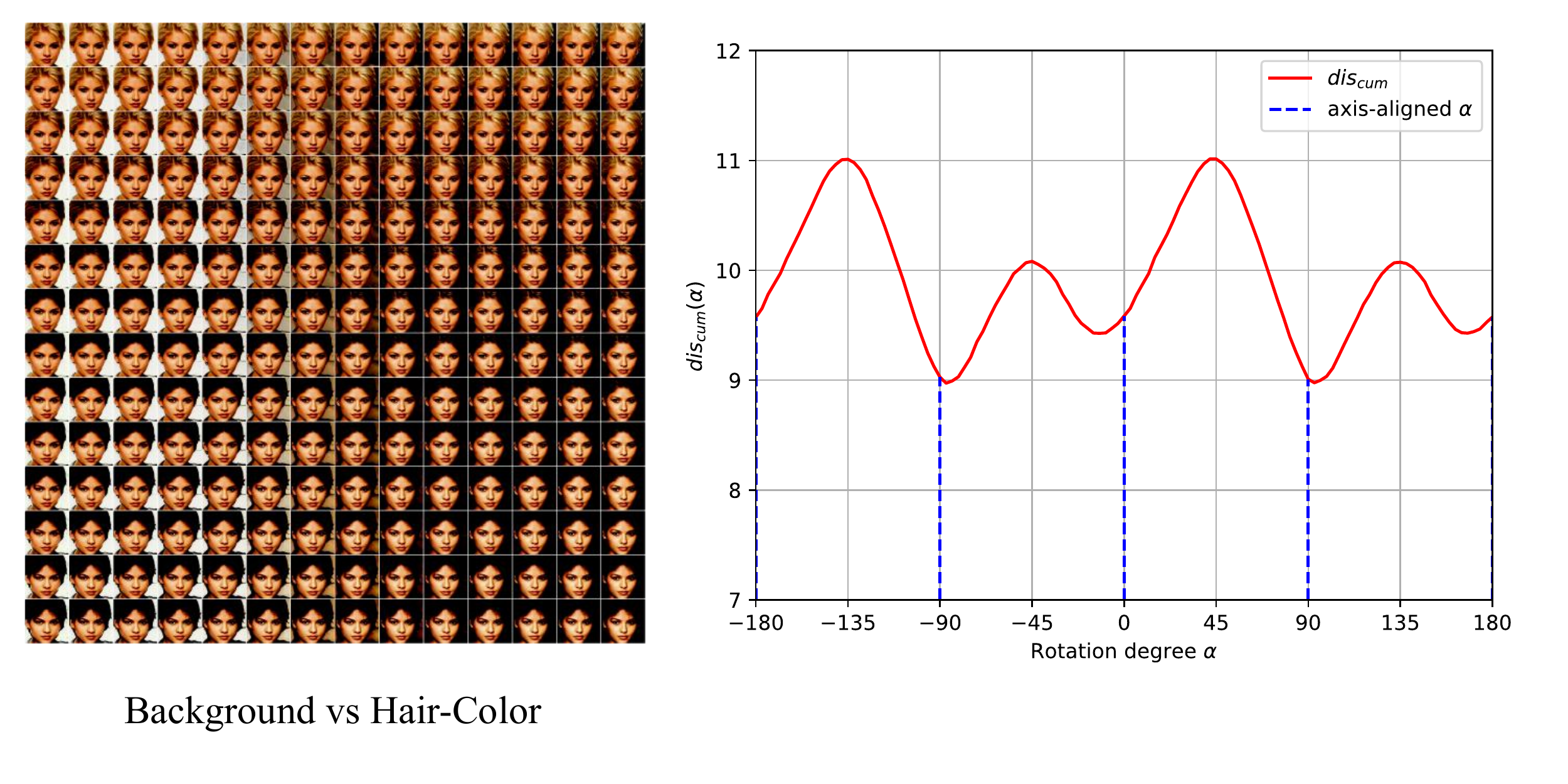}
\end{center}
    %\vspace{-8pt}
    \caption{Background vs Hair-Color.}
    %\vspace{-8pt}
\label{fig:background_vs_haircolor}
\end{figure}

The plot for different semantic-pairs on 3DShapes are shown in Fig.
\ref{fig:shape_vs_orientation} - \ref{fig:size_vs_wallcolor}.
Similar results for CelebA are shown in Fig. \ref{fig:gender_vs_smile} -
\ref{fig:background_vs_haircolor}.
For most semantic-pairs, the corresponding $\alpha$ rotation plots
agree with the assumption of Perceptual Simplicity, \ie the accumulated
perceptual distance scores become local minima when the accumulation
directions are aligned with the latent axes ($\alpha=-180, -90, 0, 90, 180$).
However, there exist still some exceptions, such as Shape vs Size,
and some attributes plotted against Azimuth, \etc.
These are (1) sometimes due to the imperfectly learned
representations which are
not fully disentangled, and (2) sometimes due to the domination of
variations encoded by one dimension over another (\eg Azimuth),
leading the Perceptual Simplicity phenomenon not obvious.
But in general, the assumption holds for most semantic-pairs.

\section{TPL Pros and Cons}
\label{ap:tpl_pros_cons}
\textbf{Pros}: 1) Unlike \cite{Duan2020AHF}, the TPL computation does not
rely on pair-wise comparisons among a herd of models (trained with
different hyper-parameters and seeds) to assign a score to a single model.
This ensures the TPL is a more efficient method for model selection,
and also enables its ability to work as a rough unsupervised metric
to evaluate disentanglement quality.
2) Unlike \cite{VPdis_eccv20}, the TPL does not need to train an extra
classifier to assign a score to a model, indicating it is a more
general and efficient approach.
3) Unlike \cite{Karras2020ASG}, the TPL leverage the perceptual anisotropy
in a disentangled representation, which can select more interpretable
ones than the PPL proposed in \cite{Karras2020ASG} which only
assign a score to a model by only evaluating the perceptual smoothness in the
latent space.

\textbf{Cons}: 1) The TPL scores can be biased by the generative ability
of the generator, \eg it usually ranks a blurred-image generator higher
than a more-detailed-image generator even their
disentanglement levels are similar.
To alleviate this problem, we recommend using TPL together with
some generative quality metrics (\eg FID, IS) to filter out the
\emph{cheating} models that achieve disentanglement at the severe cost of
generative quality.
2) The TPL is based on the assumption that disentangled representations
contain perceptually simple variations along their latent axes.
This assumption may not hold for every dataset or for every concept.
In those case, supervised disentanglement learning methods may be
more preferable.

\section{More Results of TPL Experiments}
\label{ap:tpl_other_experiments}
\begin{figure}[t]
    \begin{center}
        \includegraphics[width=\linewidth]{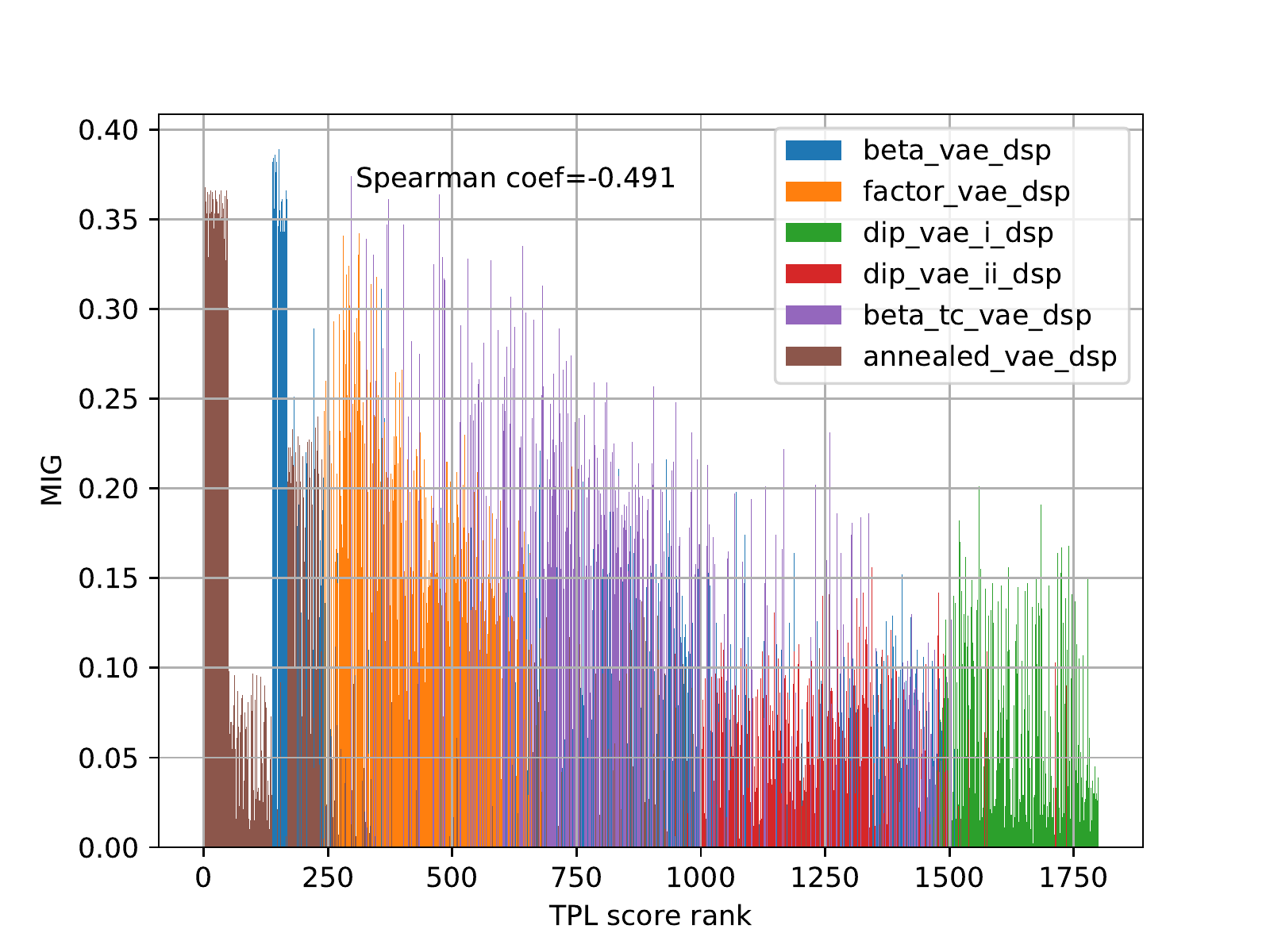}
        %\vspace{-20pt}
    \end{center}
    \caption{TPL (act$>$0) vs MIG. Ranked by TPL scores.}
    %\vspace{-10pt}
    \label{fig:tpl_0_vs_mig}
\end{figure}

\begin{figure}[t]
    \begin{center}
        \includegraphics[width=\linewidth]{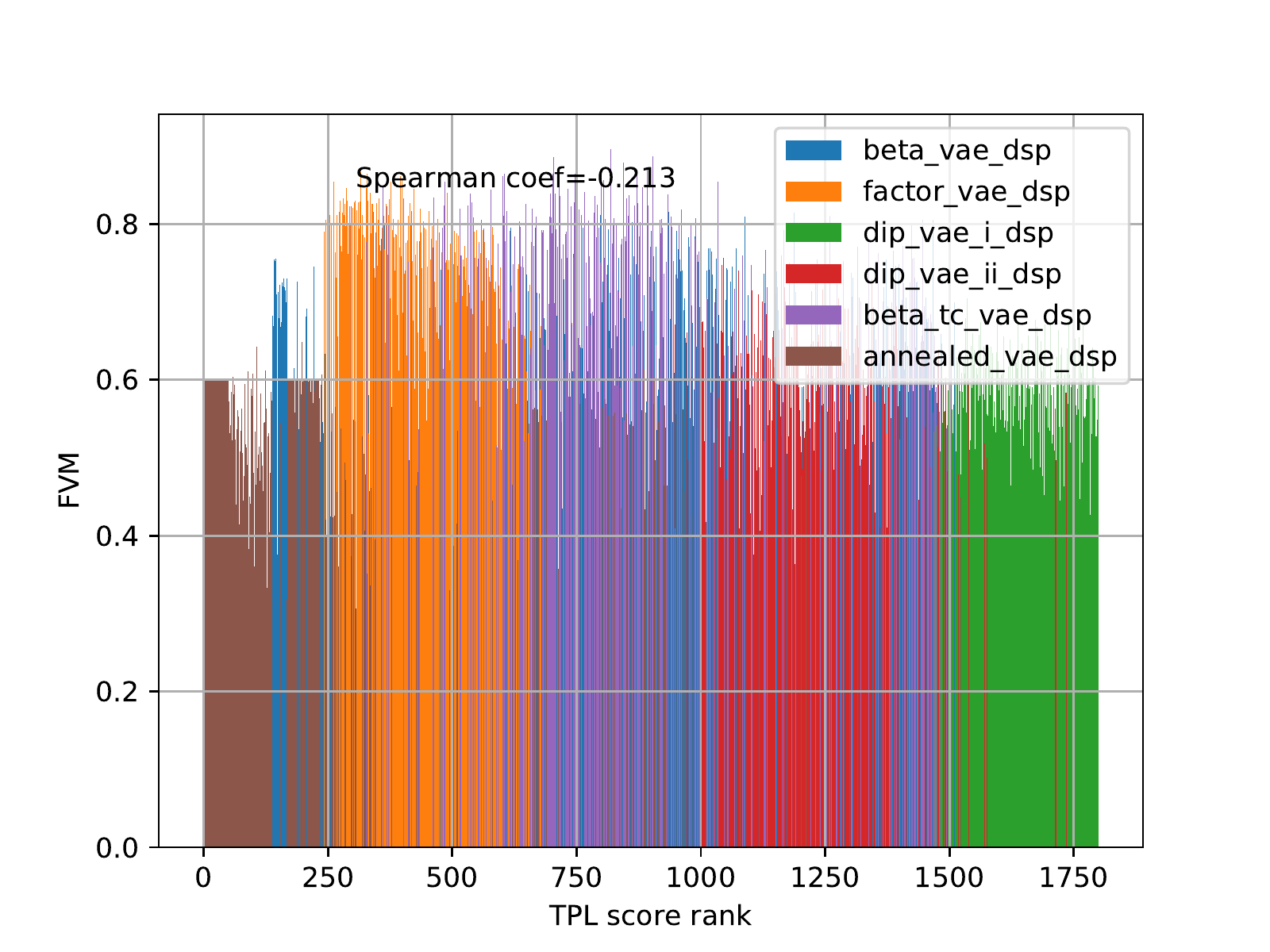}
        %\vspace{-20pt}
    \end{center}
    \caption{TPL (act$>$0) vs FVM. Ranked by TPL scores.}
    %\vspace{-10pt}
    \label{fig:tpl_0_vs_fvm}
\end{figure}

\begin{figure}[t]
    \begin{center}
        \includegraphics[width=\linewidth]{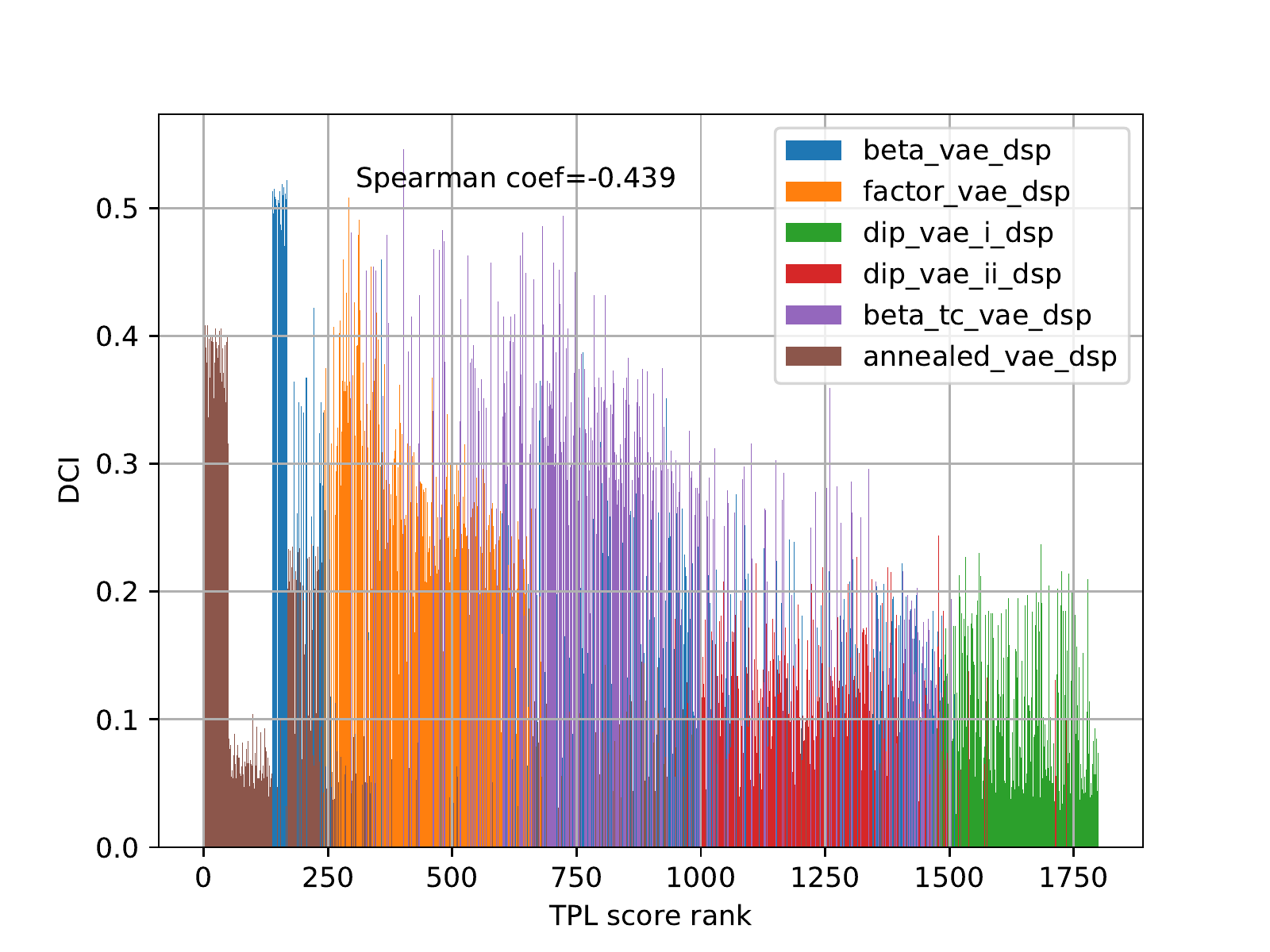}
        %\vspace{-20pt}
    \end{center}
    \caption{TPL (act$>$0) vs DCI. Ranked by TPL scores.}
    %\vspace{-10pt}
    \label{fig:tpl_0_vs_dci}
\end{figure}

\begin{figure}[t]
    \begin{center}
        \includegraphics[width=\linewidth]{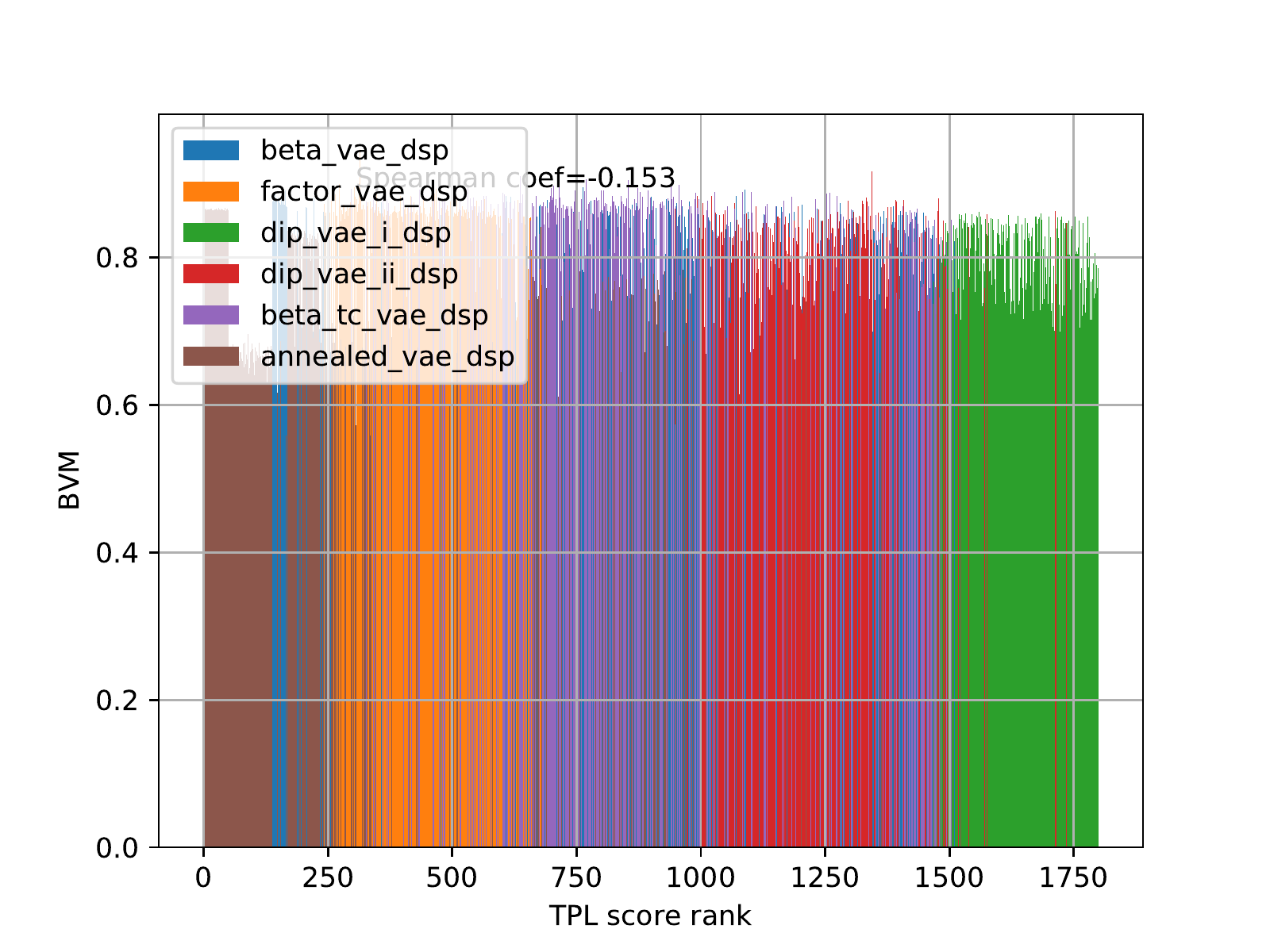}
        %\vspace{-20pt}
    \end{center}
    \caption{TPL (act$>$0) vs BVM. Ranked by TPL scores.}
    %\vspace{-10pt}
    \label{fig:tpl_0_vs_bvm}
\end{figure}

\begin{figure}[t]
    \begin{center}
        \includegraphics[width=\linewidth]{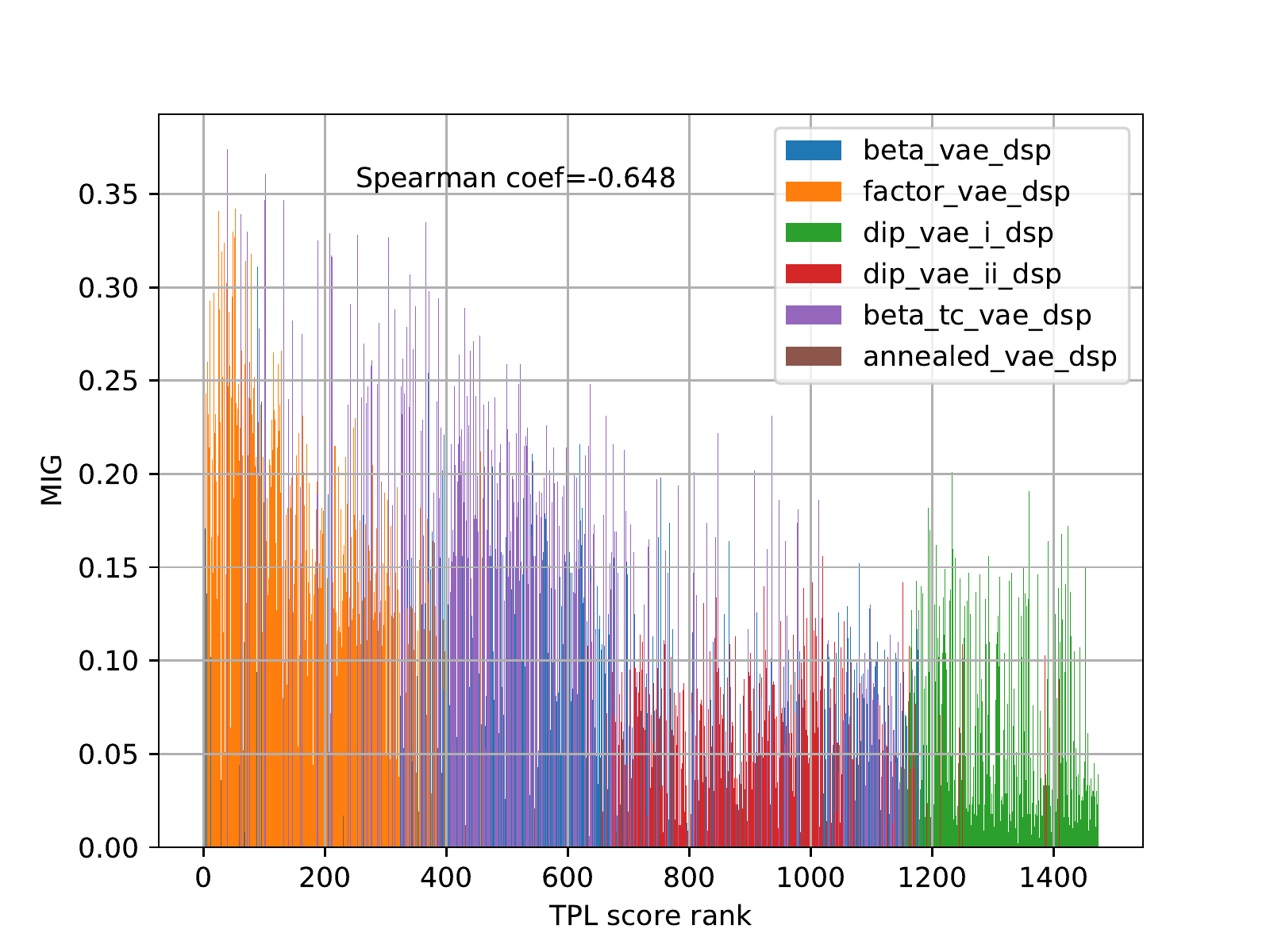}
        %\vspace{-20pt}
    \end{center}
    \caption{TPL (act$>$3) vs MIG. Ranked by TPL scores.}
    %\vspace{-10pt}
    \label{fig:tpl_3_vs_mig}
\end{figure}

\begin{figure}[t]
    \begin{center}
        \includegraphics[width=\linewidth]{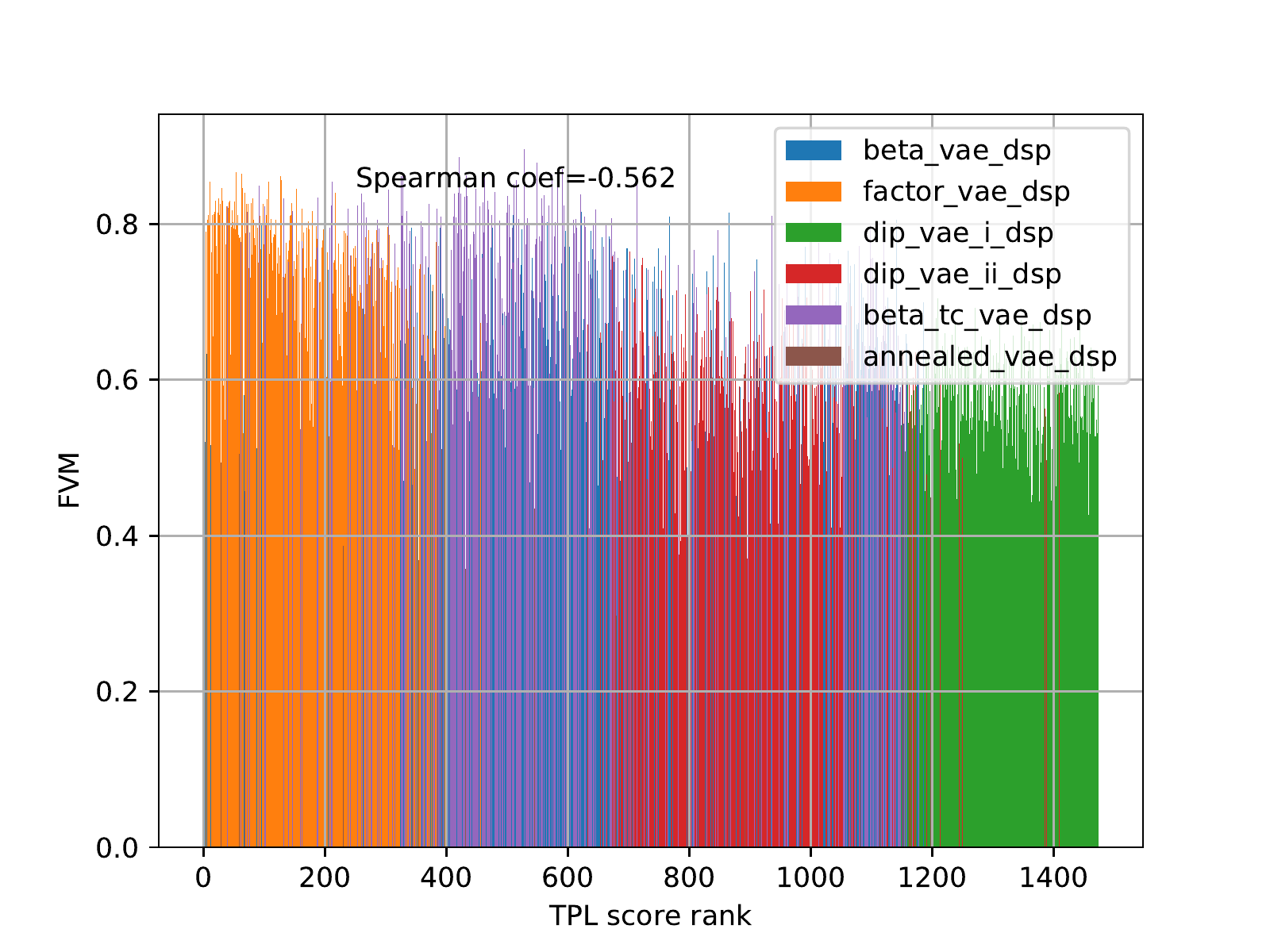}
        %\vspace{-20pt}
    \end{center}
    \caption{TPL (act$>$3) vs FVM. Ranked by TPL scores.}
    %\vspace{-10pt}
    \label{fig:tpl_3_vs_fvm}
\end{figure}

\begin{figure}[t]
    \begin{center}
        \includegraphics[width=\linewidth]{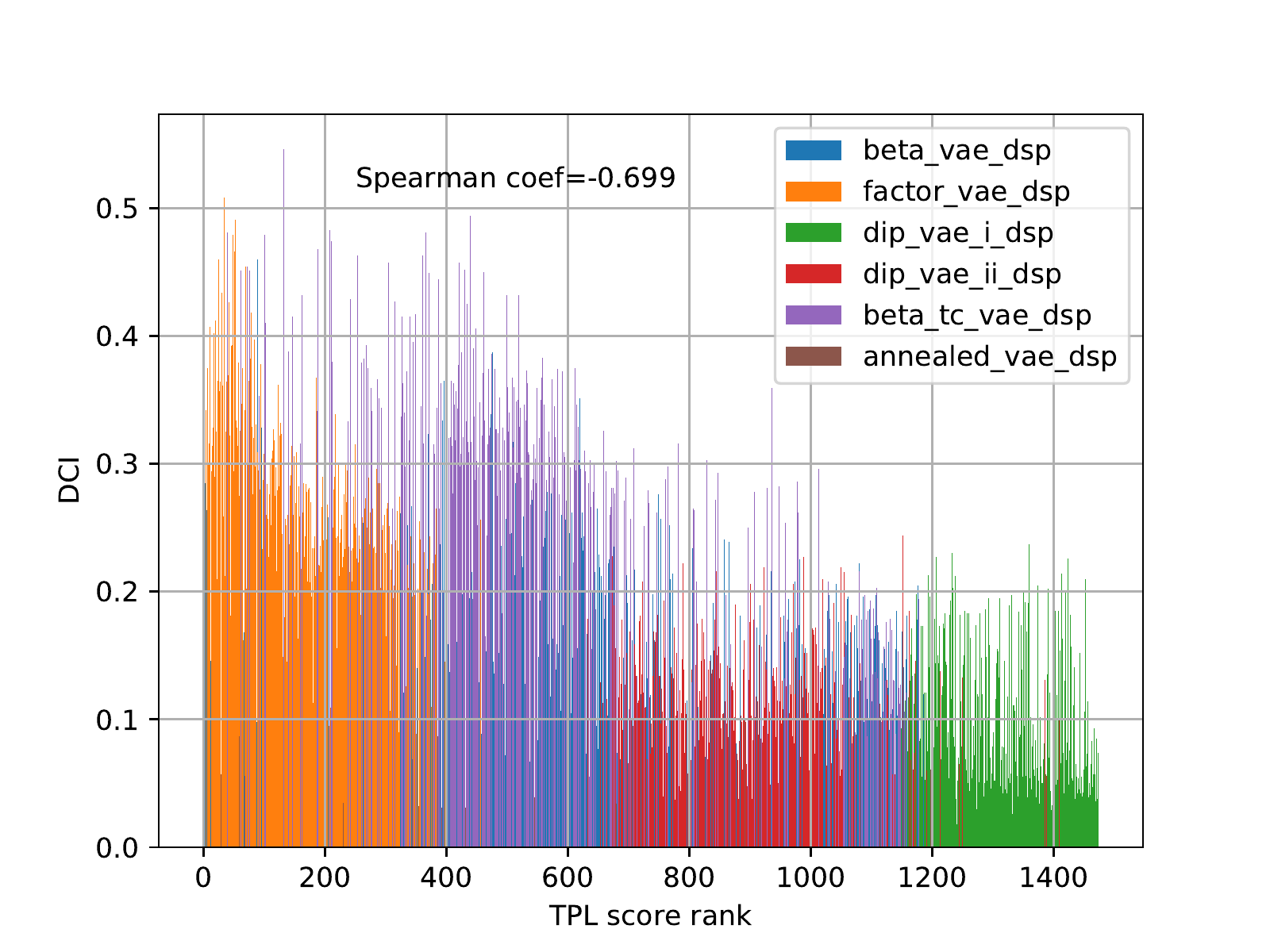}
        %\vspace{-20pt}
    \end{center}
    \caption{TPL (act$>$3) vs DCI. Ranked by TPL scores.}
    %\vspace{-10pt}
    \label{fig:tpl_3_vs_dci}
\end{figure}

\begin{figure}[t]
    \begin{center}
        \includegraphics[width=\linewidth]{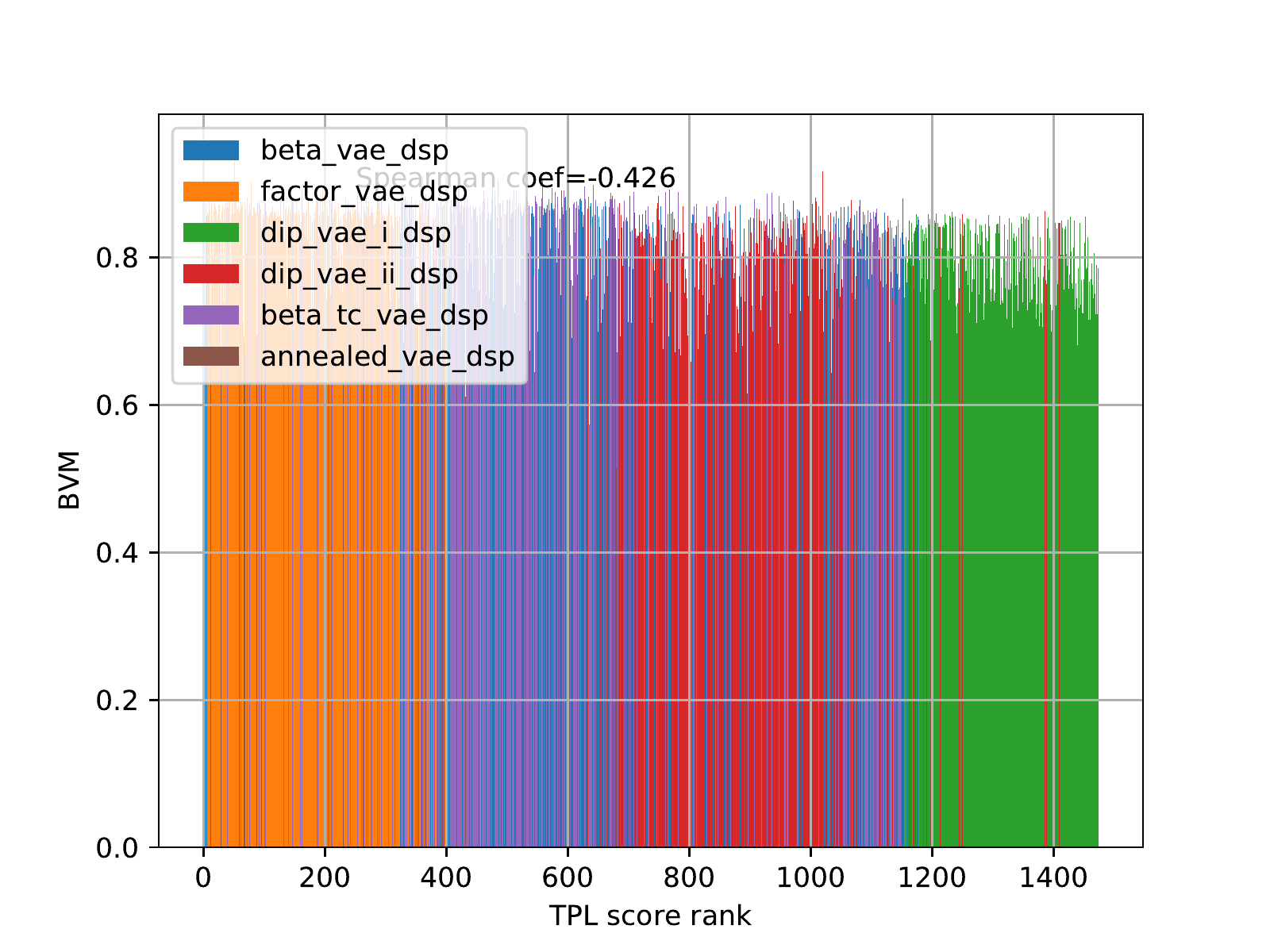}
        %\vspace{-20pt}
    \end{center}
    \caption{TPL (act$>$3) vs BVM. Ranked by TPL scores.}
    %\vspace{-10pt}
    \label{fig:tpl_3_vs_bvm}
\end{figure}

\begin{figure}[t]
    \begin{center}
        \includegraphics[width=\linewidth]{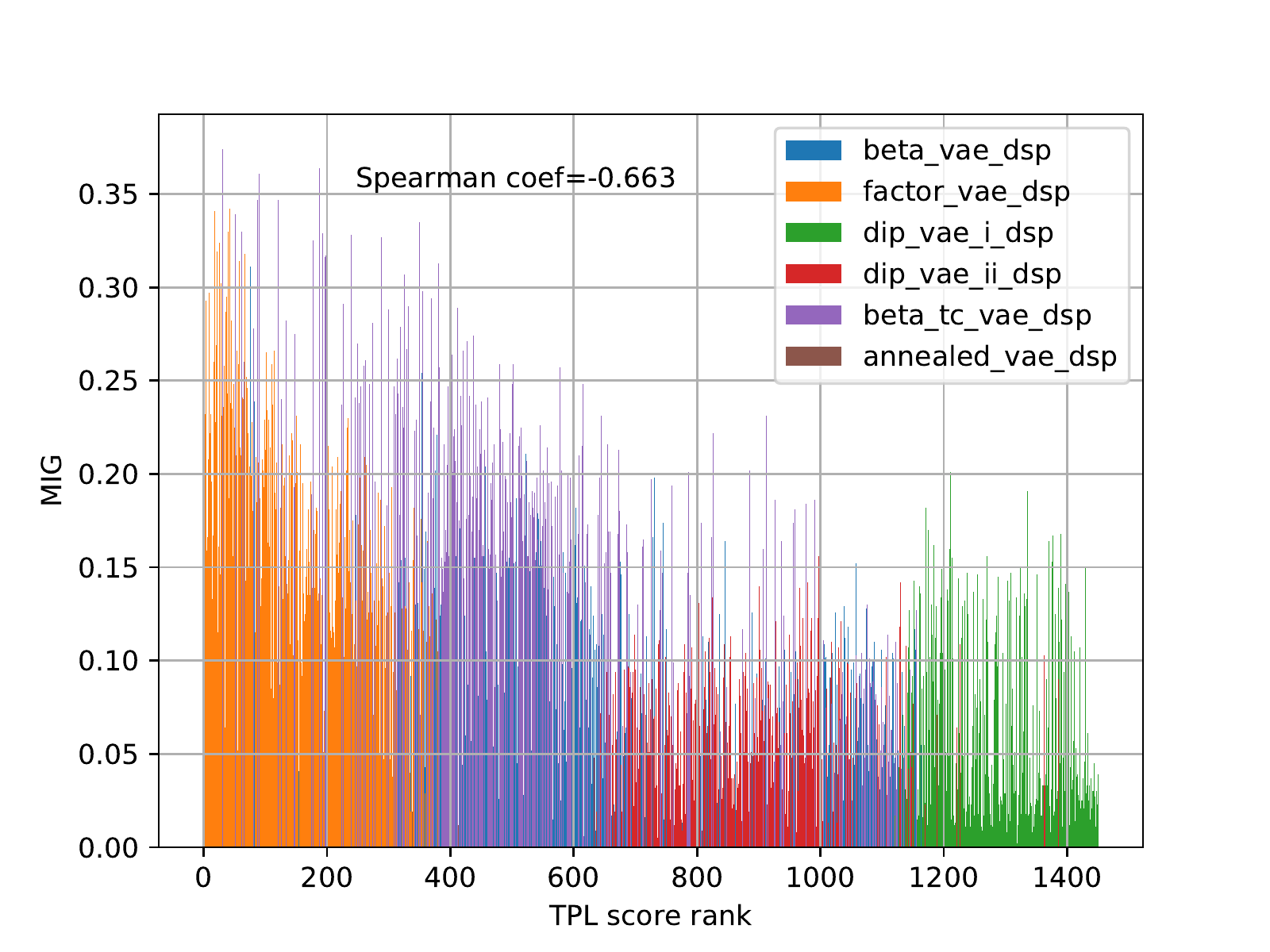}
        %\vspace{-20pt}
    \end{center}
    \caption{TPL (act$>$4) vs MIG. Ranked by TPL scores.}
    %\vspace{-10pt}
    \label{fig:tpl_4_vs_mig}
\end{figure}

\begin{figure}[t]
    \begin{center}
        \includegraphics[width=\linewidth]{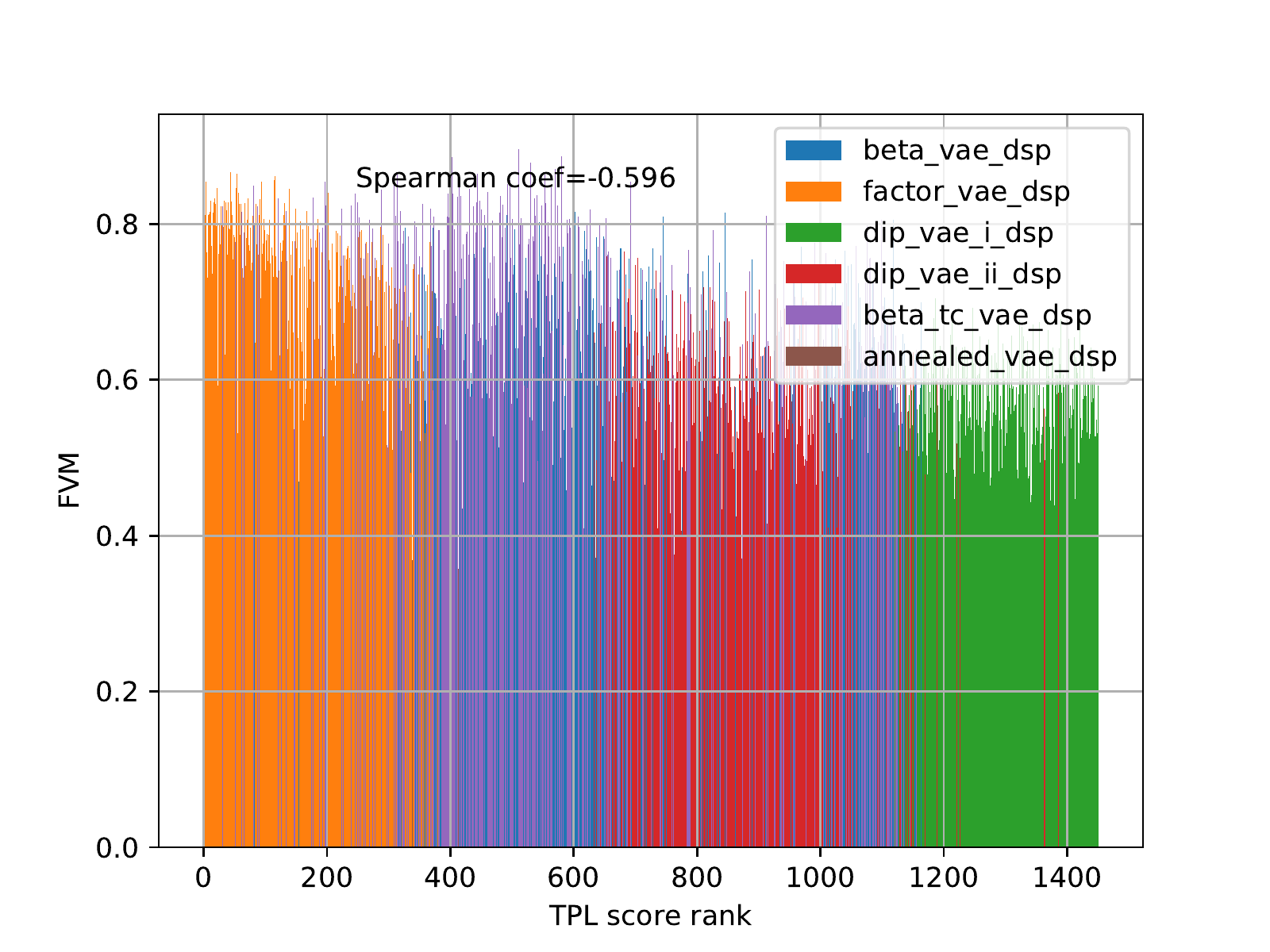}
        %\vspace{-20pt}
    \end{center}
    \caption{TPL (act$>$4) vs FVM. Ranked by TPL scores.}
    %\vspace{-10pt}
    \label{fig:tpl_4_vs_fvm}
\end{figure}

\begin{figure}[t]
    \begin{center}
        \includegraphics[width=\linewidth]{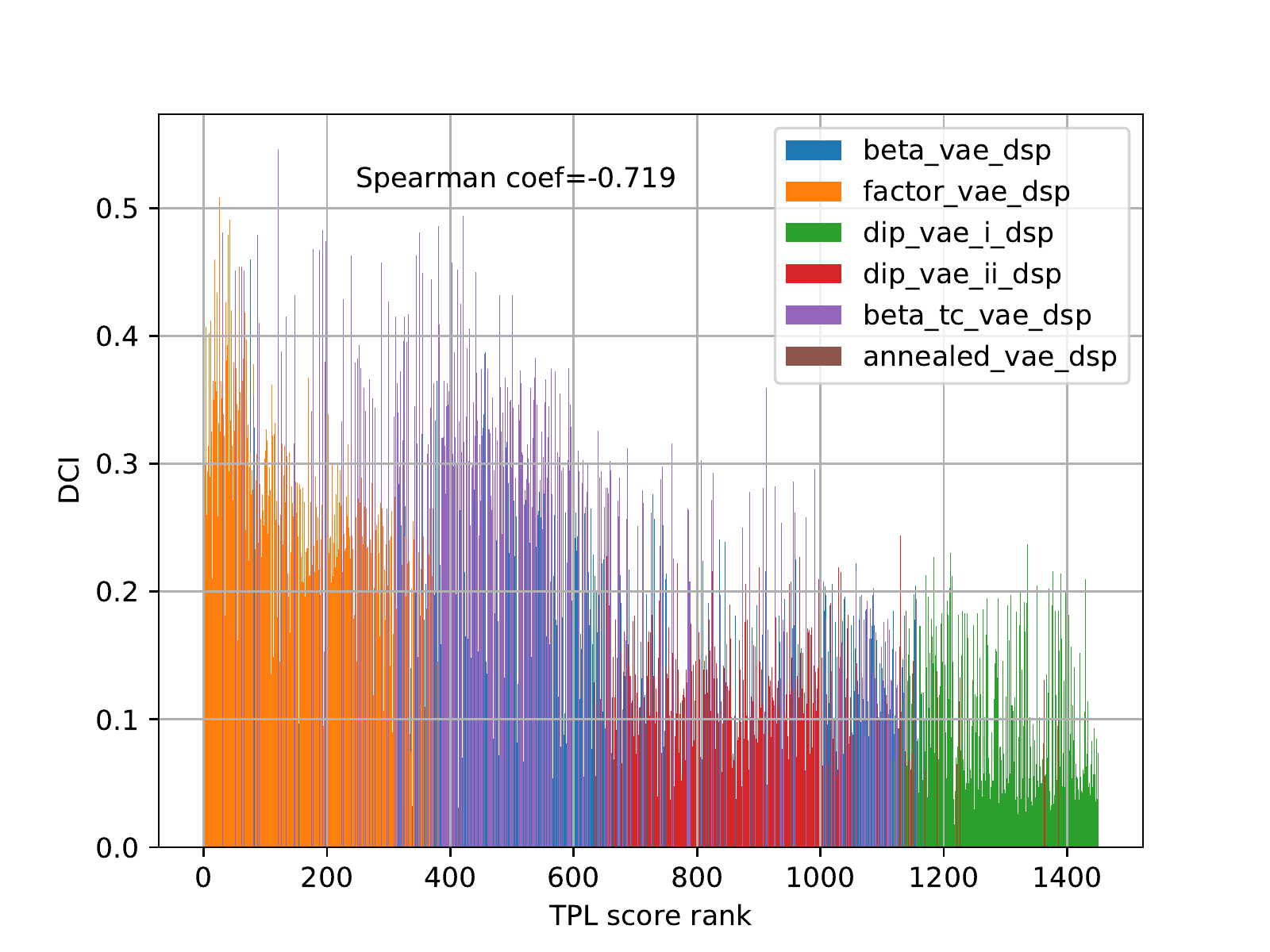}
        %\vspace{-20pt}
    \end{center}
    \caption{TPL (act$>$4) vs DCI. Ranked by TPL scores.}
    %\vspace{-10pt}
    \label{fig:tpl_4_vs_dci}
\end{figure}

\begin{figure}[t]
    \begin{center}
        \includegraphics[width=\linewidth]{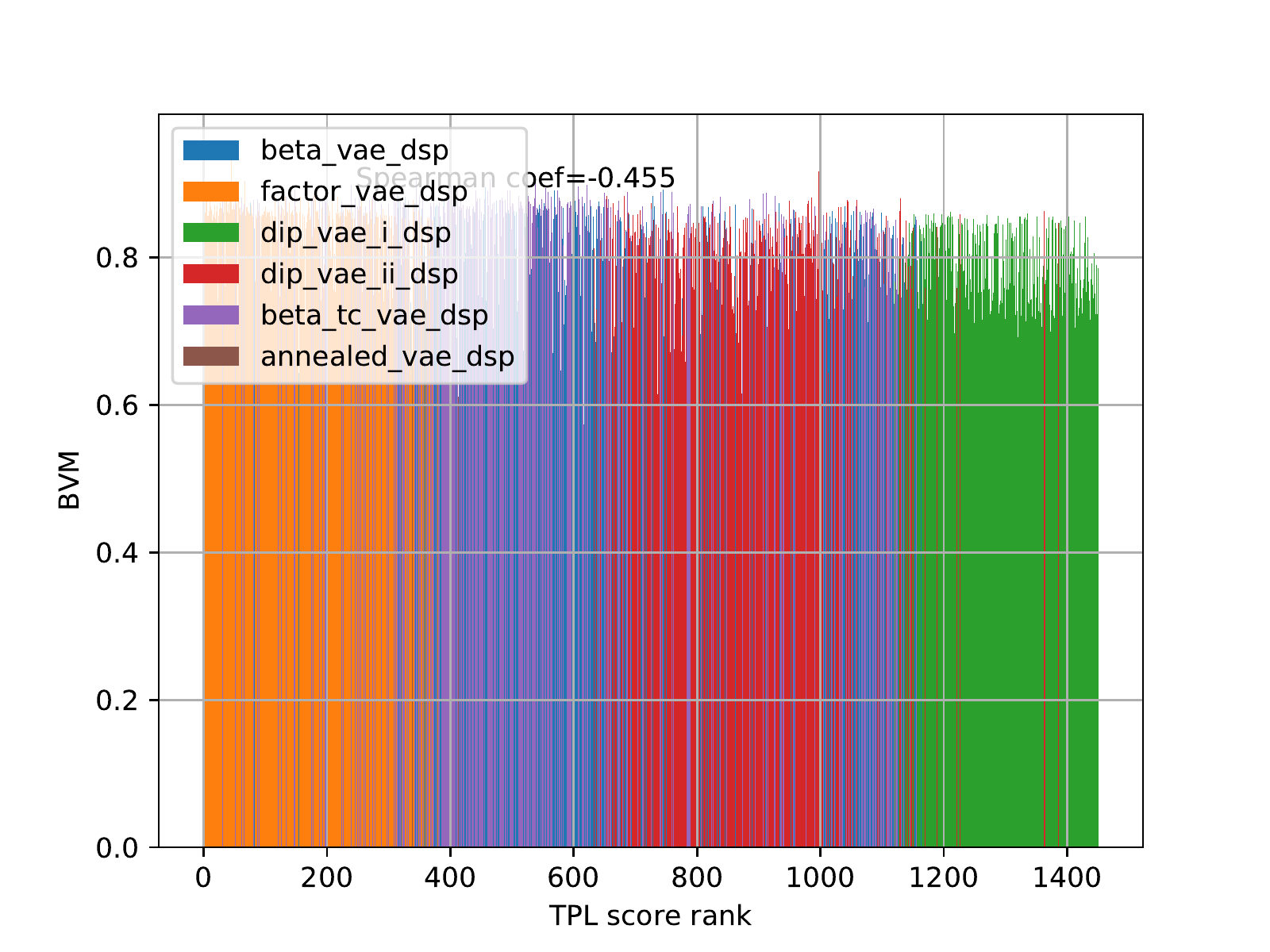}
        %\vspace{-20pt}
    \end{center}
    \caption{TPL (act$>$4) vs BVM. Ranked by TPL scores.}
    %\vspace{-10pt}
    \label{fig:tpl_4_vs_bvm}
\end{figure}

Here we show more correlation plots for more metrics and the dimensions
using the pretrained checkpoints.
The correlation coefficients for each setting are shown in the plots.
In Fig. \ref{fig:tpl_0_vs_mig}, \ref{fig:tpl_0_vs_fvm}, \ref{fig:tpl_0_vs_dci},
and \ref{fig:tpl_0_vs_bvm}, we show the plots with all dimensions of
activation (act$>$0) taken into account.
There are descending trends shown in each figure, but the reason that
the correlation scores are low is due to the left most samples (around $<$250)
shown in each plot. These samples are strangely positioned at the top
of the whole rank by TPL, but are generally not scored high by
supervised metrics. This is because these samples encode only a
subset of factors in the dataset (3 out of 5), and some of them are
detected by the supervised metrics and are assigned with low scores.
In Fig. \ref{fig:tpl_3_vs_mig}, \ref{fig:tpl_3_vs_fvm}, \ref{fig:tpl_3_vs_dci},
\ref{fig:tpl_3_vs_bvm}, \ref{fig:tpl_4_vs_mig}, \ref{fig:tpl_4_vs_fvm},
\ref{fig:tpl_4_vs_dci} and \ref{fig:tpl_4_vs_bvm}, we show the plots of
samples with active dimensions larger than 3 and 4. In these figures,
the TPL ranks these models much better, and the correlation coefficients
also agree with the descending trends better in these plots.

\section{Ablation Study on the Number of Rectangles $J$}
\label{ap:n_rectangles}
We show the impact of the number of rectangles used in our SC modules
on the CelebA dataset. We vary $J$ from 1 to 9 while keeping all other
factors unchanged. The results are shown in Table \ref{table:ablation_J}.
We can see using
too few rectangles harms FID (variation controlled by each code is
too simple) while using too many harms TPL (each code controls
entangled variations). A balanced $J$ is about 3$\sim$6.

\begin{table}[t]
    \begin{center}
        %\small
        \begin{tabular}{lccc}
            %\toprule
            \hline
            $J$ & TPL & PPL & FID \\
            %\midrule
            \hline
            1 & \textbf{5.4} & \textbf{29.2} & 9.0 \\
            3 & 7.0 & 35.8 & 7.7 \\
            4 & 7.1 & 38.4 & \textbf{5.9} \\
            6 & 8.1 & 38.9 & 6.0 \\
            9 & 8.2 & 40.0 & 6.2 \\
            %\bottomrule
            \hline
            %\hline
        \end{tabular}
    \end{center}
    \caption{Ablation on $J$ on CelebA dataset.}
    \label{table:ablation_J}
\end{table}

\section{More Qualitative Results}
\label{ap:more_qualitative_results}
\begin{figure}[t]
\begin{center}
   \includegraphics[width=\linewidth]{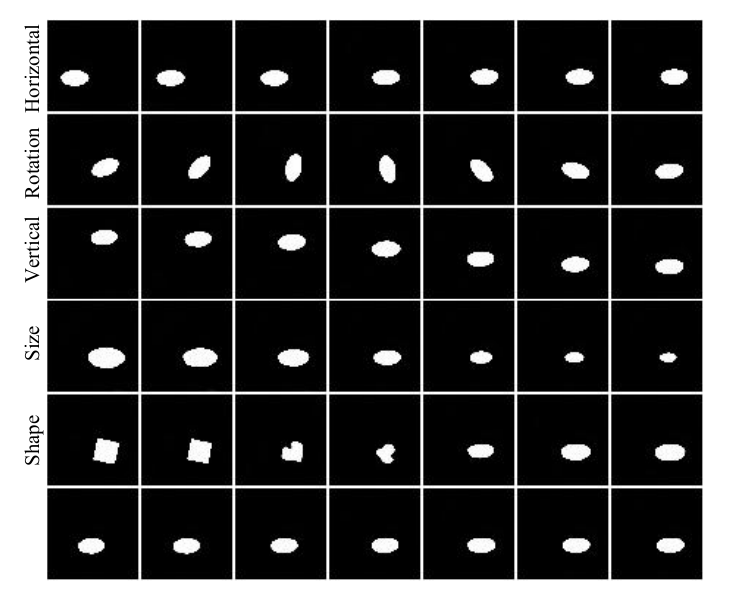}
\end{center}
    %\vspace{-8pt}
    \caption{DSprites 1.}
    %\vspace{-8pt}
\label{fig:dsprites_travs_1}
\end{figure}

\begin{figure}[t]
\begin{center}
   \includegraphics[width=\linewidth]{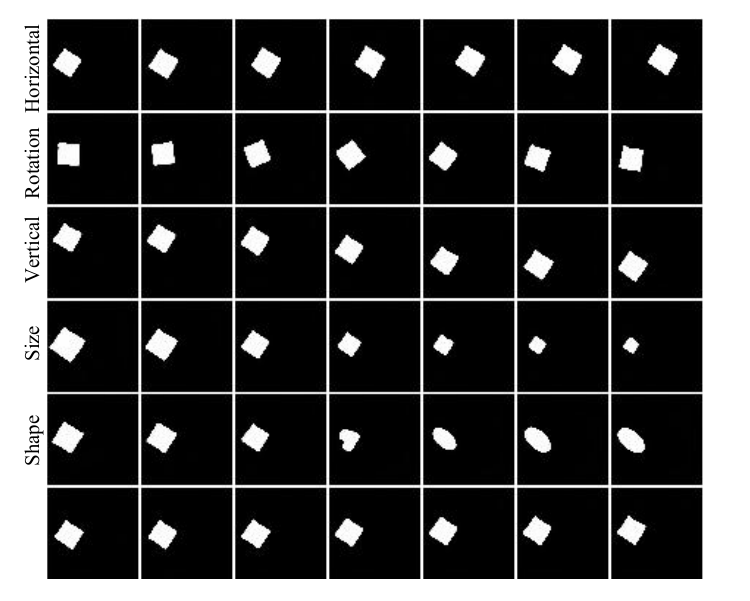}
\end{center}
    %\vspace{-8pt}
    \caption{DSprites 2.}
    %\vspace{-8pt}
\label{fig:dsprites_travs_2}
\end{figure}

\begin{figure}[t]
\begin{center}
   \includegraphics[width=\linewidth]{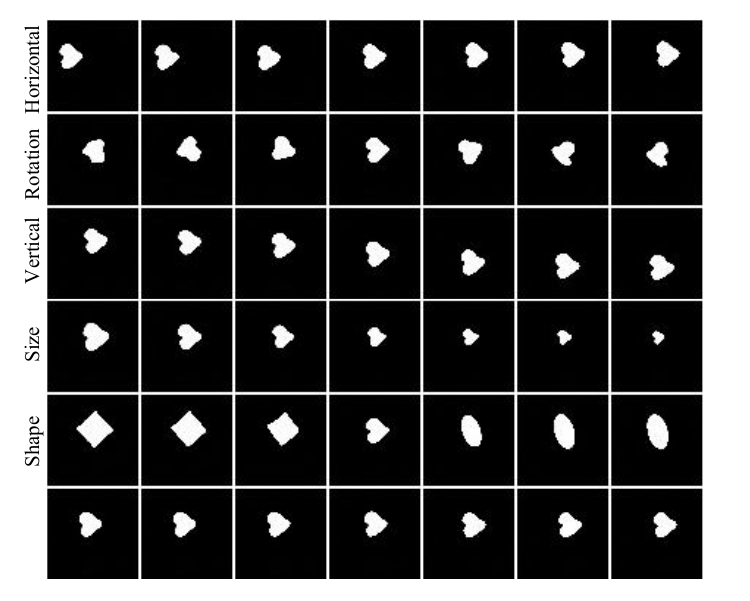}
\end{center}
    %\vspace{-8pt}
    \caption{DSprites 3.}
    %\vspace{-8pt}
\label{fig:dsprites_travs_3}
\end{figure}

\begin{figure}[t]
\begin{center}
   \includegraphics[width=\linewidth]{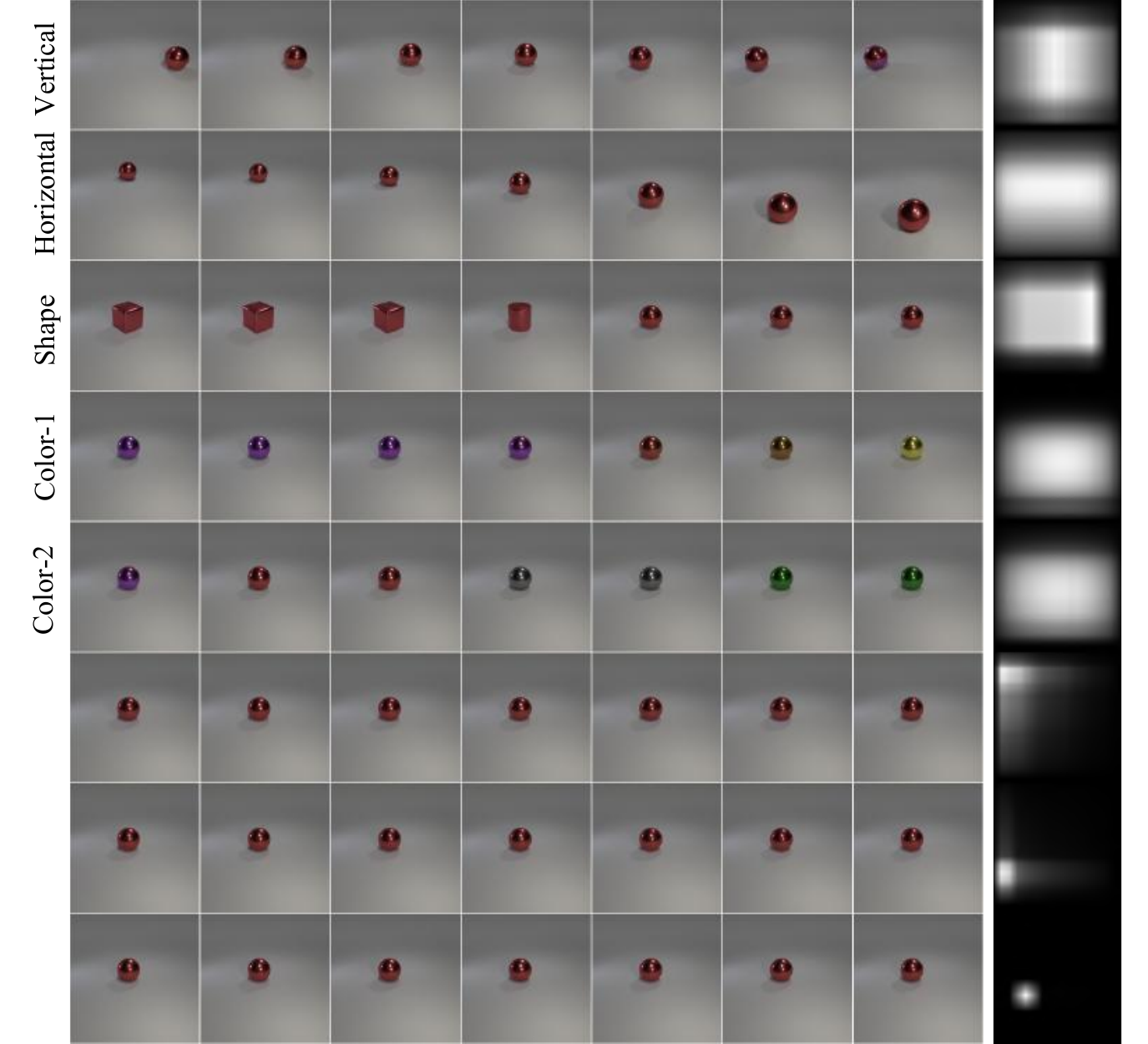}
\end{center}
    %\vspace{-8pt}
    \caption{Clevr-Simple 3.}
    %\vspace{-8pt}
\label{fig:clevr_simple_travs_3}
\end{figure}

\begin{figure}[t]
\begin{center}
   \includegraphics[width=\linewidth]{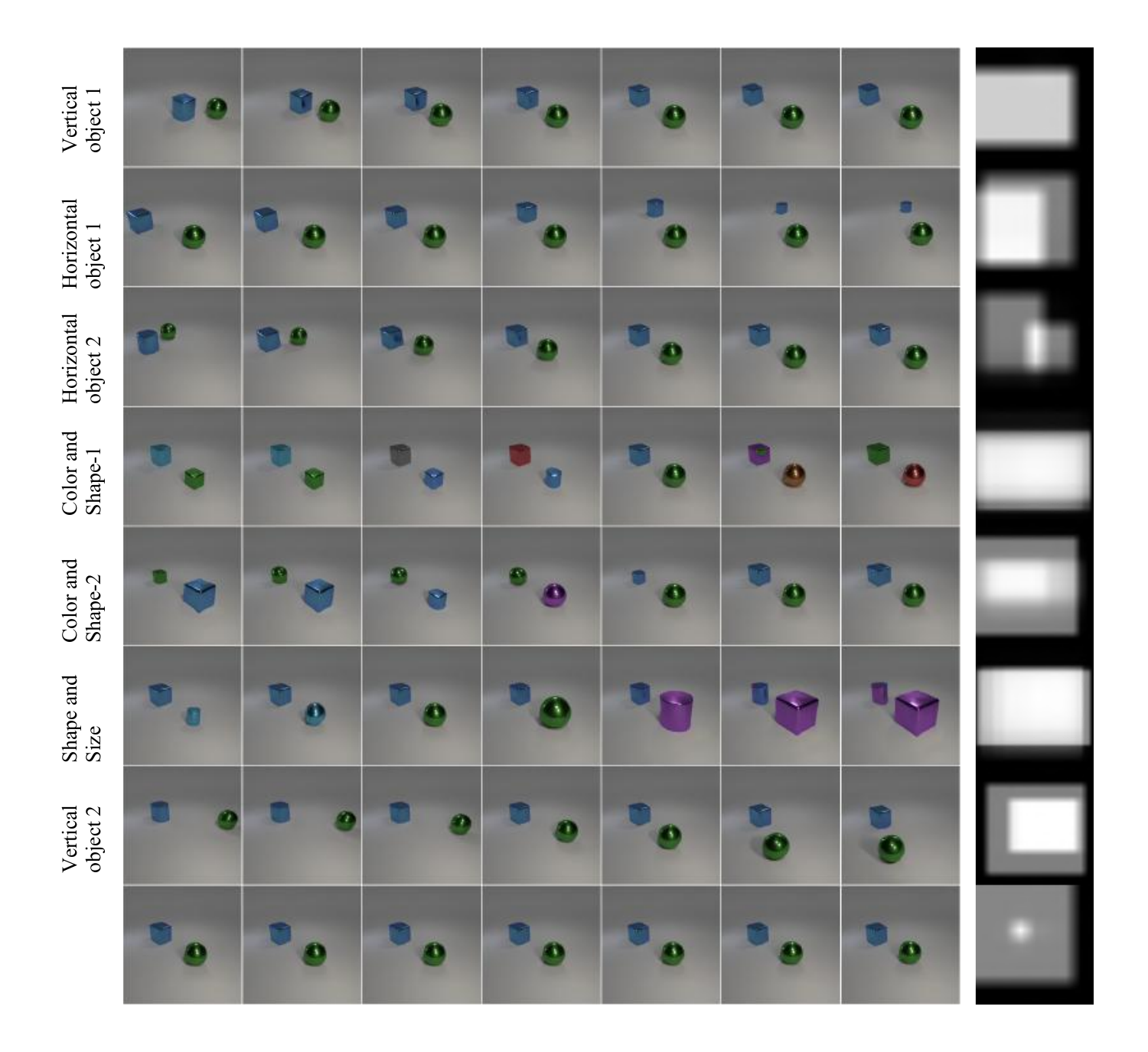}
\end{center}
    %\vspace{-8pt}
    \caption{Clevr-Complex 1.}
    %\vspace{-8pt}
\label{fig:clevr_comp_travs_1}
\end{figure}

\begin{figure}[t]
\begin{center}
   \includegraphics[width=\linewidth]{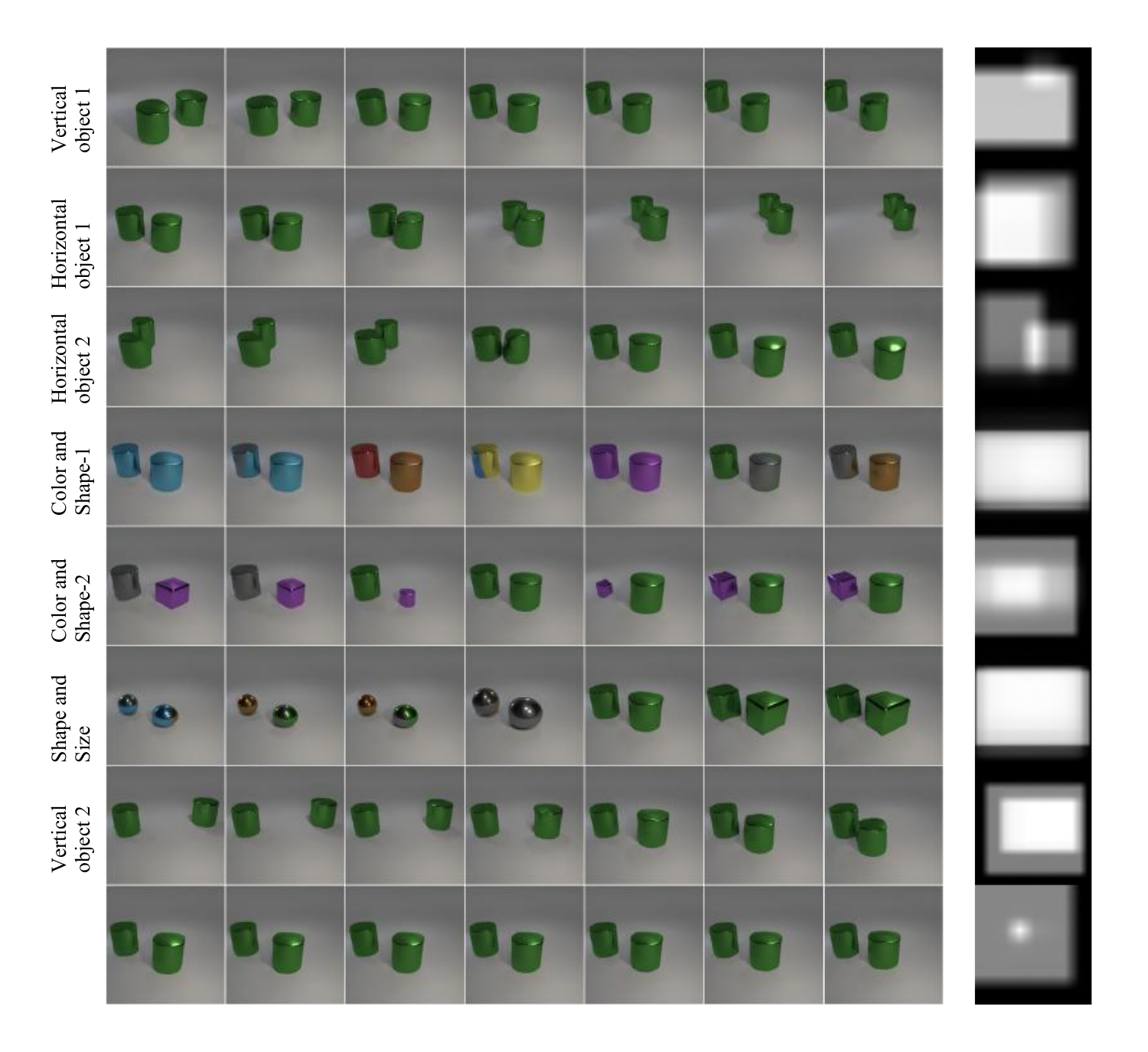}
\end{center}
    %\vspace{-8pt}
    \caption{Clevr-Complex 2.}
    %\vspace{-8pt}
\label{fig:clevr_comp_travs_2}
\end{figure}

\begin{figure}[t]
\begin{center}
   \includegraphics[width=\linewidth]{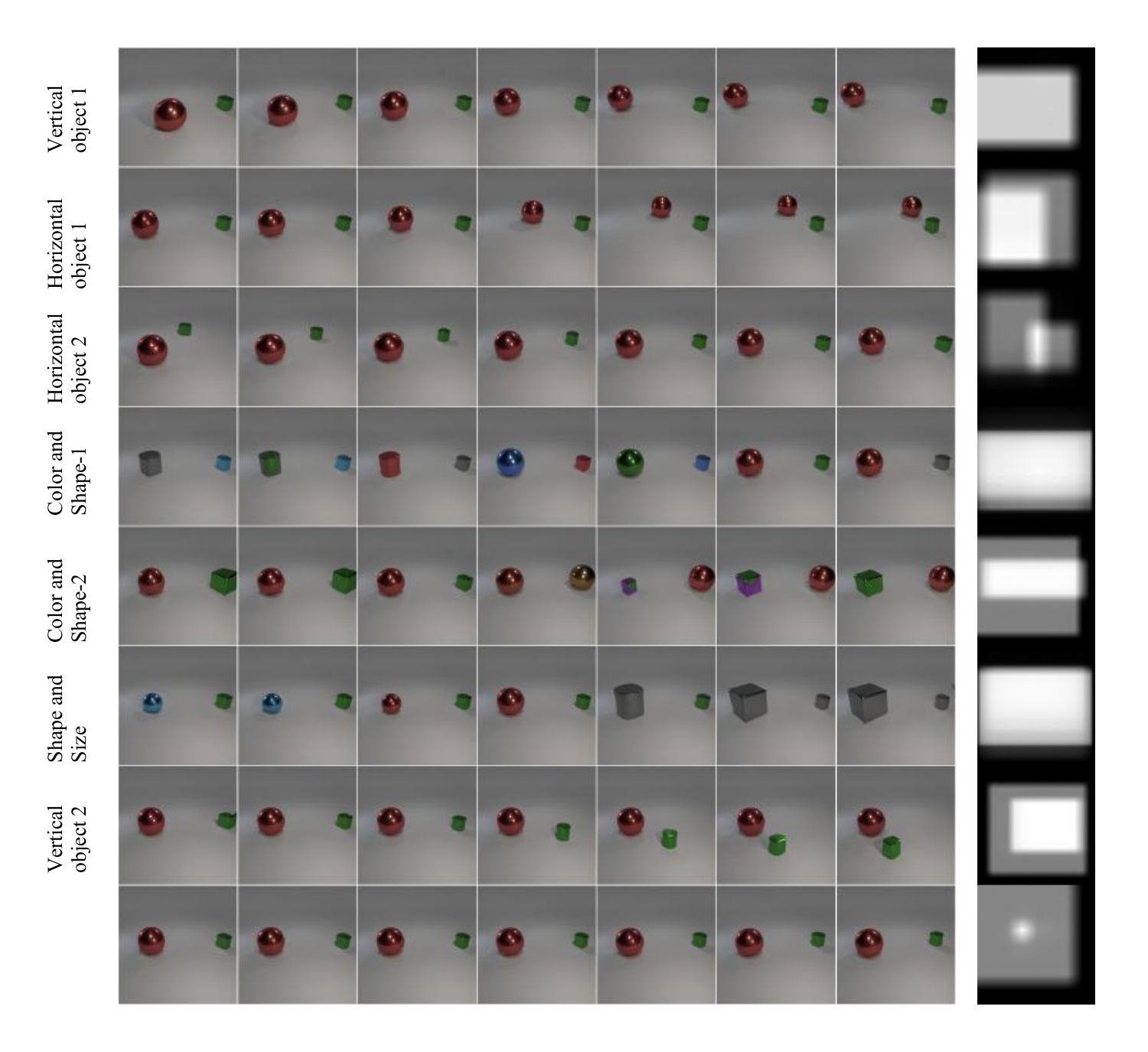}
\end{center}
    %\vspace{-8pt}
    \caption{Clevr-Complex 3.}
    %\vspace{-8pt}
\label{fig:clevr_comp_travs_3}
\end{figure}

\begin{figure}[t]
\begin{center}
   \includegraphics[width=\linewidth]{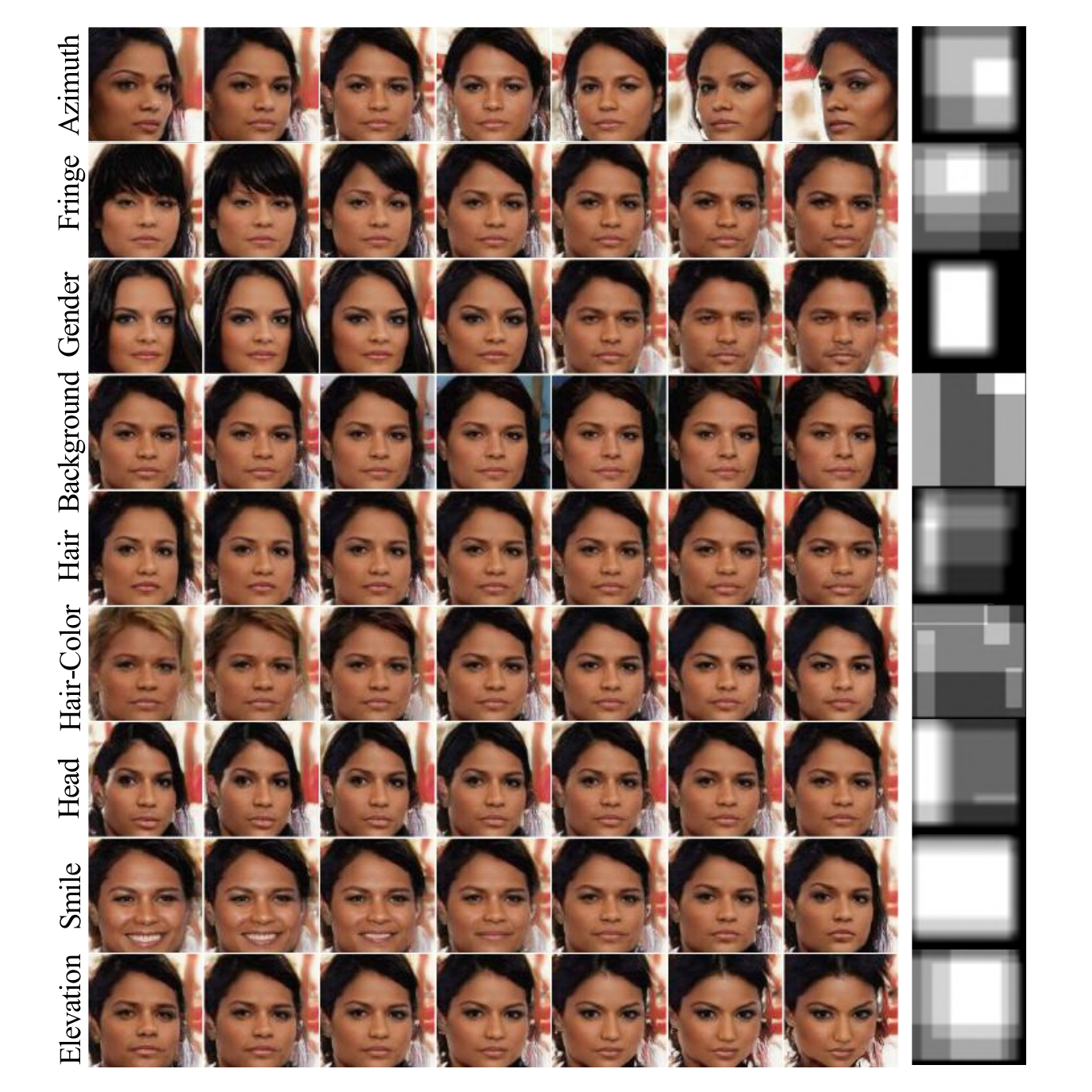}
\end{center}
    %\vspace{-8pt}
    \caption{CelebA 1.}
    %\vspace{-8pt}
\label{fig:celeba_travs_1}
\end{figure}

\begin{figure}[t]
\begin{center}
   \includegraphics[width=\linewidth]{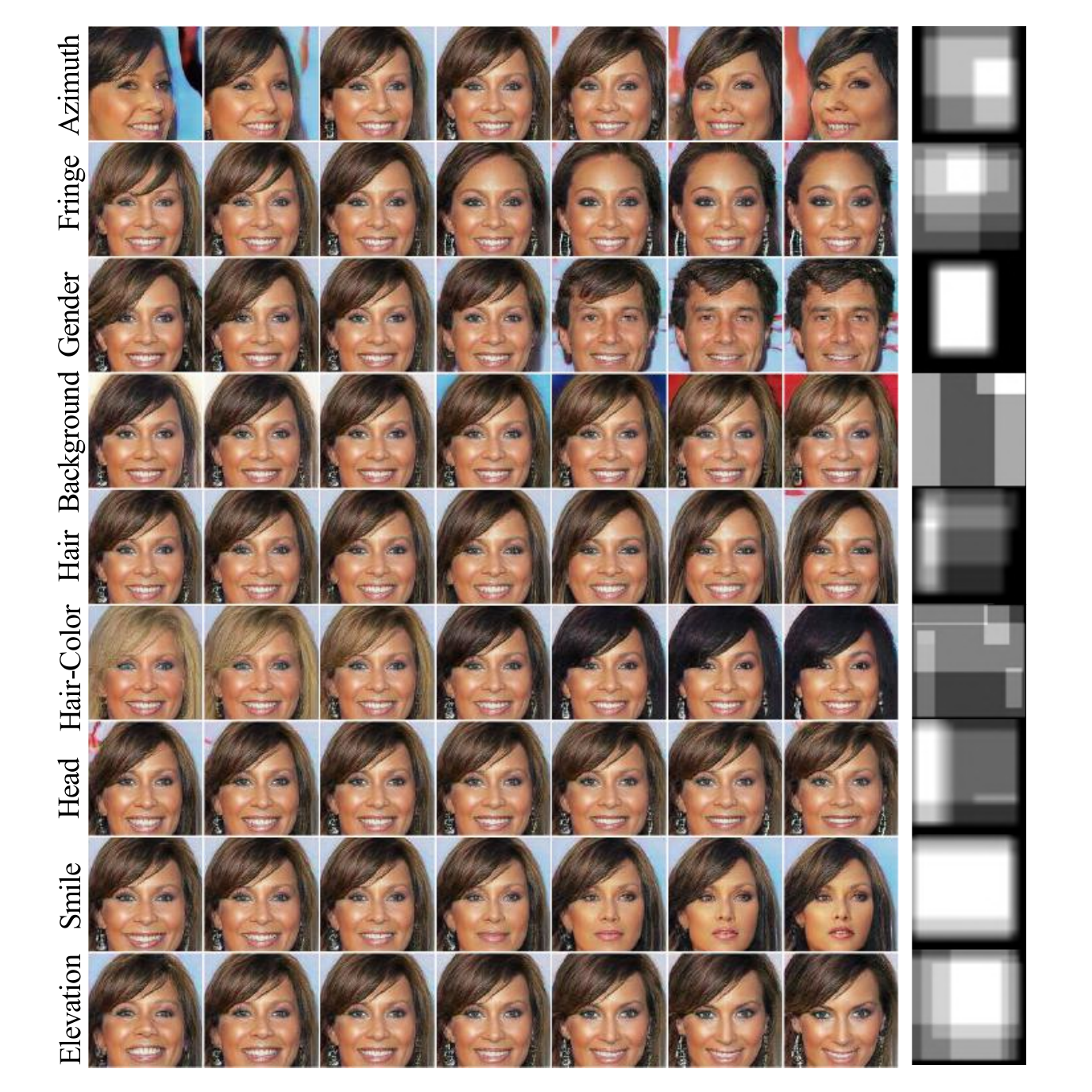}
\end{center}
    %\vspace{-8pt}
    \caption{CelebA 2.}
    %\vspace{-8pt}
\label{fig:celeba_travs_2}
\end{figure}

\begin{figure}[t]
\begin{center}
   \includegraphics[width=\linewidth]{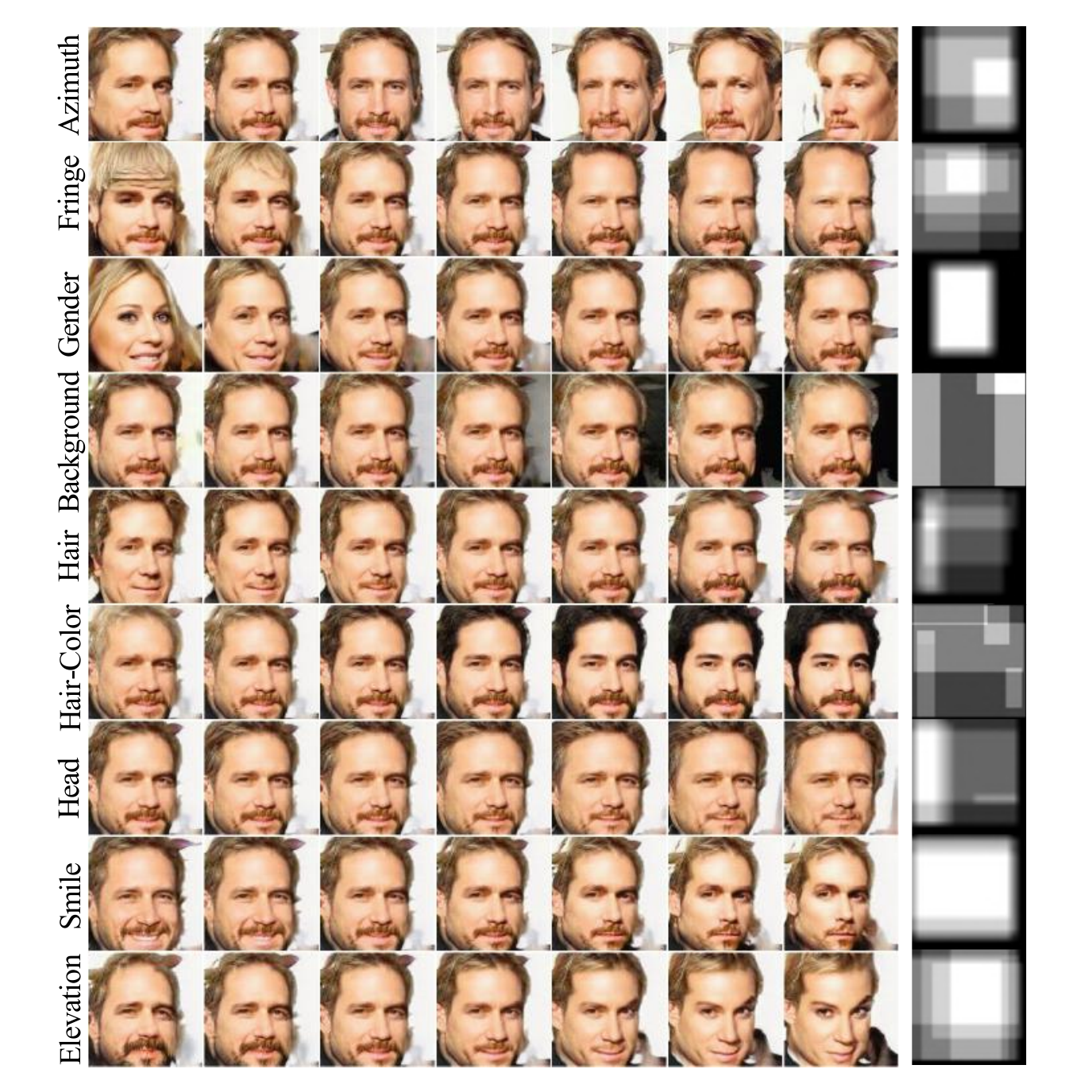}
\end{center}
    %\vspace{-8pt}
    \caption{CelebA 3.}
    %\vspace{-8pt}
\label{fig:celeba_travs_3}
\end{figure}

More traversals are shown in Fig. \ref{fig:dsprites_travs_1} - Fig.
\ref{fig:celeba_travs_3}.
For Clevr-Complex, we can see our model learns the position concepts
for both objects, but other concepts like size and shape of objects
are still entangled. This indicates the multi-object disentangled
representation learning is still a hard and unsolved problem.

\begin{figure}[t]
\begin{center}
   \includegraphics[width=\linewidth]{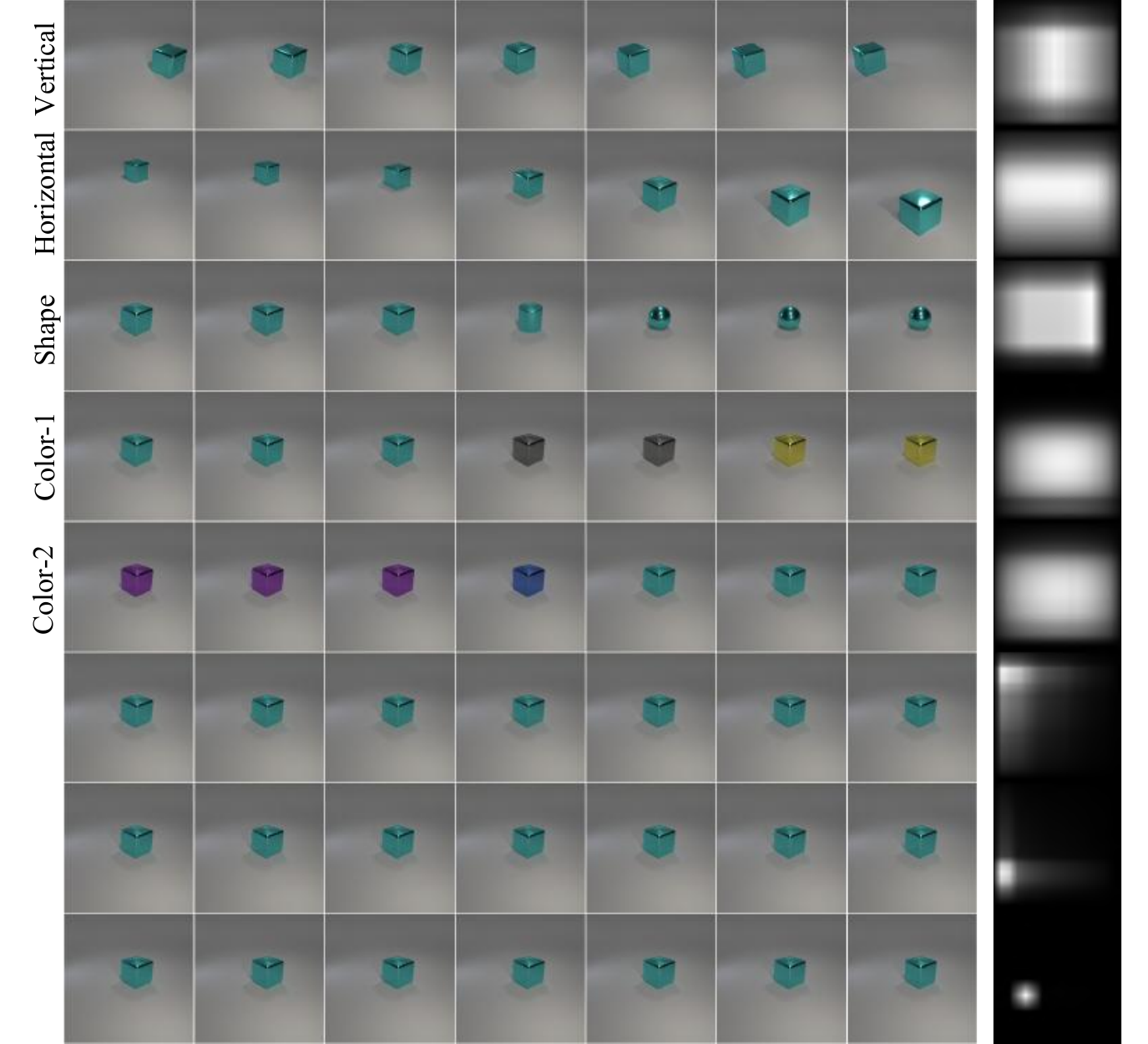}
\end{center}
    %\vspace{-8pt}
    \caption{Clevr-Simple 1.}
    %\vspace{-8pt}
\label{fig:clevr_simple_travs_1}
\end{figure}
\begin{figure}[t]
\begin{center}
   \includegraphics[width=\linewidth]{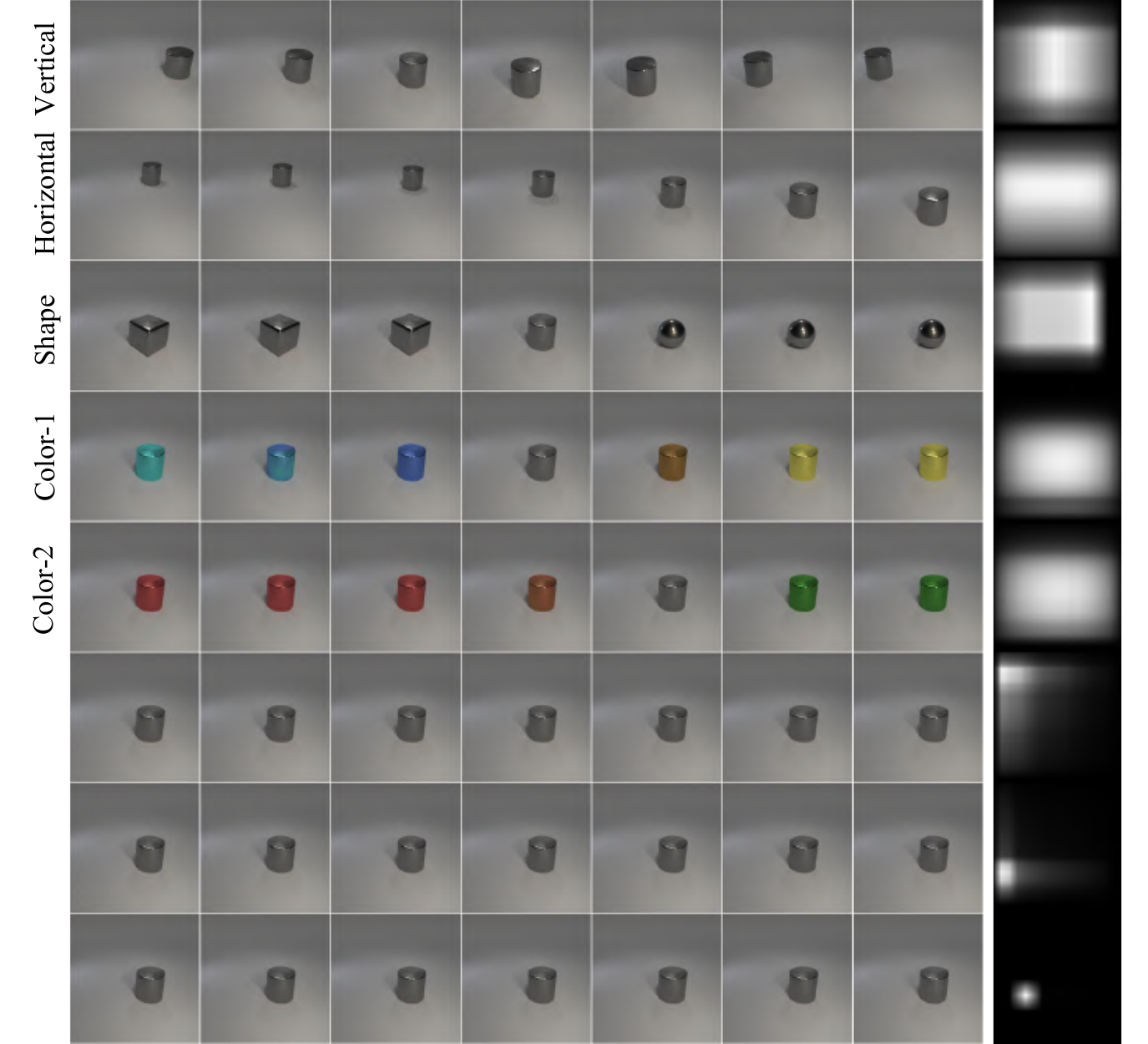}
\end{center}
    %\vspace{-8pt}
    \caption{Clevr-Simple 2.}
    %\vspace{-8pt}
\label{fig:clevr_simple_travs_1}
\end{figure}

\section{More Image Editing}
\label{ap:more_editing}
\begin{figure}[t]
\begin{center}
   \includegraphics[width=\linewidth]{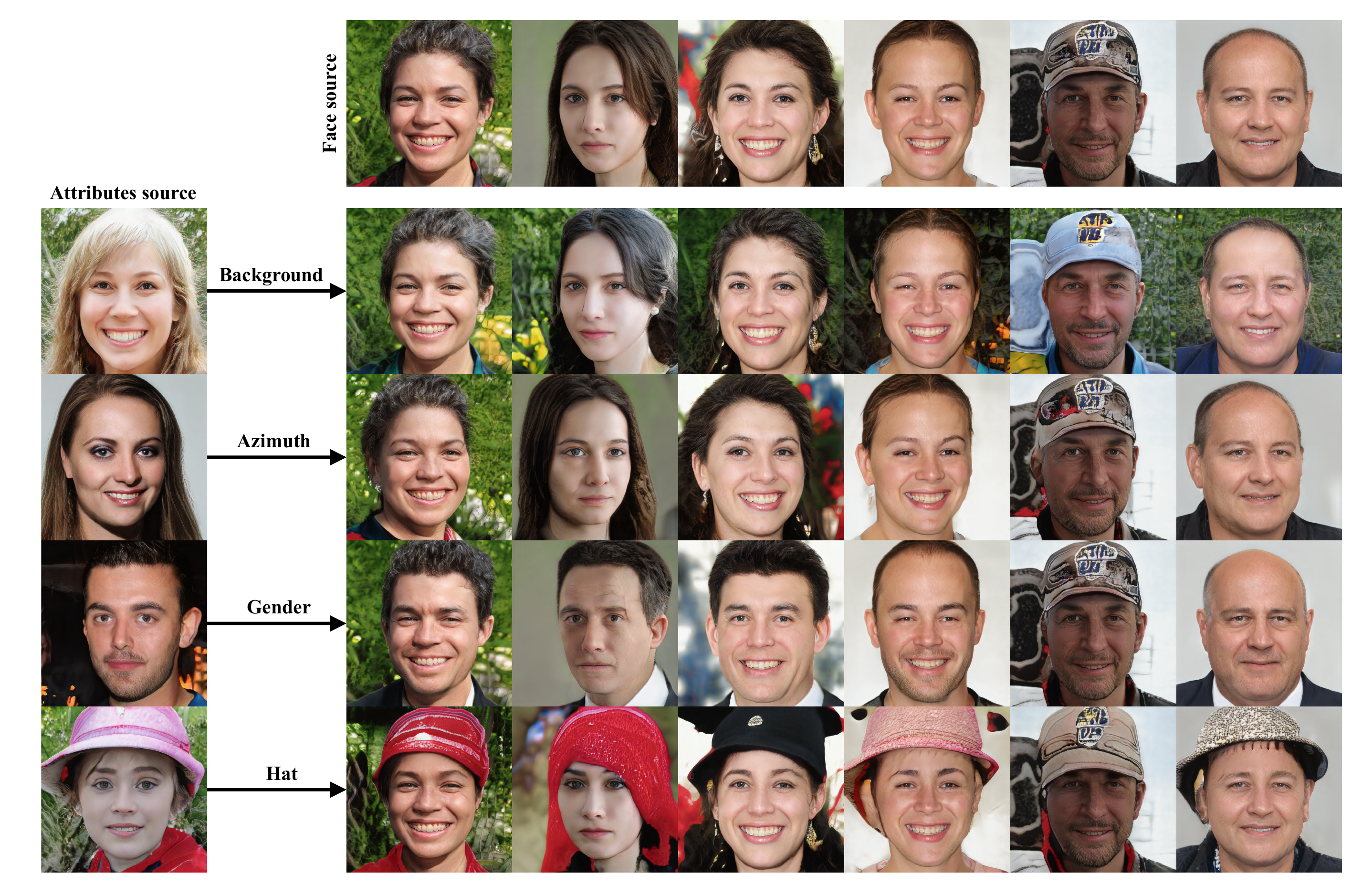}
\end{center}
    %\vspace{-8pt}
    \caption{Image editing by transferring attributes.}
    %\vspace{-8pt}
\label{fig:ap_image_editing_1}
\end{figure}

\begin{figure}[t]
\begin{center}
   \includegraphics[width=\linewidth]{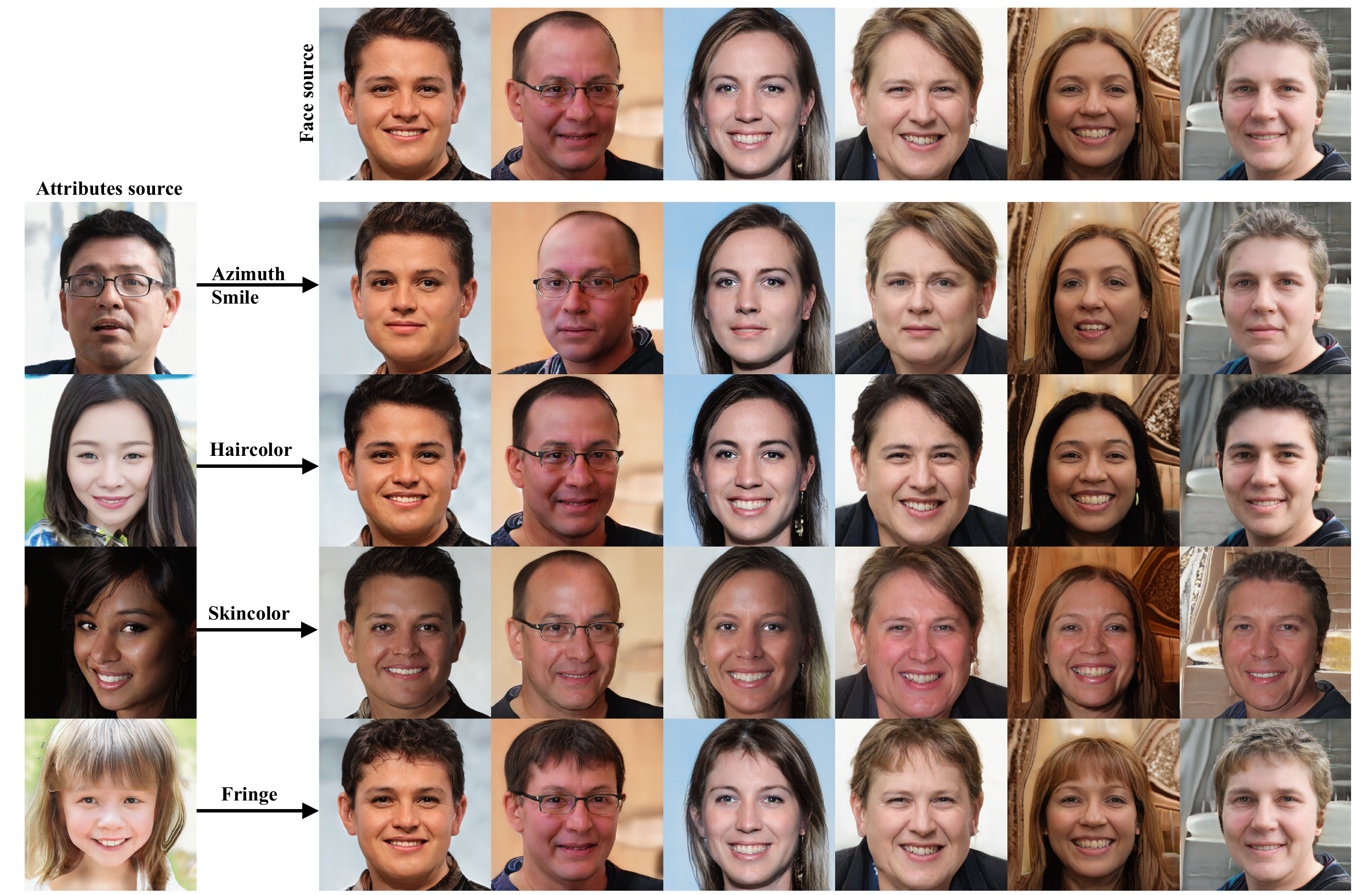}
\end{center}
    %\vspace{-8pt}
    \caption{Image editing by transferring attributes.}
    %\vspace{-8pt}
\label{fig:ap_image_editing_2}
\end{figure}

More image editing experiments (similar to Fig. \ref{fig:image_editing}) are
shown in Fig. \ref{fig:ap_image_editing_1} and Fig.
\ref{fig:ap_image_editing_2}.
We can see the attributes of background, azimuth, smile, hat, fringe,
skin color are successfully disentangled in the learned representation.
However, there are still some flaws, such as the transfer of \emph{hat}
in the last row of Fig. \ref{fig:ap_image_editing_2} changes the
style and color of hats in the resulting images.
This problem may be caused by the lack of data of various hats during
training, but may also be alleviated by better models.

\section{Implementations}
\label{ap:implementations}
The network architectures are shown in multiple Tables:
on CelebA: \ref{table:gen_celeba},
\ref{table:dis_celeba}, \ref{table:recog_celeba};
on Shoes+Edges: \ref{table:gen_shoes},
\ref{table:dis_shoes}, \ref{table:recog_shoes};
on Clevr-Simple and Clevr-Complex: \ref{table:gen_clevr},
\ref{table:dis_clevlr}, \ref{table:recog_clevr};
on FFHQ: \ref{table:gen_ffhq},
\ref{table:dis_ffhq}, \ref{table:recog_ffhq};
on DSprites:
\ref{table:gen_dsprites}, \ref{table:dis_dsprites},
\ref{table:recog_dsprites};
on 3DShapes: \ref{table:gen_3dshapes},
\ref{table:dis_3dshapes}, \ref{table:recog_3dshapes};
All models are trained with the Adam optimizer. For both generator and
discriminator optimizers, $\beta_{1}=0$, $\beta_{2}=0.99$,
initial learning rate is 0.002, except for 3DShapes we set 0.005. Note that
the $Q$ network is trained together with the generator.
The $\lambda$ for Shoes+Edges, Clevr-Complex, CelebA, FFHQ is 0.01,
for Clevr-Simple is 0.05,
for DSprites and 3DShapes is 0.001.
The $p_{\text{var}}$ is 0.2 for DSprites and 3DShapes,
and is 1 for other datasets.
For CelebA dataset, models are trained for 4,000k images (around 19 epochs).
For FFHQ dataset, models are trained until FID starts to saturate
(around 28 epochs).
For DSprites dataset, models are trained for 15,000k images (around 20 epochs).
For 3DShapes dataset, models are trained for 8,000k images (around 18 epochs).
For DSprites and 3DShapes, instead of randomly sampling the perturbed
dimension $k$ in the PC loss every iteration,
we loop the $k$ sequentially for each
dimension every 1k images to achieve a stabler convergence.

\begin{table}
    \begin{center}
        \begin{tabular} {l l}
            \toprule
                    Layer & Out Shape \\
                \midrule
                    Const & 4x4x512 \\
                \midrule
                    ResConv-up-1 & 8x8x512 \\
                \midrule
                    SC-4-5 & 8x8x512 \\
                \midrule
                    ResConv-id-1 & 8x8x512 \\
                \midrule
                    Noise-2 & 8x8x512 \\
                \midrule
                    ResConv-up-1 & 16x16x512 \\
                \midrule
                    SC-6-5 & 16x16x512 \\
                \midrule
                    ResConv-id-1 & 16x16x512 \\
                \midrule
                    Noise-2 & 16x16x512 \\
                \midrule
                    ResConv-up-1 & 32x32x512 \\
                \midrule
                    SC-6-5 & 32x32x512 \\
                \midrule
                    ResConv-id-1 & 32x32x512 \\
                \midrule
                    Noise-2 & 32x32x256 \\
                \midrule
                    ResConv-up-1 & 64x64x256 \\
                \midrule
                    SC-6-5 & 64x64x256 \\
                \midrule
                    ResConv-id-1 & 64x64x256 \\
                \midrule
                    Noise-2 & 64x64x256 \\
                \midrule
                    ResConv-up-1 & 128x128x256 \\
                \midrule
                    SC-4-5 & 128x128x256 \\
                \midrule
                    ResConv-id-1 & 128x128x128 \\
                \midrule
                    Noise-2 & 128x128x128 \\
                \midrule
                    ResConv-id-2 & 128x128x128 \\
                \midrule
                    ResConv-id-1 & 128x128x3 \\
                \bottomrule
            \end{tabular}
    \end{center}
    \caption{Generator on CelebA.}
    \label{table:gen_celeba}
\end{table}
\begin{table}
    \begin{center}
        \begin{tabular} {l l}
            \toprule
                    Layer & Out Shape\\
                \midrule
                    ResConv-down-2 & 128x128x64 \\
                \midrule
                    ResConv-down-2 & 64x64x64 \\
                \midrule
                    ResConv-down-2 & 32x32x128 \\
                \midrule
                    ResConv-down-2 & 16x16x256 \\
                \midrule
                    ResConv-down-2 & 8x8x512 \\
                \midrule
                    ResConv-down-2 & 4x4x512 \\
                \midrule
                    Conv-id-1 & 4x4x512 \\
                \midrule
                    Dense-1 & 1 \\
                \bottomrule
        \end{tabular}
    \end{center}
    \caption{Discriminator on CelebA.}
    \label{table:dis_celeba}
\end{table}
\begin{table}
    \begin{center}
        \begin{tabular} {l l}
            \toprule
                    Layer & Out Shape\\
                \midrule
                    ResConv-down-2 & 256x256x64 \\
                \midrule
                    ResConv-down-2 & 128x128x64 \\
                \midrule
                    ResConv-down-2 & 64x64x64 \\
                \midrule
                    ResConv-down-2 & 32x32x128 \\
                \midrule
                    ResConv-down-2 & 16x16x256 \\
                \midrule
                    ResConv-down-2 & 8x8x512 \\
                \midrule
                    ResConv-down-2 & 4x4x512 \\
                \midrule
                    Conv-id-1 & 4x4x512 \\
                \midrule
                    Dense-1 & 25 \\
                \bottomrule
        \end{tabular}
    \end{center}
    \caption{Recognizer $Q$ on CelebA.}
    \label{table:recog_celeba}
\end{table}

\begin{table}
    \begin{center}
        \begin{tabular} {l l}
            \toprule
                    Layer & Out Shape \\
                \midrule
                    Const & 4x4x512 \\
                \midrule
                    ResConv-up-1 & 8x8x512 \\
                \midrule
                    SC-4-3 & 8x8x512 \\
                \midrule
                    ResConv-id-1 & 8x8x512 \\
                \midrule
                    Noise-2 & 8x8x512 \\
                \midrule
                    ResConv-up-1 & 16x16x512 \\
                \midrule
                    SC-6-3 & 16x16x512 \\
                \midrule
                    ResConv-id-1 & 16x16x256 \\
                \midrule
                    Noise-2 & 16x16x256 \\
                \midrule
                    ResConv-up-1 & 32x32x256 \\
                \midrule
                    SC-6-3 & 32x32x256 \\
                \midrule
                    ResConv-id-1 & 32x32x256 \\
                \midrule
                    Noise-2 & 32x32x128 \\
                \midrule
                    ResConv-up-1 & 64x64x128 \\
                \midrule
                    SC-6-3 & 64x64x128 \\
                \midrule
                    ResConv-id-1 & 64x64x128 \\
                \midrule
                    Noise-2 & 64x64x128 \\
                \midrule
                    ResConv-up-1 & 128x128x128 \\
                \midrule
                    SC-4-3 & 128x128x128 \\
                \midrule
                    ResConv-id-1 & 128x128x64 \\
                \midrule
                    Noise-2 & 128x128x64 \\
                \midrule
                    ResConv-id-2 & 128x128x64 \\
                \midrule
                    ResConv-id-1 & 128x128x3 \\
                \bottomrule
            \end{tabular}
    \end{center}
    \caption{Generator on Shoes+Edges.}
    \label{table:gen_shoes}
\end{table}
\begin{table}
    \begin{center}
        \begin{tabular} {l l}
            \toprule
                    Layer & Out Shape\\
                \midrule
                    ResConv-down-2 & 128x128x64 \\
                \midrule
                    ResConv-down-2 & 64x64x64 \\
                \midrule
                    ResConv-down-2 & 32x32x64 \\
                \midrule
                    ResConv-down-2 & 16x16x128 \\
                \midrule
                    ResConv-down-2 & 8x8x256 \\
                \midrule
                    ResConv-down-2 & 4x4x512 \\
                \midrule
                    Conv-id-1 & 4x4x512 \\
                \midrule
                    Dense-1 & 1 \\
                \bottomrule
        \end{tabular}
    \end{center}
    \caption{Discriminator on Shoes+Edges.}
    \label{table:dis_shoes}
\end{table}
\begin{table}
    \begin{center}
        \begin{tabular} {l l}
            \toprule
                    Layer & Out Shape\\
                \midrule
                    ResConv-down-2 & 128x128x64 \\
                \midrule
                    ResConv-down-2 & 64x64x64 \\
                \midrule
                    ResConv-down-2 & 32x32x64 \\
                \midrule
                    ResConv-down-2 & 16x16x128 \\
                \midrule
                    ResConv-down-2 & 8x8x256 \\
                \midrule
                    ResConv-down-2 & 4x4x512 \\
                \midrule
                    Conv-id-1 & 4x4x512 \\
                \midrule
                    Dense-1 & 15 \\
                \bottomrule
        \end{tabular}
    \end{center}
    \caption{Recognizer $Q$ on Shoes+Edges.}
    \label{table:recog_shoes}
\end{table}

\begin{table}
    \begin{center}
        \begin{tabular} {l l}
            \toprule
                    Layer & Out Shape \\
                \midrule
                    Const & 4x4x512 \\
                \midrule
                    ResConv-up-1 & 8x8x512 \\
                \midrule
                    SC-1-2 & 8x8x512 \\
                \midrule
                    Noise-1 & 8x8x512 \\
                \midrule
                    ResConv-up-1 & 16x16x512 \\
                \midrule
                    SC-1-2 & 16x16x512 \\
                \midrule
                    Noise-1 & 16x16x256 \\
                \midrule
                    ResConv-up-1 & 32x32x256 \\
                \midrule
                    SC-1-2 & 32x32x256 \\
                \midrule
                    Noise-1 & 32x32x256 \\
                \midrule
                    ResConv-up-1 & 64x64x128 \\
                \midrule
                    SC-1-2 & 64x64x128 \\
                \midrule
                    Noise-1 & 64x64x128 \\
                \midrule
                    ResConv-up-1 & 128x128x128 \\
                \midrule
                    SC-1-2 & 128x128x128 \\
                \midrule
                    Noise-1 & 128x128x64 \\
                \midrule
                    ResConv-up-1 & 256x256x64 \\
                \midrule
                    Noise-1 & 256x256x64 \\
                \midrule
                    ResConv-id-1 & 256x256x3 \\
                \bottomrule
            \end{tabular}
    \end{center}
    \caption{Generator on Clevr-Simple and -Complex.}
    \label{table:gen_clevr}
\end{table}
\begin{table}
    \begin{center}
        \begin{tabular} {l l}
            \toprule
                    Layer & Out Shape\\
                \midrule
                    ResConv-down-2 & 256x256x64 \\
                \midrule
                    ResConv-down-2 & 128x128x64 \\
                \midrule
                    ResConv-down-2 & 64x64x64 \\
                \midrule
                    ResConv-down-2 & 32x32x64 \\
                \midrule
                    ResConv-down-2 & 16x16x128 \\
                \midrule
                    ResConv-down-2 & 8x8x256 \\
                \midrule
                    ResConv-down-2 & 4x4x512 \\
                \midrule
                    Conv-id-1 & 4x4x512 \\
                \midrule
                    Dense-1 & 1 \\
                \bottomrule
        \end{tabular}
    \end{center}
    \caption{Discriminator on Clevr-Simple and -Complex.}
    \label{table:dis_clevlr}
\end{table}
\begin{table}
    \begin{center}
        \begin{tabular} {l l}
            \toprule
                    Layer & Out Shape\\
                \midrule
                    ResConv-down-2 & 256x256x64 \\
                \midrule
                    ResConv-down-2 & 128x128x64 \\
                \midrule
                    ResConv-down-2 & 64x64x64 \\
                \midrule
                    ResConv-down-2 & 32x32x64 \\
                \midrule
                    ResConv-down-2 & 16x16x128 \\
                \midrule
                    ResConv-down-2 & 8x8x256 \\
                \midrule
                    ResConv-down-2 & 4x4x512 \\
                \midrule
                    Conv-id-1 & 4x4x512 \\
                \midrule
                    Dense-1 & 10 \\
                \bottomrule
        \end{tabular}
    \end{center}
    \caption{Recognizer $Q$ on Clevr-Simple and -Complex.}
    \label{table:recog_clevr}
\end{table}

\begin{table}
    \begin{center}
        \begin{tabular} {l l}
            \toprule
                    Layer & Out Shape \\
                \midrule
                    Const & 4x4x512 \\
                \midrule
                    ResConv-up-1 & 8x8x512 \\
                \midrule
                    SC-4-5 & 8x8x512 \\
                \midrule
                    ResConv-id-1 & 8x8x512 \\
                \midrule
                    Noise-2 & 8x8x512 \\
                \midrule
                    ResConv-up-1 & 16x16x512 \\
                \midrule
                    SC-4-5 & 16x16x512 \\
                \midrule
                    ResConv-id-1 & 16x16x512 \\
                \midrule
                    Noise-2 & 16x16x512 \\
                \midrule
                    ResConv-up-1 & 32x32x512 \\
                \midrule
                    SC-4-5 & 32x32x512 \\
                \midrule
                    ResConv-id-1 & 32x32x512 \\
                \midrule
                    Noise-2 & 32x32x256 \\
                \midrule
                    ResConv-up-1 & 64x64x256 \\
                \midrule
                    SC-4-5 & 64x64x256 \\
                \midrule
                    ResConv-id-1 & 64x64x256 \\
                \midrule
                    Noise-2 & 64x64x256 \\
                \midrule
                    ResConv-up-1 & 128x128x256 \\
                \midrule
                    SC-4-5 & 128x128x256 \\
                \midrule
                    ResConv-id-1 & 128x128x128 \\
                \midrule
                    Noise-2 & 128x128x128 \\
                \midrule
                    ResConv-up-1 & 256x256x128 \\
                \midrule
                    SC-4-5 & 256x256x128 \\
                \midrule
                    ResConv-id-1 & 256x256x128 \\
                \midrule
                    Noise-2 & 256x256x64 \\
                \midrule
                    ResConv-id-1 & 256x256x64 \\
                \midrule
                    ResConv-up-1 & 512x512x64 \\
                \midrule
                    ResConv-id-2 & 512x512x64 \\
                \midrule
                    ResConv-id-1 & 512x512x3 \\
                \bottomrule
            \end{tabular}
    \end{center}
    \caption{Generator on FFHQ.}
    \label{table:gen_ffhq}
\end{table}
\begin{table}
    \begin{center}
        \begin{tabular} {l l}
            \toprule
                    Layer & Out Shape\\
                \midrule
                    ResConv-down-2 & 256x256x64 \\
                \midrule
                    ResConv-down-2 & 128x128x64 \\
                \midrule
                    ResConv-down-2 & 64x64x64 \\
                \midrule
                    ResConv-down-2 & 32x32x128 \\
                \midrule
                    ResConv-down-2 & 16x16x256 \\
                \midrule
                    ResConv-down-2 & 8x8x512 \\
                \midrule
                    ResConv-down-2 & 4x4x512 \\
                \midrule
                    Conv-id-1 & 4x4x512 \\
                \midrule
                    Dense-1 & 1 \\
                \bottomrule
        \end{tabular}
    \end{center}
    \caption{Discriminator on FFHQ.}
    \label{table:dis_ffhq}
\end{table}
\begin{table}
    \begin{center}
        \begin{tabular} {l l}
            \toprule
                    Layer & Out Shape\\
                \midrule
                    ResConv-down-2 & 256x256x64 \\
                \midrule
                    ResConv-down-2 & 128x128x64 \\
                \midrule
                    ResConv-down-2 & 64x64x64 \\
                \midrule
                    ResConv-down-2 & 32x32x128 \\
                \midrule
                    ResConv-down-2 & 16x16x256 \\
                \midrule
                    ResConv-down-2 & 8x8x512 \\
                \midrule
                    ResConv-down-2 & 4x4x512 \\
                \midrule
                    Conv-id-1 & 4x4x512 \\
                \midrule
                    Dense-1 & 30 \\
                \bottomrule
        \end{tabular}
    \end{center}
    \caption{Recognizer $Q$ on FFHQ.}
    \label{table:recog_ffhq}
\end{table}

\begin{table}
    \begin{center}
        \begin{tabular} {l l}
            \toprule
                    Layer & Out Shape \\
                \midrule
                    Const & 4x4x512 \\

                \midrule
                    AdaIN-1 & 4x4x128 \\
                \midrule
                    Conv-id-1 & 4x4x128 \\
                \midrule
                    AdaIN-1 & 4x4x128 \\
                \midrule
                    Conv-id-1 & 4x4x64 \\
                \midrule
                    AdaIN-1 & 4x4x64 \\
                \midrule
                    Conv-up-1 & 8x8x64 \\

                \midrule
                    AdaIN-1 & 8x8x64 \\
                \midrule
                    Conv-id-1 & 8x8x32 \\
                \midrule
                    AdaIN-1 & 8x8x32 \\
                \midrule
                    Conv-id-1 & 8x8x32 \\
                \midrule
                    AdaIN-1 & 8x8x32 \\
                \midrule
                    Conv-up-1 & 16x16x16 \\

                \midrule
                    AdaIN-1 & 16x16x16 \\
                \midrule
                    Conv-id-1 & 16x16x16 \\
                \midrule
                    AdaIN-1 & 16x16x16 \\
                \midrule
                    Conv-id-1 & 16x16x16 \\
                \midrule
                    AdaIN-1 & 16x16x16 \\
                \midrule
                    Conv-up-1 & 32x32x16 \\

                \midrule
                    Conv-id-1 & 32x32x16\\
                \midrule
                    Conv-up-1 & 64x64x16\\
                \midrule
                    Conv-id-1 & 64x64x1\\
                \bottomrule
            \end{tabular}
    \end{center}
    \caption{Generator on DSprites.}
    \label{table:gen_dsprites}
\end{table}
\begin{table}
    \begin{center}
        \begin{tabular} {l l}
            \toprule
                    Layer & Out Shape\\
                \midrule
                    ResConv-down-2 & 64x64x16 \\
                \midrule
                    ResConv-down-2 & 32x32x16 \\
                \midrule
                    ResConv-down-2 & 16x16x32 \\
                \midrule
                    ResConv-down-2 & 8x8x64 \\
                \midrule
                    ResConv-down-2 & 4x4x128 \\
                \midrule
                    Conv-id-1 & 4x4x128 \\
                \midrule
                    Dense-1 & 1 \\
                \bottomrule
        \end{tabular}
    \end{center}
    \caption{Discriminator on DSprites.}
    \label{table:dis_dsprites}
\end{table}
\begin{table}
    \begin{center}
        \begin{tabular} {l l}
            \toprule
                    Layer & Out Shape\\
                \midrule
                    ResConv-down-2 & 64x64x16 \\
                \midrule
                    ResConv-down-2 & 32x32x16 \\
                \midrule
                    ResConv-down-2 & 16x16x16 \\
                \midrule
                    ResConv-down-2 & 8x8x16 \\
                \midrule
                    ResConv-down-2 & 4x4x32 \\
                \midrule
                    Conv-id-1 & 4x4x32 \\
                \midrule
                    Dense-1 & 9 \\
                \bottomrule
        \end{tabular}
    \end{center}
    \caption{Recognizer $Q$ on DSprites.}
    \label{table:recog_dsprites}
\end{table}

\clearpage

\begin{table}
    \begin{center}
        \begin{tabular} {l l}
            \toprule
                    Layer & Out Shape \\
                \midrule
                    Const & 4x4x512 \\
                \midrule
                    Conv-up-1 & 8x8x256 \\
                \midrule
                    SC-3-2 & 8x8x256 \\
                \midrule
                    ResConv-id-1 & 8x8x256 \\
                \midrule
                    SC-4-3 & 8x8x256 \\
                \midrule
                    ResConv-id-1 & 8x8x128 \\
                \midrule
                    Conv-up-1 & 16x16x128 \\
                \midrule
                    SC-4-3 & 16x16x128 \\
                \midrule
                    ResConv-id-1 & 16x16x64 \\
                \midrule
                    SC-4-3 & 16x16x64 \\
                \midrule
                    ResConv-id-1 & 16x16x64 \\
                \midrule
                    Conv-up-1 & 32x32x64 \\
                \midrule
                    ResConv-id-1 & 32x32x32 \\
                \midrule
                    Conv-up-1 & 64x64x32 \\
                \midrule
                    ResConv-id-1 & 64x64x32 \\
                \midrule
                    Conv-id-1 & 64x64x3 \\
                \bottomrule
            \end{tabular}
    \end{center}
    \caption{Generator on 3DShapes.}
    \label{table:gen_3dshapes}
\end{table}
\begin{table}
    \begin{center}
        \begin{tabular} {l l}
            \toprule
                    Layer & Out Shape\\
                \midrule
                    ResConv-down-2 & 64x64x32 \\
                \midrule
                    ResConv-down-2 & 32x32x32 \\
                \midrule
                    ResConv-down-2 & 16x16x32 \\
                \midrule
                    ResConv-down-2 & 8x8x64 \\
                \midrule
                    ResConv-down-2 & 4x4x128 \\
                \midrule
                    Conv-id-1 & 4x4x128 \\
                \midrule
                    Dense-1 & 1 \\
                \bottomrule
        \end{tabular}
    \end{center}
    \caption{Discriminator on 3DShapes.}
    \label{table:dis_3dshapes}
\end{table}
\begin{table}
    \begin{center}
        \begin{tabular} {l l}
            \toprule
                    Layer & Out Shape\\
                \midrule
                    ResConv-down-2 & 64x64x32 \\
                \midrule
                    ResConv-down-2 & 32x32x32 \\
                \midrule
                    ResConv-down-2 & 16x16x32 \\
                \midrule
                    ResConv-down-2 & 8x8x64 \\
                \midrule
                    ResConv-down-2 & 4x4x128 \\
                \midrule
                    Conv-id-1 & 4x4x128 \\
                \midrule
                    Dense-1 & 12 \\
                \bottomrule
        \end{tabular}
    \end{center}
    \caption{Recognizer $Q$ on 3DShapes.}
    \label{table:recog_3dshapes}
\end{table}

\end{document}